\documentclass[Afour,sageh,times]{sagej}

\usepackage{moreverb,url}

\usepackage[utf8]{inputenc}
\usepackage{nomencl}
\makenomenclature
\usepackage{etoolbox}
\usepackage{imakeidx}
\usepackage{gensymb}
\usepackage{fancyhdr,graphicx}

\renewcommand\nomgroup[1]{%
  \item[\bfseries
  \ifstrequal{#1}{R}{Robot Trajectory}{%
  \ifstrequal{#1}{C}{Coordinate System}{%
  \ifstrequal{#1}{S}{Synthetic Aperture Radar}{}}}%
]}

\setcitestyle{authoryear,open={(},close={)},aysep={,}}
\usepackage{multicol}
\usepackage{commath}
\usepackage{amssymb}
\usepackage{siunitx}
\usepackage{graphicx}
\usepackage{xcolor}
\usepackage[normalem]{ulem}
\usepackage{soul}

\newtheorem{theorem}{Theorem}
\newtheorem{problem}{Problem}
\newtheorem{lemma}[theorem]{Lemma}

\newtheoremstyle{definition}
  {\topsep}
  {\topsep}
  {}
  {}
  {\itshape}
  {}
  {.3em}
  {\thmname{#1}\thmnumber{ #2}\thmnote{ (#3)}}
\theoremstyle{definition}
\newtheorem{definition}{Definition}

\newtheoremstyle{assumption}
  {\topsep}
  {\topsep}
  {}
  {}
  {\itshape}
  {}
  {.3em}
  {\thmname{#1}\thmnumber{ #2}\thmnote{ (#3)}}
\theoremstyle{assumption}

\newtheorem*{assumption*}{Assumptions}

\newtheoremstyle{case}
  {\topsep}
  {\topsep}
  {}
  {}
  {\itshape}
  {:}
  {.5em}
  {\thmname{#1}\thmnumber{ #2}\thmnote{ (#3)}}
\theoremstyle{definition}

\usepackage[font=small,labelfont=bf,
   justification=justified,
   format=plain]{caption} 

\usepackage{float}
\newcommand\BibTeX{{\rmfamily B\kern-.05em \textsc{i\kern-.025em b}\kern-.08em
T\kern-.1667em\lower.7ex\hbox{E}\kern-.125emX}}

\fancypagestyle{firstpage}
{
    \fancyhead[OL]{
    Please cite as follows: \textbf{Jadhav N, Wang W, Zhang D, Khatib O, Kumar S, Gil S. A wireless signal-based sensing framework for robotics. The International Journal of Robotics Research. 2022;41(11-12):955-992. doi:10.1177/02783649221097989}}
    
}
\renewcommand\headrule{
\begin{minipage}{1\textwidth}
\hrule width \hsize \kern 1mm \hrule width \hsize height 0pt 
\end{minipage}}%
\begin{document}

\title{A wireless signal-based sensing framework for robotics}

\author{Ninad Jadhav\affilnum{1}\affilnum{*}, Weiying Wang\affilnum{1}\affilnum{*}, Diana Zhang\affilnum{2}, Oussama Khatib\affilnum{3}, Swarun Kumar\affilnum{2} and Stephanie Gil\affilnum{1}}

\affiliation{\affilnum{1}Harvard University, US\\
\affilnum{2}Carnegie Mellon University, US\\
\affilnum{3}Stanford University, US\\
\affilnum{*}are co-primary authors}

\corrauth{Ninad Jadhav, REACT Lab, Harvard University, Boston, MA, US.\\
Weiying Wang, REACT Lab, Harvard University, Boston, MA, US
}
\email{\{njadhav, weiyingwang\}@g.harvard.edu}

\begin{abstract}
In this paper we develop the analytical framework for a novel Wireless signal-based Sensing
capability for Robotics (WSR) by leveraging robots' mobility. It allows robots to primarily measure relative direction, or Angle-of-Arrival (AOA), to other robots, while operating in non-line-of-sight unmapped environments and without requiring external infrastructure. We do so by capturing all of the paths that a wireless signal traverses as it travels from a transmitting to a receiving robot in the team, which we term as an \emph{AOA profile}. The key intuition behind our approach is to enable a robot to emulate \emph{antenna arrays} as it moves freely in 2D and 3D space. The small differences in the phase of the wireless signals are thus processed with knowledge of robots' local displacement to obtain the profile, via a method akin to Synthetic Aperture Radar (SAR). The main contribution of this work is the development of i) a framework to accommodate arbitrary 2D and 3D motion, as well as continuous mobility of both signal transmitting and receiving robots, while computing AOA profiles between them and ii) a Cramer-Rao Bound analysis, based on antenna array theory, that provides a lower bound on the variance in AOA estimation as a function of the geometry of robot motion. This is a critical distinction with previous work on SAR-based methods that restricts robot mobility to prescribed motion patterns, does not generalize to the full 3D space, and/or requires transmitting robots to be stationary during data acquisition periods. We show that allowing robots to use their full mobility in 3D space while performing SAR, results in \emph{more accurate} AOA profiles and thus better AOA estimation. We formally characterize this observation as the \emph{informativeness} of the robots' motion; a computable quantity for which we derive a closed form. All analytical developments are substantiated by extensive simulation and hardware experiments on air/ground robot platforms using $5~\text{GHz}$ WiFi. Our experimental results bolster our analytical findings, demonstrating that 3D motion provides enhanced and consistent accuracy, with total AOA error of less than $10\degree$ for $95\%$ of trials. We also analytically characterize the impact of displacement estimation errors on the measured AOA and validate this empirically using robot displacements obtained from off-the-shelf Intel Tracking Camera T265. Finally, we demonstrate the performance of our system on a multi-robot task where a heterogeneous air/ground pair of robots continuously measure AOA profiles over a WiFi link to achieve dynamic rendezvous in an unmapped, $300~\text{m}^2$ environment with occlusions.
\end{abstract}

\maketitle
\thispagestyle{firstpage}
\section{Introduction} \label{sec:intro}
\begin{figure}
  \centering
   \textbf{Angle-of-Arrival profile}
   \vspace{5pt}
  \includegraphics[width=8cm,height=5.5cm]{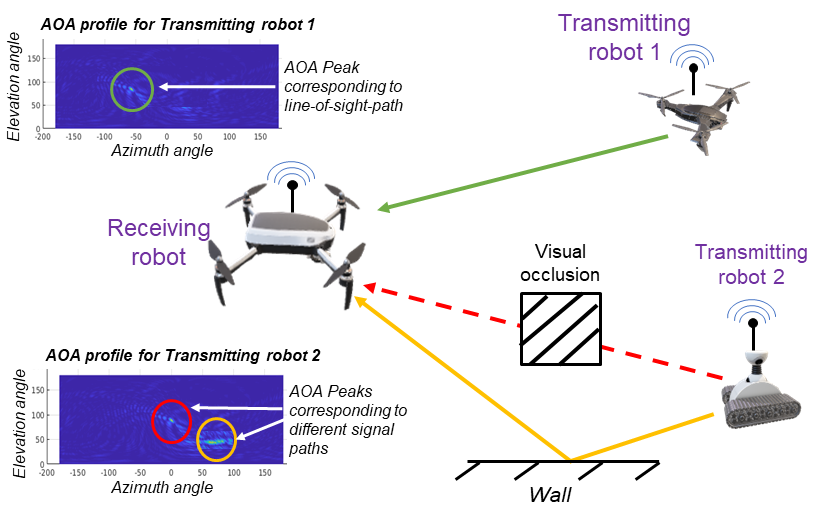}\caption{\footnotesize{\emph{Schematic representation of AOA profiles obtained by a signal receiving UAV for transmitting aerial and ground robot. Green, Red and orange lines represent the line-of-sight, direct and reflected signal paths respectively. These multipaths are captured in the corresponding AOA profiles as different AOA peaks. This profile can be used to obtain relative bearing between robots, establishing adhoc robot networks or as a fingerprint to verify uniqueness of a transmitting robot.}}}
  \label{fig:AOA_Profile}
  \vspace{-0.075in}
\end{figure}

Multi-robot coordination often requires robots to gather information about neighboring robots and their shared environment in order to collaborate effectively. Sensing state information of other robots such as their relative positions, proximity, or nature (friend or foe) is often needed for coordination. This information is critical for various multi-robot applications, ranging from coverage, to exploration, to rendezvous and beyond ~\citep{bulloCoverage, almadhoun2019survey, cunningham2012fully, Freda20193DMP, Choudhary2016MultiRO, graphslam}. However, how to obtain this information in an unknown and unmapped environment without a shared
coordinate reference frame, without line-of-sight
and without GPS or other external infrastructure, is a significant challenge. A promising approach for locally obtaining relative position related information locally, exploits information extracted from exchanged wireless signals \citep{Gil2015AdaptiveCI}. As a wireless signal travels between robots, it physically interacts with the environment. Its amplitude and phase are affected by the distance it travels, the obstacles it passes through or is reflected off of, and the relative positions and velocities of the communicating robots in a phenomenon called \emph{multipath}. This multipath can be measured as an Angle-of-Arrival (AOA) profile, where the arriving signal's \emph{direction and power along each path} is measured as peaks along different spatial angles (see Fig.~\ref{fig:AOA_Profile}), for a pair of communicating robots in a team. Measuring these AOA profiles would thus allow robots to obtain information about the relative direction of other robots that they are able to receive any communication from, including lightweight ping packets. \emph{We note that ping packets incur a very small fraction of available bandwidth, as little as 5 KB/s; thus robots can perform AOA estimation of one another at much longer distances, and through occlusions, even if they are not able to communicate other information, such as maps, typically requiring much higher bandwidths~\citep{Schmuck2019CCMSLAMRA}.} In this way, wireless signals could inform about a distant or occluded neighboring robot which would be difficult or impossible to obtain using traditional on-board sensors alone such as LiDAR or camera.

\begin{figure}
    \centering
   \textbf{Virtual antenna array}\\
   \vspace{5pt}
  \includegraphics[width=8.7cm,height=5.0cm]{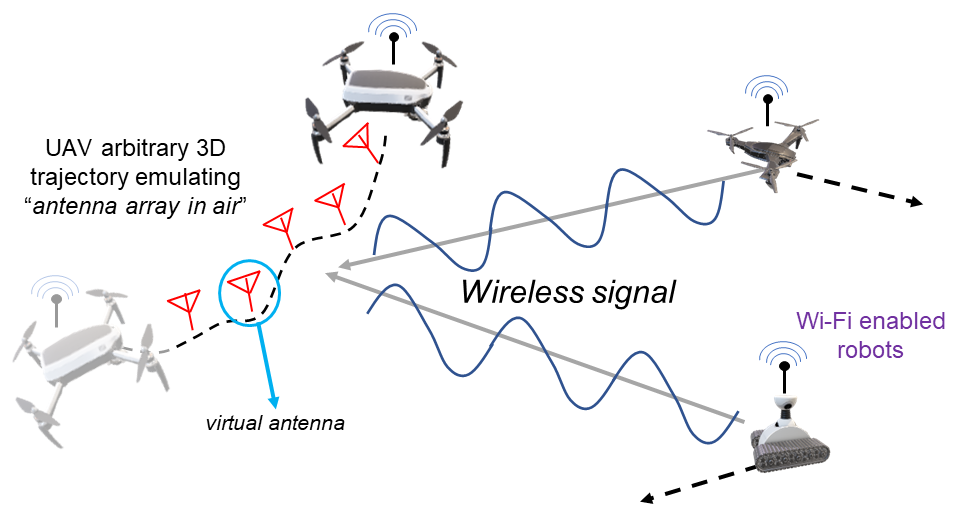}\caption{\footnotesize{\emph{A schematic representation of our approach that enables robots to use their mobility and received communication signals (e.g. WiFi) for emulating an antenna array. The dotted black lines indicate arbitrary robot displacements in 2D and 3D.}}}
  \label{fig:Paper_schematic}
  \vspace{-0.1in}
\end{figure}

In this paper, we develop the algorithmic and analytical framework for a \emph{Wireless signal-based Sensing capability for Robotics (WSR)}. This framework is applicable to any radio frequency wireless signal. It enables the robots to measure full AOA profiles between robots that are i) communicating locally by broadcasting packets over local wireless
network (e.g. WiFi), ii)  using only on-board sensors native to robotic
platforms for estimating local displacement and iii) freely mobile in 3D space. This framework would constitute a highly amenable design for fully distributed robotic systems. The key intuition for our approach builds off of Synthetic Aperture Radar (SAR) to measure small phase differences of a transmitted signal as the receiving robot moves through space. In this way, the receiving robot effectively emulates \emph{``antenna arrays''}, (See Fig. \ref{fig:Paper_schematic}) tracing out a virtual multi-antenna array (with one \emph{``antenna element''} for each received packet). The shape or \emph{array geometry} is dictated by the \emph{path traversed} by the robot. Antenna array theory indicates that a full AOA profile can be reconstructed from an array of minimum length of two times the signal wavelength (about $12~\text{cm}$ for $5~\text{GHz}$ WiFi) ~\citep{orfanidis2016electromagnetic}. This allows the receiving robot to resolve the incoming signal directions, using minimal displacements as it moves through space. \textbf{This paper shows for the first time that arbitrary displacements on general air/ground robotic platforms are indeed compatible with SAR-type approaches}.

Previous work has shown that measuring AOA profiles is possible, and that incorporating this information into multi-robot systems can be used to address several important long-standing challenges in robotics. For example: i) several AOA profiles collected from the same transmitting source can be used to triangulate the position of the sender, leading to accurate positioning and tracking in GPS-denied environments~\citep{ubicarse,spotFi,Xiong2013ArrayTrackAF,fadel3dTracking,Vasisht2016DecimeterLevelLW, fadelSeeThroughWalls,Karanam20173DTI}, ii) an AOA profile can be used by robots establish ad-hoc networks that would improve communication quality to other robots in the team, allowing for more reliable communication in complex and cluttered environments~\citep{Gil2015AdaptiveCI,Wang2019ActiveRF}, and iii) the full AOA profile can be used as a signal
multipath \emph{signature} of a robot to verify its uniqueness, with
implications for security and authentication in multi-agent
systems~\citep{Gil2015Spoof-Resilient,securearray}. However, a main limitation of many of these approaches is that they do not extend to general robotics platforms and full 3D mobility of multiple robots, while maintaining the ability to collect AOA profile information. \emph{In order to achieve true adoption of the WSR framework, we need to resolve this important limitation.} 

We do not focus on a particular application domain but rather focus on the development of their critical primitive -- AOA profile measurement capabilities and a characterization of their accuracy -- for general air/ground multi-robot systems. Advances in accurate pose estimation for a robot, using on-board sensors such as camera and LiDAR~\citep{Schmuck2019CCMSLAMRA, Choudhary2017DistributedMW,Qin2018VINSMonoAR} enable this new possibility of applying SAR over \emph{arbitrary} 2D and 3D displacement of robots. However, key challenges must be addressed. Namely, whether SAR-type approaches can be successfully employed over i) arbitrary 3D motion subject to displacement estimation error, and ii) in a scenario where both the transmitting and receiving robot are moving simultaneously. The focus of this paper is to address these challenges.

\subsection{Key challenges and approach}
We simultaneously address the challenges and opportunities imposed by arbitrary robot motion by appealing to antenna array theory~\citep{Vu2013-3D}. We first formulate a SAR equation to effectively isolate the phase change in the signal resulting from any arbitrary 2D or 3D robot displacement. Isolation here refers to isolation of the phase change from other sources of noise, such as Carrier Frequency Offset (CFO), and/or simultaneous mobility at the transmitting and receiving robots i.e \emph{``moving ends''}. Thus our developed methods allow for continuous data collection and AOA profile estimation, between any robots that can communicate (even at low rates of $\sim 5~\text{KB/s}$). 

Interestingly, the ability to perform SAR over arbitrary 3D robot motion holds potential for \emph{improving the accuracy of AOA profiles}. It is well known that different antenna array geometries display different sensitivity to measuring phase differences in a received signal \citep{manikas2004differential}. 
Hence, it is intuitive to expect that \emph{different robot motions should also impact the AOA profile and specifically the accuracy of the AOA peak that corresponds to the direction of maximum power path of received signal}. We formalize this intuition via a Cramer Rao Bound analysis. The Cramer Rao Bound (CRB) captures the sensitivity of AOA peak estimation, as it is a lower bound on its variance, resulting from a particular array geometry. In particular, the CRB is proportional to the inverse of the \emph{Fisher Information} for a specific antenna array geometry. We thus treat the inverse of the CRB as the \emph{informativeness} of the antenna array and seek to characterize this value for different motion patterns. Essentially, a key question that we address in this work is: \textbf{\emph{how to relate the informativeness of a robot's displacement to the resulting AOA profile?}} Along these lines we derive a new form of the CRB for analyzing the effect of arbitrary displacement geometry on the quality of the resulting AOA estimation. Our analysis reveals that indeed the ability to compute AOA profiles over a robot's arbitrary 3D motion provides a precise advantage as compared to 2D motion due to high informativeness of the former.

Additionally, we analytically characterize the impact of the robot's displacement estimation error, resulting from the use of on-board sensors, on the AOA estimation accuracy. Our analytical development shows that drift and high frequency noise about the estimated displacement mean manifests itself as a shift and attenuation of maximum magnitude AOA peak in the resulting profile. Moreover, we show that using short robot displacement over a small time window makes our system resilient to accumulating error in the robots' local pose estimation when operating for longer duration.

\medskip
\noindent Our analysis reveals several insights about the nature of how robots' motion influences AOA profiles, some of which are described below:
\begin{itemize}
\item \textbf{\ul{AOA profile reconstruction using arbitrary robot motion:}}. Short local displacement of a robot can be used to simultaneously measure AOA profiles for \emph{all} communication neighbor robots.
\item \textbf{\ul{Informativeness of robot displacement:}} Displacement along a random curve in 3D space is more informative (as determined by the CRB) than a regular linear displacement, as might typically be prescribed for SAR. Hardware experiments, using $5~\text{GHz}$ WiFi signals, support this analysis showing an error of less than $10\degree$ in AOA for $95\%$ of trials for arbitrary 3D displacement, whereas for planar 2D curved and linear displacement only $50\%$ and $7\%$ of trials show error below $10\degree$, respectively (c.f. Section~\ref{sec:cr-bound}).
\item \textbf{\ul{Influence of displacement estimation error on AOA estimation accuracy:}} AOA estimation will suffer a shift $\Delta$ and a peak attenuation of $\cos(\sigma/2)$ where $\Delta$ is linearly proportional to the drift in robots' displacement estimation and $\sigma$ is the maximum angular variation of the displacement estimation error about its mean (c.f. Lemma~\ref{lemma:trajError}). Our hardware experiments, indicate that this translates to a median estimation error of $7.58\degree$ in azimuth, $3.21\degree$ in elevation (use Fig. \ref{fig:AOA_SAR} schematic for reference) on average for $7\%$ displacement error ($0.2~\text{m}$ displacement estimation error for a $2.8~\text{m}$ robot displacement). 
\item \textbf{\ul{AOA profile estimation with moving ends:}} Obtaining AOA estimation using a SAR-based technique is possible even when both the transmitting and receiving robots are mobile during the signal sampling. We show that this is due to the translation agnostic property of phase change in wireless signals. AOA estimation can thus be accomplished under certain conditions (that we define) so long as robots can share their position estimates in their own local reference frame (c.f. Lemma~\ref{lemma:movingEnds}). 
\end{itemize}  


\noindent We substantiate the above observations with quantitative bounds on errors. These analytical results are applicable to any radio frequency (RF) signals, with practical hardware implementation requiring access to the signals' phase. WiFi is easily accessible on robots today with off-the-shelf modules capable of accessing signal phase~\citep{Halperin_csitool,nexmon_tool,atheros_tool}. Thus, to further validate our analytical results we use WiFi for extensive hardware experiments, demonstrating agreement in theory and in practice for every derived claim in the paper. Our implementation uses an off-the-shelf  Intel 5300 WiFi card to enable packet broadcasting with no pre-installed wireless routers and a VIO camera -- thereby demonstrating how our methods are designed for distributed implementation and seamless integration on today's robot platforms.

\subsection{Paper contributions} 
\begin{figure}
  \centering
   \textbf{Obtaining relative direction using AOA profile}\\
   \vspace{5pt}
  \includegraphics[width=8.7cm,height=7.0cm]{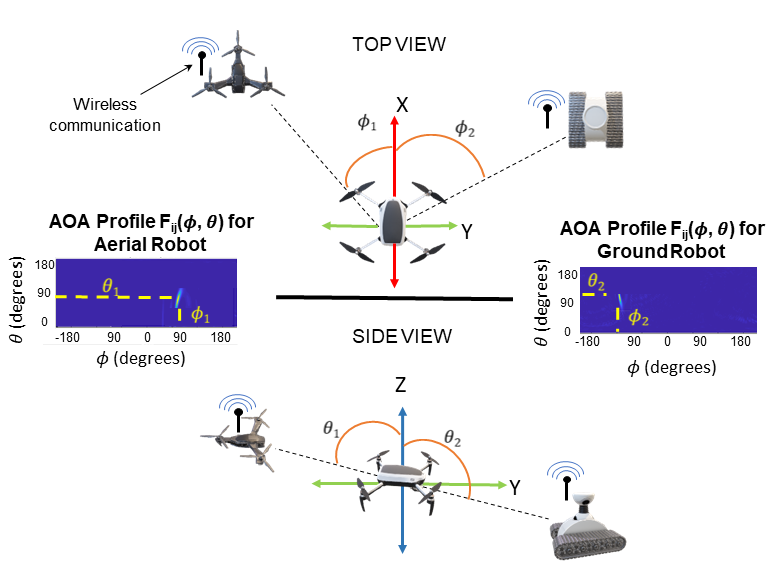}\caption{\footnotesize{\emph{Schematic showing AOA profile obtained by signal receiving UAV for transmitting ground and aerial robots. The Angle-of-Arrival ($\phi,\theta$) in azimuth (top view) and elevation (side view) respectively obtained from these profiles is highlighted in yellow.}}}
  \label{fig:AOA_SAR}
  \vspace{-0.1in}
\end{figure}

\begin{enumerate}
    \item We develop the algorithmic machinery for using robots' local displacement in 3D for achieving incoming signal direction i.e., AOA profile estimation (Fig. \ref{fig:AOA_SAR}).
    \item We derive the algorithmic framework for allowing simultaneous motion of both ends (transmitting and receiving robots) while obtaining a full and accurate AOA profile. This accommodates continuous motion and continuous AOA estimation during robot operation.
    \item We analytically characterize the impact of i) the \emph{informativeness} of robots' motion on the accuracy of the AOA estimation and ii) the displacement estimation error on the AOA estimation error.
    \item All analytical and algorithmic claims are supported with in-depth hardware experiments using $5~\text{GHz}$ WiFi signals, including a proof-of-concept application where AOA, based on our \emph{WSR} framework, is used to achieve fully dynamic rendezvous between an aerial robot and an occluded ground robot.
\end{enumerate}

\subsection{Paper roadmap} In the next section we provide a summary of previous work done in the context of robotics and wireless signal. Sec.~\ref{sec:Problem Formulation} formally states the problem statements and the definitions relevant to our development and analysis. Sec.~\ref{sec:Development} details the formulation of arbitrary 3D SAR and  Sec.~\ref{sec:moving_ends} shows how our system can handle signal transmitting robots that are mobile. Sec.~\ref{sec:Analysis_impact_factors} provides an analysis of the different factors that impact AOA calculation. We show the results of our hardware experiments for 5 $\texttt{GHz}$ WiFi signal, in Sec.~\ref{sec:results} and evaluate the capability of our system using a proof-of-concept application in Sec.~\ref{sec:hardware_application}. Directions for future research and concluding remarks are provided in Sec.~\ref{sec:discussion} and~\ref{sec:conclusion} respectively.

\section{Related Work}\label{sec:related_works}
In this section, we provide an overview of prior work on applications of wireless signals, specifically in the context of robotics. We first highlight key limitations of traditional sensors in obtaining relative information in multi-robot systems, as well as developments in the wireless community that explore the use of wireless signals as a sensing modality. Next we look at previous approaches on obtaining relative direction information between communicating robots and the corresponding open challenges.






\subsubsection*{\textbf{Relative state estimation in multi-robot systems:}}
Many multi-robot coordination problems such as coverage~\citep{bulloCoverage}, mapping~\citep{robotMapping_fingerprints_Roland,mappingMonocularSLAM,featuresMapping,cooperativeMapping_leonard}, frontier exploration~\citep{Gautam2017FASTS, Solanas2004CoordinatedME, Fox2005DistributedME,multiagentExploration_BurgardThrun}, and persistent surveillance~\citep{persistentSurveillance_experiments,persistentSurveillance} rely on knowledge of relative positions and other state information between collaborating robots. Although on-board GPS devices or other sensors such as camera and LiDAR, commonly used for such problems, provide information about inter-agent relative positions and the environment, they have limitations. In particular they often don't work in GPS-denied and communication constrained environments, in non-line-of-sight, over long distances, and/or when there is a lack of a shared coordinate frame and map. Thus, for many common environments such as those with clutter, those that require robots to operate at far distances from one another or those that lack common features required for fixing a shared coordinate frame, the current sensing methods easily get disoriented~\citep{SajadSaeedi2016MultipleRobotSL,multislamKshirsagar2018ASO}. In comparison, our method
uses wireless signals to obtain information about robots’ state
and its environment in ways that traditional sensors cannot easily achieve.
\begin{figure}
    \centering
        \hspace{15pt}\textbf{Robots used for hardware experiments}\\
        \vspace{5pt}
        \includegraphics[width=8.5cm,height=4.0cm]{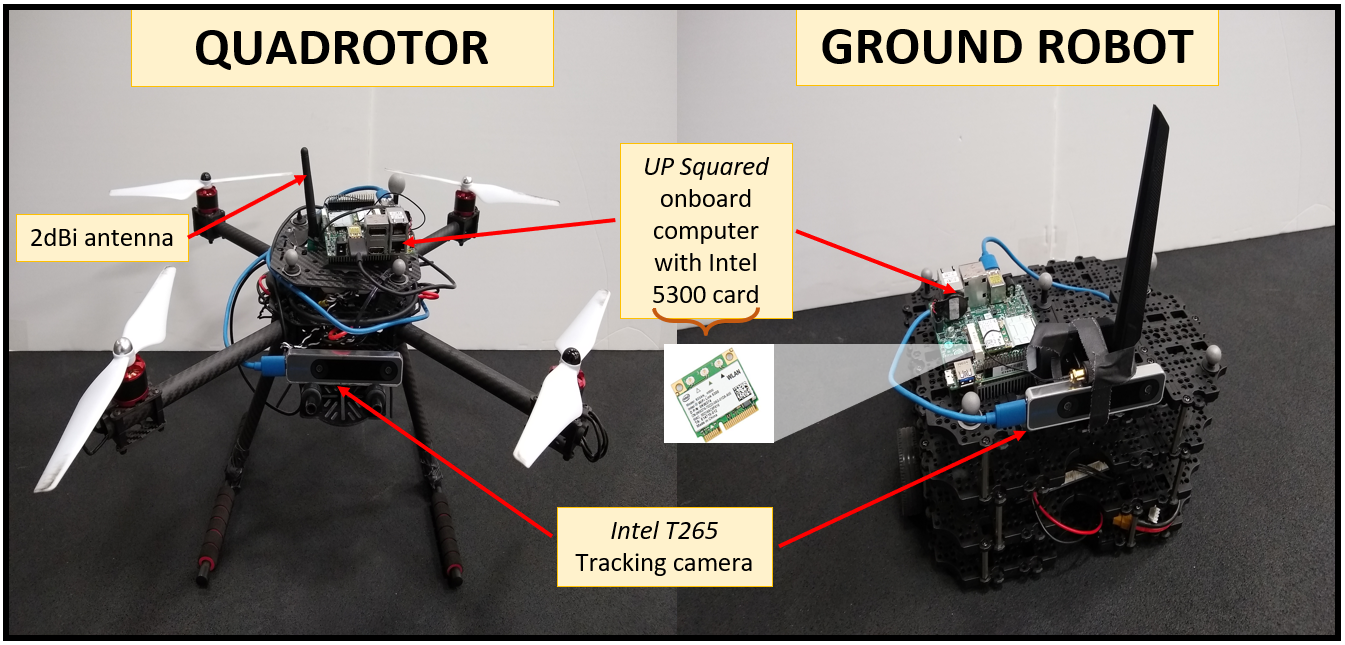}
    	\caption{\footnotesize{\emph{Aerial and ground robots used in hardware experiments equipped with different off-the-shelf on-board sensors. Each robot is equipped with an UP Squared on-board compute device that has Intel 5300 WiFi card, an omnidirectional antenna, and an Intel RealSense Tracking Camera T265 for obtaining local displacement estimates.}}}
    	\label{fig:hardware_robots}
    	\vspace{-0.1in}
\end{figure}

\subsubsection*{\textbf{Sensing using wireless signals:}}
Robotic platforms are generally equipped with different types of wireless communication hardwares \citep{CalvoFullana2019MultimodeAC}. Beyond enabling communication, wireless signals processing has shown great promise for sensing as well. For example, it has been shown that transmitted wireless signals between devices can be used to extract useful information such as device position~\citep{spotFi,Xiong2013ArrayTrackAF,lteye,PinIt,mostofiAOA,Cominelli2019DeadOA,Verma2018DOAEF,Cao2020HomecomingAW}. Other works have shown the ability to sense rich information about the environment as well, from tracking and imaging behind walls~\citep{fadelThroughWallSensing,mostofi_throughWall_wirelessSensing,fadelSeeThroughWalls,mostofi_wirelessSensing_mapping}, to materials sensing~\citep{swarun_materialSensing, fadel_foodSensing, Dhekne2018LiquIDAW}, imaging in harsh visual conditions \citep{Guan2020ThroughFH} and even shape sensing~\citep{swarun_shapeSensing}. While these works undoubtedly show the promise of using wireless signals to obtain useful information about a robots' state and its environment in ways that more traditional sensors like cameras and LiDAR cannot, they are not necessarily compatible with general robotics applications. These methods sometimes use specialized hardware, pre-deployed stationary infrastructure or even specialized wireless signals that are not readily available on robot platforms. Often, they also do not consider mobility (both signal transmitting and receiving robot), which can have a profound impact on the accuracy of these methods. Thus, without accounting for substantial deployment effort, they are not inherently designed to be seamlessly integrated with robotic platforms.

For ease of deployment, some methods use Received Signal Strength (RSS), which refers to the magnitude of the signal \citep{Zhu2013RSSIbasedAF}. However capturing substantial changes in signal magnitude require large displacements of the receiving robot and/or requires sampling along multiple directions. This makes RSS coarse and hence ill-suited for extracting information with accuracy and precision over short robot displacements. A more promising approach is to use signal phase, which is sensitive to even small amounts of displacement. This leads to higher granularity and accuracy of information compared to RSS ~\citep{Yang2013FromRT,Ma2019WiFiSW}. However, this information is not easily available for many commercial off-the-shelf wireless signal modules. 

Significant attention has also been given to UWB as sensing modalities for robotic platforms. TrackIO~\citep{trackIO} deployed UAV-UWB system to estimate first responders' location and velocity even in deep indoors environment, while the autonomous Puffer project~\citep{Boroson2020InterRobotRM} used UWB for pose graph optimization. However, UWB typically need to operate on wide bandwidths resulting in significant FCC restrictions on operating range and power~\citep{fccban_uwb}. Thus, UWB is not yet native to current robot platforms and commercial UWB modules require pre-calibration and additional customization using physical antennas to obtain AOA in both azimuth and elevation~\citep{ULoc}.

Our method is designed to be completely native to robotic platforms. Although our framework is applicable to any RF signal, we use WiFi for experimental evaluation since it is more ubiquitous to current robot platforms. The signal phase for WiFi can be obtained from its Channel State Information (CSI) for off-the-shelf WiFi module -- Intel 5300 WiFi card. CSI can be obtained from the PHY layer of 802.11 protocol using the open-sourced CSI Toolbox \citep{Halperin2010Predictable8P}. Our framework allows for using the \emph{virutal antennas} to measure the signal phase difference.

\subsubsection*{\textbf{Synthetic Aperture Radar (SAR):}}
Using their mobility, robots can improve the quality of information obtained \citep{Twigg2019RoboticPA, trackIO}, however they lack the ability of controlling robot motion to optimize the measurement value (e.g. range or AOA) or capture additional information like signal multipath. Our previous work \emph{exploits mobility} and WiFi signals, both inherent to robotic platforms, to emulate a virtual antenna array along the displacement of the robot, a method akin to SAR~\citep{sar1998,Watts20162DA3,Barrie2004ThroughWallSA}, for accurately measuring the change in phase of an incoming signal. As these signals traverse between a transmitting robot and a receiving robot, they interact with the environment such as being absorbed, reflected, and scattered \citep{Goldsmith2005WirelessC} by objects in the environment. Therefore, the signals paths can reveal directional information about the robots. This proves useful when robots are separated by large distances and/or are visually occluded from each other. Indeed, exploiting this fact has led to many uses like positioning~\citep{ubicarse}, adaptive networking~\citep{Gil2015AdaptiveCI}, mapping~\citep{Wang2019ActiveRF}, and security~\citep{Gil2015Spoof-Resilient,icra2019,Gil2019ResilientMC}. These works use short linear or 2D circular displacements (e.g. turn-in-place) of the signal receiving robot obtained using IMU and require that the transmitting robot be stationary for a few seconds when the receiving robot collects data.

However, emulating a linear or circular antenna array leads to \emph{sensing ambiguities} where the receiving robot cannot uniquely identify a transmitting robot's direction. This is a well-know problem in antenna array theory~\citep[chapter 6]{manikas2004differential}. Antenna array research shows a possible solution of using 3D antenna arrays that are capable of resolving such ambiguities~\citep{Vu2013-3D, Vu2010PerformanceAO}. However, traditional arrays are bulky~\citep{Bjrnson2019MassiveMI} and emulating SAR with robot's arbitrary displacement in 3D has not been done before. Another open challenge not addressed by previous works is the possible mobility of the transmitting robot for the duration of SAR. The change in phase measurements for even small displacements (e.g 24cm) of the receiving robot are desirable for obtaining highly accurate AOA profiles \citep{Gil2015AdaptiveCI}. However, not accounting for the simultaneous motion of the transmitting robot results in considerable errors in these measurements. These factors are important motivators for developing the ability for robots to \emph{emulate 3D antenna arrays over their arbitrary paths as they traverse the environment while both ends (i.e transmitting and receiving robots) remain mobile at all times.}

\subsubsection*{\textbf{Displacement estimation:}}
Advances in vision-based sensing and optimization in the past couple of years have enabled a robot to obtain its local displacement estimates accurately. Monocular systems such as \citep{Qin2018VINSMonoAR}, utilize one camera and one low-cost inertial measurement unit as sensing components, achieving 0.21m ATE in a 500m real world trajectory. Pose Graph Optimization also plays an essential role in state-of-the-art SLAM tasks as a back-end optimizer, for example as with distributed mapping \citep{Choudhary2017DistributedMW} and rotation estimation \citep{PGOSurvey_Luca}. Energy-efficient accelerator for VIO in \citep{karaman_droneVIO,karaman_miniVIO} also enables the localization and mapping on miniaturized robots. Several other works~\citep{karaman_sparsetoDense,scaramuzza_eventCameras,scaramuzza_motionEstimation} provide a multitude of techniques ranging from specialized visual sensing capability to estimation of depth and motion in the view with monocular system.
\subsubsection*{\textbf{Antenna array theory:}}
Traditional antenna arrays that provide AOA information are often bulky and hence not suitable for small agile robot platforms~\citep{Bjrnson2019MassiveMI}. To this end, we consider a SAR inspired approach that uses robot motion and signal measurements over several communicated packets, to obtain a full AOA profile.
This paper builds closely off of antenna array theory in order to characterize AOA performance metrics resulting from 3D SAR applied arbitrary robot displacement in 3D space.
The Cramer Rao Bound (CRB) is a often utilized tool for the performance analysis of different antenna array geometries\citep{Gazzah2006CramerRaoBF,Mirkin1991CramerRaoBO}. Specifically, the CRB provides important lower bound estimates on the variance of the resulting angle-of-arrival estimates as a function of antenna array geometry~\citep{Moriya2012CRLB3D,Mohamed2019ASO}; in other words, CRB analysis \emph{reveals the relationship between the robot displacement and the richness of information attainable from processing received communication signals}. This key observation opens the possibility of thinking of SAR as an enabling technology for a new kind of \emph{sensing} for robotics. 

\medskip
Thus, in comparison to previous work done in SAR, our proposed system is a generalized formulation of a direction finding capability for robots that combines local displacement estimation methods with antenna array theory. It enables a robot to obtain relative AOA of a wireless signal transmitter (stationary or moving) by using arbitrary 3D motion. However, this requires a mathematical, algorithmic, and experimental framework for characterizing the impact of i) displacements in 3D on the \emph{informativeness} (or richness of information) derived from processing collected signals, ii) displacement estimation error on the resulting accuracy of the AOA estimates, and iii) mobility at \emph{both ends} meaning mobility of both the transmitting and receiving robots, in order to enable continuous data collection that is most compatible with general multi-robot tasks.  We address these objectives in the current paper.

\section{Background and Notation}
In this section we present important notational conventions used as well as essential background on wireless signals and Synthetic Aperture Radar (SAR) that we will use throughout.

\subsection{Coordinate system}\label{notation_coordinate_system} As our system returns angular information, we formulate our problem in a Spherical coordinate system. Fig.~\ref{fig:3DcoordinateSystem} shows a schematic representation of a robot's displacement over a time window and it's local spherical coordinate system. $\varphi$ and $\xi$ are the azimuth and elevation angles of the receiving robot $i$ at a given timestep. The AOA tuple ($\phi$, $\theta$) represents the transmitting robot's relative azimuth and elevation direction with respect to the initial position (at time $t_k$) of robot $i$. $\phi$ and $\varphi$ are measured counter-clockwise from the x-axis while $\theta$ and $\xi$ are measured clockwise from the z-axis. Robot $i$'s position at time $t$ denoted by $p_i(t)$, is represented as ($\rho_i(t)$,$\varphi_i(t)$,$\xi_i(t)$), $\rho$ being the distance from the center of the frame. In the case that the robot's displacement is obtained via on-board sensors or estimation techniques, the local positions are denoted by $\hat{p_i}(t)$. This position measurement is subject to error which we denote as $\kappa_{\rho}$, $\kappa_{\phi}$ and $\kappa_{\theta}$ for errors in $\rho$, $\varphi$, $\xi$ respectively. We denote estimated relative AOA (from our system) between robots $i$ and $j$ as $(\hat{\phi},\hat{\theta})$ and the true AOA (from motion capture system) as $(\phi_g,\theta_g)$. Our experimental evaluation uses the Cartesian coordinates for a robot position, represented as ($x_i(t)$, $y_i(t)$, $z_i(t)$).\footnote{A position ($\rho$,$\phi$,$\theta$) in Spherical coordinates has corresponding Cartesian coordinates ($x,y,z$) as ($\rho\sin\theta\cos\phi, \rho\sin\theta\sin\phi, \rho\cos\theta$).}
\subsection{Robot Displacement}\label{notation_robot_trajectory} 
\begin{figure}
    \centering
    \hspace{15pt}\textbf{Coordinate System}\\
  \includegraphics[width=8.0cm,height=7.5cm]{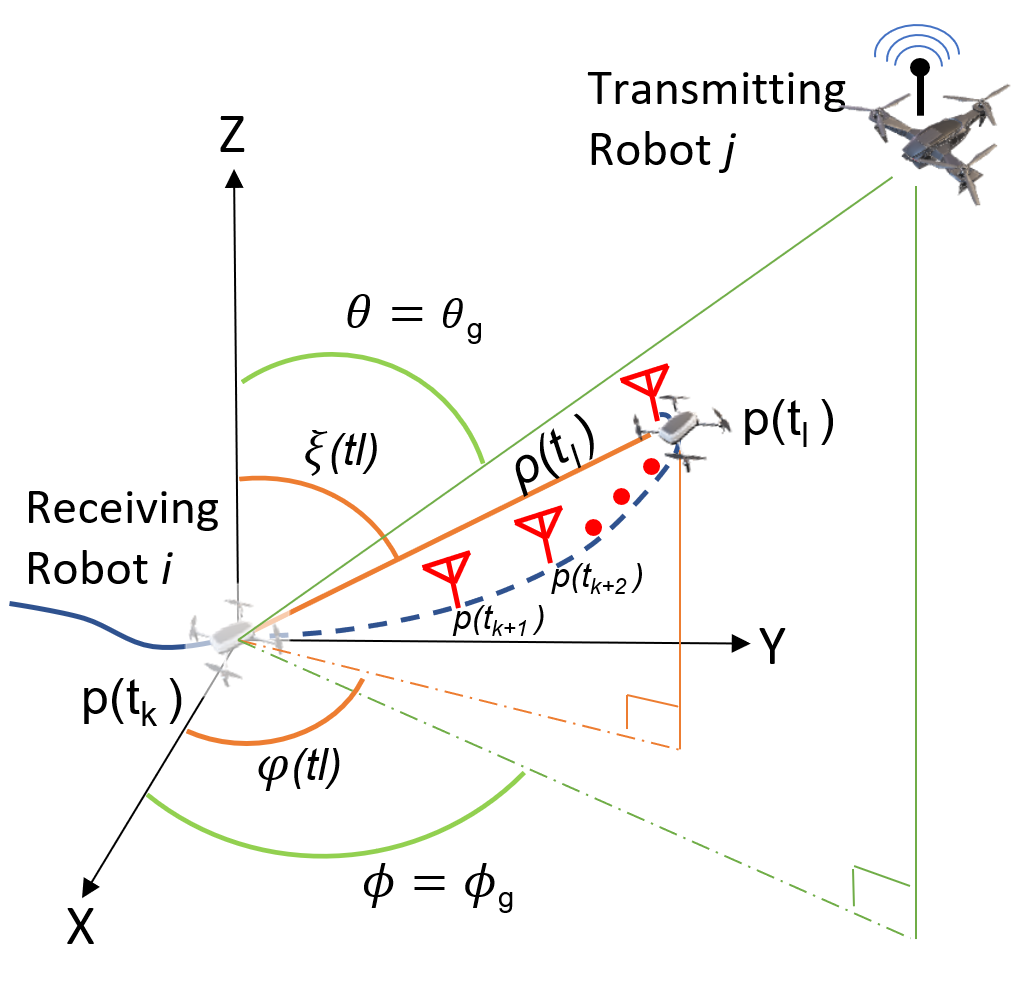}
  \caption{\footnotesize{\emph{Our formulation uses the Spherical coordinate system in receiving robot $i$'s local frame of reference. Robot $i$ uses M wireless signal measurements and their corresponding position estimates $p(t)$ = ($\rho$,$\varphi$,$\xi$) from time $t_{k}$ to $t_{l}$, l = k + M for each position (red antennas) to emulate M virtual antenna array elements and obtain relative angles $\phi$ (azimuth) and $\theta$ (elevation) to the signal transmitting robot $j$. Ground truth AOA ,obtained from motion capture, is denoted by $(\phi_g, \theta_g)$.}}}
  \label{fig:3DcoordinateSystem}
  \vspace{-0.1in}
\end{figure}

For our experiments, robot displacement is obtained via on-board sensors or from an external motion capture system (groundtruth). $\boldmath{\chi_i}\textbf{(t)}$ denotes the groundtruth displacement of the robot $i$ from time $t\in[t_k,\hdots,t_l]$ over which we intend to perform AOA estimation (Fig.~\ref{fig:3DcoordinateSystem}). Estimated displacement is denoted as $\boldmath{\hat{\chi_i}}\textbf{(t)}$. We assume that robot displacements are at least $2\lambda$, similar to the required minimum array length for attaining full AOA profile as stated in antenna array theory. For $5$ GHz WiFi, it is on the order of 12cm of total robot displacement. 

$\boldmath{\chi_i}\textbf{(t)}$ is the vector of all poses $\chi_i(t) = (R_i(t), p_i(t))$ where $R_i(t) \in SO(3)$ is orientation and $p_i(t) \in \mathbb{R}^3$ is position, for all $t\in[t_k,\hdots,t_l]$. $k$ is any timestep when the robot wants to initiate data collection for 3D SAR; i.e. to obtain relative AOA to any transmitting robots $j\in\mathcal{N}_i$ in its \emph{neighborhood}. $\mathcal{N}_i$ includes all transmitting robots $j$ for which robot $i$ can at least obtain ping packets (i.e. broadcasting at $5$ KB/s over 3 seconds). Note that signal packets from all $j\in\mathcal{N}_i$ simultaneously collected over robot displacement $\boldmath{\chi_i}\textbf{(t)}$, can be used to compute individual AOA profiles for \emph{all} robots in parallel. An estimated pose of robot $i$ in $\boldmath{\hat{\chi_i}}\textbf{(t)}$ at time $t_{k}$ is represented as $\hat{\chi_i}(t_{k}) = (R_i(t_k)R_{\epsilon}(t_k), p_i(t_k)+p_{\epsilon}(t_k))$. Let $R_{\epsilon}(t)$ and $p_{\epsilon}(t)$ denote the accumulated error in estimated position and orientation from time $t_0$ (The very first timestep when robot start to move) to any time $t_k$.

We use $\hat{p_i}(t)$ to denote robot $i$'s estimated position for a pose in $\boldmath{\hat{\chi_i}}\textbf{(t)}$ i.e $\hat{p_i}(t)$ = $p_i(t_k)+p_{\epsilon}(t_k)$. As our system can handle the motion of transmitting robots, let $\boldmath{\chi_{ij}}{(t_{k:l})}$ denote the relative displacement of robot $i$ with respect to another robot $j\in\mathcal{N}_i$ over the time window from $t=t_k$ to $t=t_l$, calculated as $\boldmath{\chi_{ij}}{(t_{k:l})}$ = $\boldmath{\chi_i}{(t_{k:l})}$ - $\boldmath{\chi_{j}}{(t_{k:l})}$. For brevity we denote this \emph{relative displacement} over the time vector $[t_k,\hdots,t_l]$ as $\boldmath{\chi_{ij}}\textbf{(t)}$ in the sequel. In Sec. \ref{sec:moving_ends} we show its validity even when the coordinate frames of robot are different. 

\subsection{A primer on Synthetic Aperture Radar (SAR)}\label{notation_SAR} An application of SAR over robot motion requires that a robot travel over some small local displacement of at least $2\lambda$, $\lambda$ being the signal wavelength while taking several measurements $h_{ij}(t)$ of the wireless channel broadcasted by its neighboring robots $j\in\mathcal{N}_i$. Following the far-distance assumption for planar waves, the ideal channel at time $t$ is modeled as~\citep{Tse2005FundamentalsOW} :

\vspace{-0.1in}
\begin{align}\label{eqn:channel_basic_eqn}
    h_{ij}(t) = \frac{1}{d_{ij}(t)}e^{(\frac{-2\pi\sqrt{-1}}{\lambda}d_{ij}(t))}
\end{align}
\noindent where $d_{ij}(t)$ is the groundtruth distance between robots $i$ and $j$ at time $t$. During its motion, the robot $i$ creates a history of packet-position tuples $<h_{ij}(t),p_i(t)>$, where each such tuple is essentially an emulation of omnidirectional ``antenna element'' in a virtual multi-antenna array. For example, thousands of lightweight WiFi ping packets can be transmitted at 5kB/s which is much lower than the 802.11 protocol's lowest rate of 6Mb/s. Thus a large number of antennas can be emulated with very little communication overhead\footnote{The total communication overhead incurred by our system is around 15 Kb for an AOA measurement.}. Thus using its local displacement and transmitted signals from robot $j\in\mathcal{N}_i$, the wireless channel at robot $i$ can be modeled using the multi-antenna array output observation model from~\citep{Vu2010PerformanceAO} as : 

\vspace{-0.1in}
\begin{equation}\label{eq:array_model}
    \mathbf{Y_{ij}(t)}=[Y(t_k),\hdots,Y(t_l)]^T=\mathbf{a}(\theta,\phi)\mathbf{(t)}\mathbf{h_{ij}(t)}+\mathbf{n(t)}
\end{equation}

\noindent where $Y(t_k)$ is received signal at robot position $p_i(t_k)$, the total number of signal packets received is $M$, and $l=k+M$. The vector $\textbf{a}(\theta,\phi)\textbf{(t)}$ is the \emph{steering vector}, which is defined by the robot's local displacement $\boldmath{\chi_i}\textbf{(t)}$ and thus dictates the \emph{geometry} of arbitrary 3D antenna array. This steering vector is a function of the geometry of the antenna array (i.e. robot $i$'s positions $p_i(t)$, $t\in[t_k,t_l]$ ) and the angle tuple ($\phi$, $\theta$) for transmitting robot $j\in\mathcal{N}_i$ (See Fig. \ref{fig:3DcoordinateSystem}) :
\begin{align}\label{eqn:steering_vector_eqn_basic}
    \textbf{a}(\theta,\phi)\textbf{(t)} = e^{(\frac{2\pi\sqrt{-1}}{\lambda}f(p_i({t}), \phi, \theta))}
\end{align}

$\textbf{n({t)}}$ is a signal noise vector, assumed to be Gaussian, circular, independent and identically distributed, with a mean of zero and a covariance of $\sigma^2I$, where $I$ is the identity matrix of appropriate dimension. The wireless channel measurement at time $t$, $h_{ij}(t)$ is a complex number capturing the phase and magnitude of the received signal. In case of 5 $\texttt{GHz}$ WiFi, $h_{ij}(t)$ can be obtained using the CSI Toolbox \citep{Halperin_csitool} for the Intel 5300 WiFi card. $\mathbf{h_{ij}(t)}$ = $[h_{ij}(t_k),\hdots,h_{ij}(t_l)]$ can thus be known at the receiving robot. 

Using direction finding algorithms like Bartlett or MUSIC from antenna array theory~\citep{Krim1996TwoDO}, the AOA profile ${F_{ij}(\phi,\theta)}$ can be obtained by measuring the signal phase difference of each array element with respect to the first element at $p(t_k)$. We use the Bartlett estimator which is give as:

\vspace{-0.1in}
\begin{equation}
    \label{eqn:bartlett_estimator}
    F_{ij}(\phi,\theta)=
    {\left|\sum_{t=t_k}^{t_l} {h_{ij}(t)}~a(\theta,\phi)(t) \right|^2}
\end{equation}

\noindent ${F_{ij}(\phi,\theta)}$ thus refers to relative paths a wireless signal traverses between a given pair of signal transmitting robot $j$ and receiving robot $i$. Mathematically, it is a 2D matrix calculated for all possible incoming signal directions along azimuth and elevation (360 x 180). 
Henceforth we refer to $AOA_{max}$ as the strongest signal direction $(\hat{\phi},\hat{\theta})$, or \emph{ the AOA peak corresponding to maximum magnitude path}, in the full AOA profile $F_{ij}(\phi,\theta)$. 

\medskip
\noindent A reference to all notations and terminology can be found in Appendix section \ref{sec:notations}. Next we formulate the problems that are addressed in this paper.

\vspace{-5pt}
\section{Problem Formulation}\label{sec:Problem Formulation}
This paper aims to enable robots in a team to infer the Angle-of-Arrival (AOA) profile $F_{ij}(\phi,\theta)$ to others in its neighborhood, based on received communication signals between them. In doing so, we aim to develop an \emph{``inter-robot sensor''} for robots relative information about each other by analyzing existing \emph{communication packets} in the network. We choose WiFi as the communication signal since it is ubiquitous to robot platforms and also allows us to empirically test our framework using off-the-shelf components. However our analysis is application to any RF signal and our algorithm can be deployed on actual hardware as long as the signal phase is accessible. Of particular interest is finding the maximum magnitude path $AOA_{max}$, between the robots referred to as the \emph{azimuth angle $\phi$} and \emph{elevation angle $\theta$} to a neighboring transmitting robot (see Fig.~\ref{fig:3DcoordinateSystem})\footnote{We note that in some cases the strongest signal path may not be the direct path which might be attenuated due to absorption or reflection of signals. However in this case there exist many methods for inferring the direct path ~\citep{ubicarse,spotFi}. In this paper we do not address the problem of finding the direct path from $F_{ij}(\phi,\theta)$ and rather refer the reader to relevant references for the solution to this problem.}. Our system uses an approach akin to \emph{Synthetic Aperture Radar} (Sec.~\ref{notation_SAR}) for this purpose. Key challenges for applying SAR-based methods to heterogeneous robot systems that we aim to address in this paper are: 
\vspace{-3pt}
\begin{enumerate}
    \item \textbf{Mobility:} Developing a framework to support AOA profile generation where both the transmitting and receiving robots are simultaneously moving in 2D or 3D during data capture. This is a commonly encountered scenario in multi-robot teams that must be accounted for in order to allow for continuous AOA sensing amongst the robots.
    \item \textbf{Displacement geometry:} We characterize \emph{informativeness} i.e the impact of the \emph{ virtual antenna array's} shape, emulated during a robot's displacement, on $F_{ij}(\phi,\theta)$ and the accuracy of the $AOA_{max}$. This is not readily known for shapes generated by arbitrary displacements. We thus characterize the impact of \emph{displacement geometry} on the sensitivity of the resulting phase measurements and the resulting $F_{ij}(\phi,\theta)$.
    \item \textbf{Estimation Noise:} Analyzing the effect of displacement estimation error on $F_{ij}(\phi,\theta)$. It is well known that local  displacement estimates for a robot obtained using onboard sensors such as IMU or VIO, are subject to noise. Thus the impact of this estimation noise on the accuracy of the resulting $AOA_{max}$ estimation must be quantified.
\end{enumerate}

\subsection{Simultaneous robot motion}\label{sec:ps_moving_ends}
\begin{figure*}
\centering
  \includegraphics[width=15.3cm,height=6.5cm]{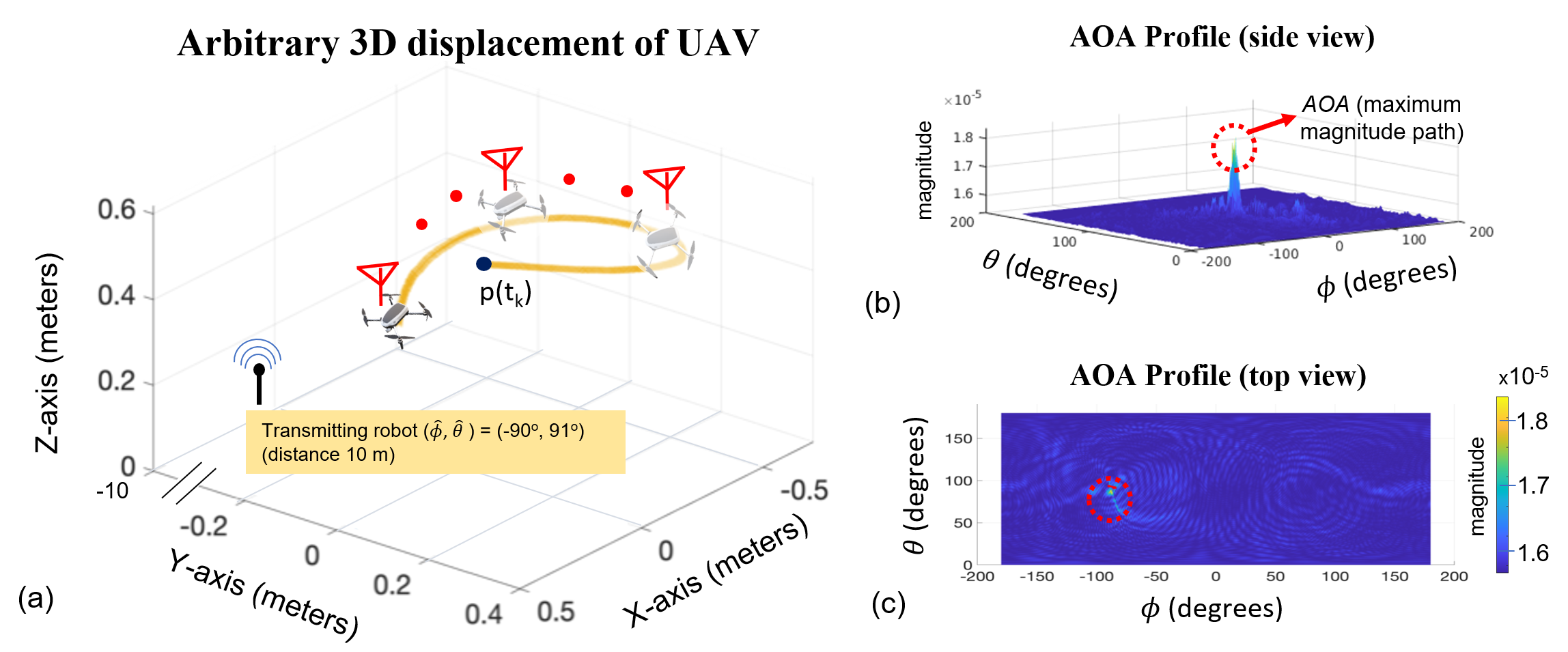}
  \caption{\footnotesize{\emph{(a) Shows the arbitrary 3D ground truth displacement for a signal receiving robot obtained from hardware experiments, the start position $p(t_k)$ indicated by a blue circle. (b) Corresponding AOA profile obtained using our formulation shows the \emph{AOA} (strongest signal direction) which corresponds to the relative direction of the transmitting robot (indicated by black antenna) located at true direction ($\phi_g$, $\theta_g$) = ( $\ang{-90}$, $\ang{91}$) and at a distance of 10 m in line-of-sight. Our system returns AOA ( $\hat{\phi}$, $\hat{\theta}$) = ( $\ang{-91.5}$, $\ang{92.5}$).}}}
  \label{fig:3D_SAR_AOA}
\end{figure*}

As many previous approaches that use SAR to obtain $AOA_{max}$ require transmitting robots to remain stationary during the period of data collection~\citep{Wang2019ActiveRF,PinIt,mostofiAOA}, they are not feasible for applications that require continuous mobility of all the robots. Additionally, they do not account for arbitrary motion of robots in 2D and 3D. This necessitates that our system account for motion of both signal transmitting and receiving robots and we refer to this as the \emph{moving ends} problem. 
\begin{problem}\underline{Moving ends problem:} 
Given the local displacements of the signal receiving robot $i$ and transmitting robot $j\in\mathcal{N}_i$, from time $t_k$ to $t_l$, obtain the AOA profile $F_{ij}(\theta,\phi)$ using the steering vector $\textbf{a}(\phi,\theta)\textbf{(t)}$ that characterizes robot $i$'s relative displacement $\boldmath{\chi_{ij}}\textbf{(t)}$ with respect to robot $j$.
\end{problem}
\noindent In Sec. \ref{sec:Development} we formulate the SAR based solution for obtaining $F_{ij}(\phi,\theta)$ and $AOA_{max}$, to a stationary transmitting robot, from arbitrary 3D motion of the receiving robot by applying the concepts from antenna array theory (Fig. \ref{fig:3D_SAR_AOA}). In Sec. \ref{sec:moving_ends} we use this formulation of 3D SAR to solve the \emph{moving ends} problem. 

\subsection{Information gain from displacement geometry}\label{sec:ps_trajectory_informativeness}
Our formulation can leverage arbitrary robot motion in 2D and 3D to emulate virtual antenna arrays of varying geometry. It is well known that an array's geometry affects its sensitivity to differentiate phases of an arriving signal from different source locations \citep{manikas2004differential}. For the case of interest in this paper where robots can operate in the full 3D space (not just 2.5D), sensitivity of the array and the ability to estimate $AOA_{max}$ consistently independent of source location, is of utmost importance. For the purposes of characterizing the sensitivity of a particular antenna array geometry, the Cramer Rao Bound (CRB) is often employed~\citep{Gazzah2006CramerRaoBF}. The Cramer Rao Bound is a lower bound on the variance of the $AOA_{max}$ estimation resulting from a particular antenna array geometry and is given by (dropping $t$ and subscript $i$,$j$ for brevity):
\begin{align}
    &CRB = 
\begin{bmatrix}
C_{\theta \theta} & C_{\theta \phi} \\
C_{\phi \theta} & C_{\phi \phi} \\
\end{bmatrix} \nonumber \\
&=\frac{\sigma^2}{2\mathbf{h}^H\mathbf{h}}{\underbrace{\begin{bmatrix}
Re(\frac{\partial a^H(\theta,\phi)}{\partial\theta} \frac{\partial a(\theta,\phi)}{\partial\theta}) &  Re(\frac{\partial a^H(\theta,\phi)}{\partial\theta} \frac{\partial a(\theta,\phi)}{\partial\phi}) \\
Re(\frac{\partial a^H(\theta,\phi)}{\partial\phi} \frac{\partial a(\theta,\phi)}{\partial\theta}) & Re(\frac{\partial a^H(\theta,\phi)}{\partial\phi} \frac{\partial a(\theta,\phi)}{\partial\phi})
\end{bmatrix}}_{FIM}}^{-1}
\label{eq:CRB_equation}
\end{align}

\noindent for a candidate source direction indicated by its azimuth angle $\theta$ and elevation angle $\phi$. Here $\mathbf{h}^H\mathbf{h}$ = $\norm{h}^2$ is signal magnitude, and $Re$ stands for real number.  $\sigma^2$ is the variance in noise of the wireless signal. FIM is the \emph{Fisher Information Matrix} which measures the amount of information that the geometry of the antenna array (captured via $\textbf{a}(\theta,\phi)\textbf{(t)}$ i.e the steering vector) provides on $F_{ij}(\phi,\theta)$. Since in our problem formulation, the geometry of the virtual antenna array is dictated by the displacement of the receiving robot, we define the \emph{informativeness} of such displacement geometry using the CRB as follows.

\begin{definition}\underline{\emph{Informativeness:}}
A displacement geometry is more \emph{informative} than another if it permits a smaller variance in $AOA_{max}$ as dictated by the Cramer Rao Bound (CRB).
\label{def:Informativeness}
\end{definition}

\noindent Thus, the second problem that we wish to tackle in this paper is the characterization of the informativeness for different robot trajectories; i.e. the effectiveness of our $AOA_{max}$ estimation capabilities as a function of the trajectory traversed by a robot as it is receiving communication packets from its neighbors.

\begin{problem} We wish to derive the formulation for the Cramer-Rao Bound (cf. Eqn.~\eqref{eq:CRB_equation}) in the $\theta$ and $\phi$ directions for arbitrary robot displacement in 3D, thereby characterizing the informativeness of different robot displacement geometries.
\end{problem}

\noindent Sec. \ref{sec:cr-bound} details the theoretical development our CRB analysis for three types of common displacement geometries for a robot: 3D helix, 2D planar circular and 2D linear. Our experimental analysis further shows that the insights garnered for these geometries generalize to overall trends in arbitrary 2D and 3D robot motion.

\subsection{Impact of displacement estimation error}\label{sec:ps_trajectory_impact}
Although our approach is independent of how robot displacement estimates are obtained, in order to make our system native to robotic platforms, our goal is to leverage local estimates from on-board sensors (e.g. VIO camera). Therefore, the third problem that we address in this paper is to characterize, both analytically and experimentally, the impact of the receiving robot's displacement estimation error on its generated $F_{ij}(\phi,\theta)$. In this paper, we use displacement estimates of the following two types to test our system's performance: 
\begin{itemize}
    \item \textit{\underline{Ground truth displacement}}: obtained using position estimates from motion capture system denoted by $p(t)$ at time $t$.
    \item \textit{\underline{Tracking camera displacement}}: obtained using position estimates from Intel Realsense Tracking Camera T265 (VIO camera). This out-of-the-box camera runs a visual SLAM algorithm which provides pose estimates at 200 Hz. We denote the estimated position at time $t$ as $\hat{p}(t)$.
\end{itemize}

\noindent Estimating displacement using the tracking camera introduces error in $F_{ij}(\phi,\theta)$ on account of noise. 
As ours is a relative direction finding system that uses spherical coordinates, we use an angular drift metric \citep{angulardrifting} to quantify this error for our theoretical analysis \footnote{For empirical evaluation, the corresponding $ATE_{trans}$ is calculated in Cartesian coordinates for simplicity.}.
\begin{definition}\underline{Angular drift:}
For every groundtruth position $p(t) = (\rho,\varphi,\xi)$ corresponding robot displacement, if we get an estimated position 
$\hat{p}(t)=((\rho-\kappa_{\rho},\varphi-\kappa_{\phi},\xi-\kappa_{\theta}))$, whenever $\kappa_{\phi}$, $\kappa_{\theta}$ are constant are non-zero constant or varying errors, we say that the displacement has an angular drift.
\label{def:angular-drift}
\end{definition}

\noindent For empirical evaluation we use the corresponding Absolute Trajectory Error in translation ($ATE_{\text{trans}}$)~\citep{Sturm2012ABF,Choudhary2017DistributedMW} given as follows in cartesian coordinates:

\begin{equation}
    \label{eq:ATEtran}
    ATE_{\text{trans}} = \left(\frac{1}{n} \sum_{t=t_{k}}^{t_{l}} ||p(t) - \hat{p}(t) ||^2\right)^{\frac{1}{2}}
\end{equation}

\noindent Since a robot's displacement is used to generate $F_ij(\phi,\theta)$ (cf. Eqn.~\ref{eqn:bartlett_estimator}), an error in displacement estimation will result in an error in $AOA_{max}$ calculation between two robots. We define this error to be \emph{AOA Estimation Error}.

\begin{definition}\underline{AOA error:}  We define error in AOA as $\phi_g - \hat{\phi}$ in azimuth and $\theta_g - \hat{\theta}$ in elevation, where ($\phi_g,\theta_g$) is the true AOA to the signal transmitting robot and ($\hat{\phi},\hat{\theta}$) is the estimated \emph{AOA} obtained by our system. 
\label{def:AoA-error}
\end{definition}
\noindent We use L2 norm between estimated $AOA_{max}$ and true AOA, as an error metric given by the equation:

\begin{equation}AOA_{Error} = \sqrt{(\phi_g - \hat{\phi})^{2} + (\theta_g-\hat{\theta})^{2}}
\label{eqn:AOA_euclidean_error}
\end{equation}

\noindent Thus we can now define our problem of characterizing the impact of displacement estimation error on AOA error as:

\begin{problem}\underline{Impact of displacement estimation error:}
For nonzero displacement estimation error, we wish to characterize the effect of angular drift  $\kappa_{\phi}>0$, $\kappa_{\theta}>0$ on the estimated $AOA_{max}$ and thus the $AOA_{Error}$ given in Eqn.~\eqref{eqn:AOA_euclidean_error}. 
\end{problem}

\noindent We note that in practice AOA error can arise from simultaneous impact of many factors, including insufficient informativeness of robot displacement geometry, artifacts from AOA estimation algorithms or errors in robot position estimation, channel noise or unaccounted phase shifts due to \emph{Carrier Frequency Offset} (CFO) or mobility of robot platforms (i.e. the \emph{moving ends} problem). We first solve for channel noise i.e., unaccounted phase shifts due CFO using previous work. Then, we independently analyze the effects of insufficient informativeness of robot displacement geometry and errors in robot local displacement estimation, on the AOA profile. Analyzing the impact of other estimators~\citep{Krim1996TwoDO} for the 3D SAR and moving ends formulations, is left for future work. 

\medskip
Finally, we aim to extensively validate our system's accuracy and performance on real robotic platforms in 2D and 3D space. In Sec. \ref{sec:hardware_application}, we evaluate the utility of our system for a dynamic rendezvous experiment between a pair of aerial and ground robot. We evaluate different displacement geometries generated during navigation in an unmapped environment in presence of \emph{moving ends} and visual occlusions in the environment.

\medskip 
\noindent We summarize key assumptions used throughout the paper:
\begin{assumption*}.
 
\begin{enumerate}
    \item Robots know the true \emph{North} and true \emph{Down} direction, perhaps by using on-board sensors like magnetometer and gyroscope/gravity sensor respectively.
    \item Each robot has the same global clock on millisecond level.
    \item All relative AOA measurements to any transmitting robot $j\in\mathcal{N}_i$ are taken with respect to the initial position $p_i(t_k)$ of the receiving robot $i$ in its local frame. 
\end{enumerate}
\end{assumption*}

\medskip
In the next section we derive our general formulation SAR. Following that, we show how our formulation can be extended to address the full mobility case, i.e. the \emph{moving ends problem}.

\section{SAR for Arbitrary Robot Motion}\label{sec:Development}
In this section, we show for the first time how a mobile receiving robot with arbitrary displacement in 3D can compute the AOA profile and thus obtain relative AOA to a stationary transmitting robot using our formulation.

Local pose estimation methods generally result in substantial orientation and position errors $R_{\epsilon}(t)$ and $p_{\epsilon}(t)$ when robots travel long distances without loop closures~\citep{drift_slam}. We use short local displacement $\boldmath{\hat{\chi_i}}\textbf{(t)}$ for receiving robot $i$ over $M$ robot positions $p_i(t_k),\hdots,p_i(t_l)$, $l=k+M$, in order to obtain the AOA profile $F_{ij}(\phi,\theta)$ (e.g. Fig.~\ref{fig:3D_SAR_AOA}) for \emph{arbitrary motion} in 2D and 3D. In practice, $M$ can be about $\sim$400 packets collected over a few seconds and with robot displacement being at least $\geq$ 2$\lambda$. This comes from the known result in antenna array theory stating that a full AOA profile is attainable over a minimum array length of $2\lambda$. We define the residual error $\epsilon(t)$ for any pose in $\boldmath{\hat{\chi_{i}}}\textbf{(t)}$ as follows: 
\begin{figure}
    \centering
    \hspace{15pt}\textbf{Cancelling CFO}\\
        \vspace{5pt}
  \includegraphics[width=7.0cm,height=4.0cm]{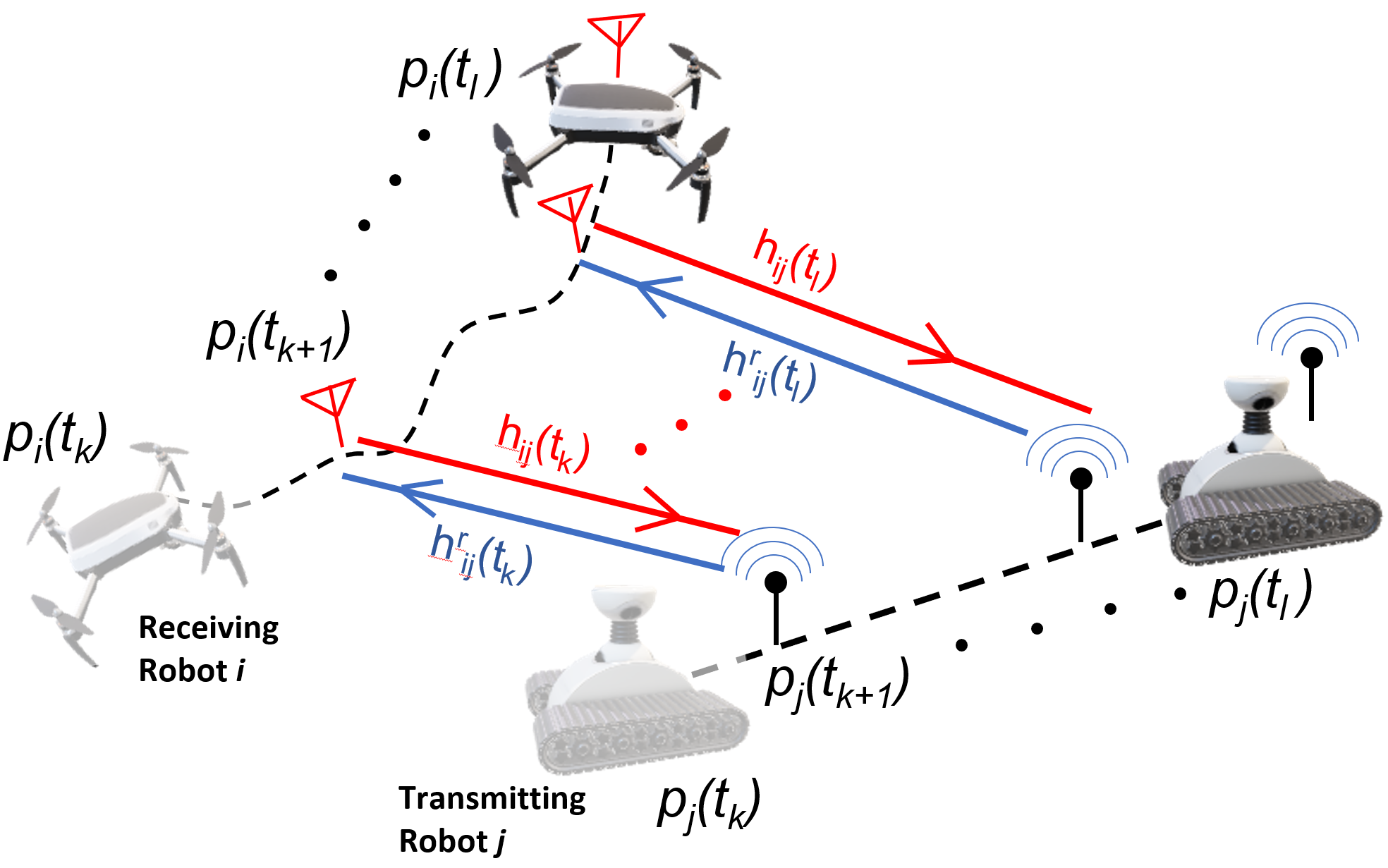}
  \caption{\footnotesize{\emph{Schematic representation of forward channel (from robot $i$ to robot $j$) and reverse channel (from robot $j$ to robot $i$). The product of these channels enables our system to cancel the Carrier Frequency Offset (CFO).}}}
  \label{fig:foward_reverse}
\end{figure}

\begin{definition}\underline{Residual Error:} Given error $p_{\epsilon}(t_k)$ in position $p(t_k)$ for the first pose $\hat{\chi_i(t_k)}$ in $\boldmath{\hat{\chi_i}}\textbf{(t)}$, for any time $t\in[t_k,\hdots,t_l]$, the residual error is given as $\epsilon(t)$ = $p_{\epsilon}(t)$ - $p_{\epsilon}(t_k)$.
\label{def:residual_error}
\end{definition}

\vspace{-0.1in}
As stated in our assumptions all measurements are  with respect to the initial position $p_i(t_k)$ of robot $i$ in its local frame. Thus, a position $p_i(t_u)$ at timestep $t_u$ becomes: 
\begin{align*}
p(t_{uk}) + \epsilon(t_{u}) = p(t_{u}) - p(t_{k}) + p_\epsilon(t_{u}) - p_\epsilon(t_{k}) 
\end{align*}
where $p_i(t_{uk})$ is position $p_i(t_{u})$ relative to the position at time $p_i(t_{k})$ and $\epsilon(t_{u})$ is the residual error. Note that our system is not impacted by orientation errors since it uses an omnidirectional antenna. In other words, our system is only impacted by the accumulation of residual errors (and displacement geometry as discussed in Sec. \ref{sec:Analysis_impact_factors}). For the rest of the paper, any position $p_i(t_{u})$ for a robot $i$ is implicitly assumed to be $p_i(t_{uk})$ for brevity.  

Given a perfect displacement estimation $\boldmath{\chi_i}\textbf{(t)}$, the error terms can be set to zero as in the case of using a motion capture system. We redefine the steering vector $\textbf{a}(\theta,\phi)\textbf{(t)}$ from Eqn. \ref{eqn:steering_vector_eqn_basic} in spherical coordinates as follows: (dropping the notation $t$ and subscript $i$ for brevity)
\begin{align}\label{eqn:steering_vector}
    \textbf{a}(\theta,\phi) = \begin{bmatrix} e^{(\frac{2\pi\rho_k\sqrt{-1}}{\lambda}\sin\theta \sin\xi_k cos(\phi-\varphi_k)+\cos\xi_k\cos\theta)}\\ \vdots \\  e^{(\frac{2\pi\rho_l\sqrt{-1}}{\lambda}\sin\theta \sin\xi_l cos(\phi-\varphi_l)+\cos\xi_l\cos\theta)}\end{bmatrix}
\end{align}

As $\mathbf{h_{ij}(t)}$ is a vector of signal measurements at each position in $\boldmath{\chi_i}\textbf{(t)}$, the steering vector enables calculating the phase difference between antenna positions in the virtual antenna array. Using this with AOA calculation algorithms like Bartlett we obtain $F_{ij}(\phi,\theta)$ for pair of communicating robots $i$,$j$. We note that the use of this steering vector formulation (Eqn.~\ref{eqn:steering_vector}) enables a robot to use its arbitrary 2D or 3D displacement to compute $F_{ij}(\phi,\theta)$. This is in contrast to formulations from previous work that only allow for strictly linear or curved 2D robot displacements~\citep{Gil2015AdaptiveCI,ubicarse}.

However, since the wireless signal transmitter and receiver are separate WiFi devices (i.e. different robots in a team), the signal oscillators in those devices have different frequencies with offset $\Delta_f$. This leads to time-varying phase offset in the signal phase called \emph{Carrier Frequency Offset} (CFO), leading to erroneous measurements. Following the development from previous work~\citep{Vasisht2016DecimeterLevelLW}, we use~\emph{channel reciprocity} to rectify this. Channel reciprocity states that i) ratio of forward and reverse channel is constant over time i.e $h_{ij}(t)/h^r_{ij}(t)=\kappa_h$, where $\kappa_h$ is constant because the forward and backward channel would be received almost simultaneously and ii) The frequency offset for the reverse channel is the negative of the offset for the forward channel i.e - $\Delta_f$. Hence, the observed forward and reverse channels, $\hat{h}_{ij}(t)$ and $\hat{h}^r_{ij}(t)$ respectively, that are affected by frequency offset $\Delta_f$ (and -$\Delta_f$), are given as follows for any position $p_i(t)$ during robot displacement:
\begin{align}
    \label{eqn:forward_backward_channel}
    &\hat{h}_{ij}(t) = h_{ij}(t)e^{-2\pi\Delta_f(t-t_k)} \\
    &\hat{h}^r_{ij}(t) = h_{ij}^r(t)e^{2\pi\Delta_f(t-t_k)}
\end{align}

\noindent The product of the forward and reverse channels can cancel out the phase offsets. Thus, rewriting Eqn. \ref{eqn:bartlett_estimator} and dropping $\kappa_h$ as it is a constant scaling factor, we get the AOA profile of transmitting robot $j$ for a receiving robot $i$ as follows:
\begin{equation}
    \label{eqn:bartlett_estimator_fin}
    F_{ij}(\phi,\theta)={\left|\sum_{t=t_k}^{t_l} \hat{h}_{ij}(t)~\hat{h_r}_{ij}(t)~\text{a}(\theta,\phi)(t)\right|^2}
\end{equation}

\noindent This formulation provides the solution to the problem of measuring $F_{ij}(\phi,\theta)$ over arbitrary robot displacement in 3D space for the stationary transmitting robot case. Thus by allowing robots to emulate an antenna array in full 3D space, we obtain maximum magnitude signal path $AOA_{max}$ to the neighboring robot. 

Next we address the more general problem of simultaneous mobility of both the transmitting and receiving robot, the \emph{moving ends problem}, in the subsequent section.

\section{Moving Ends Formulation}\label{sec:moving_ends}

\begin{figure}
    \centering
    \hspace{15pt}\textbf{Moving Ends}\\
    \includegraphics[width=7.7cm,height=6.25cm]{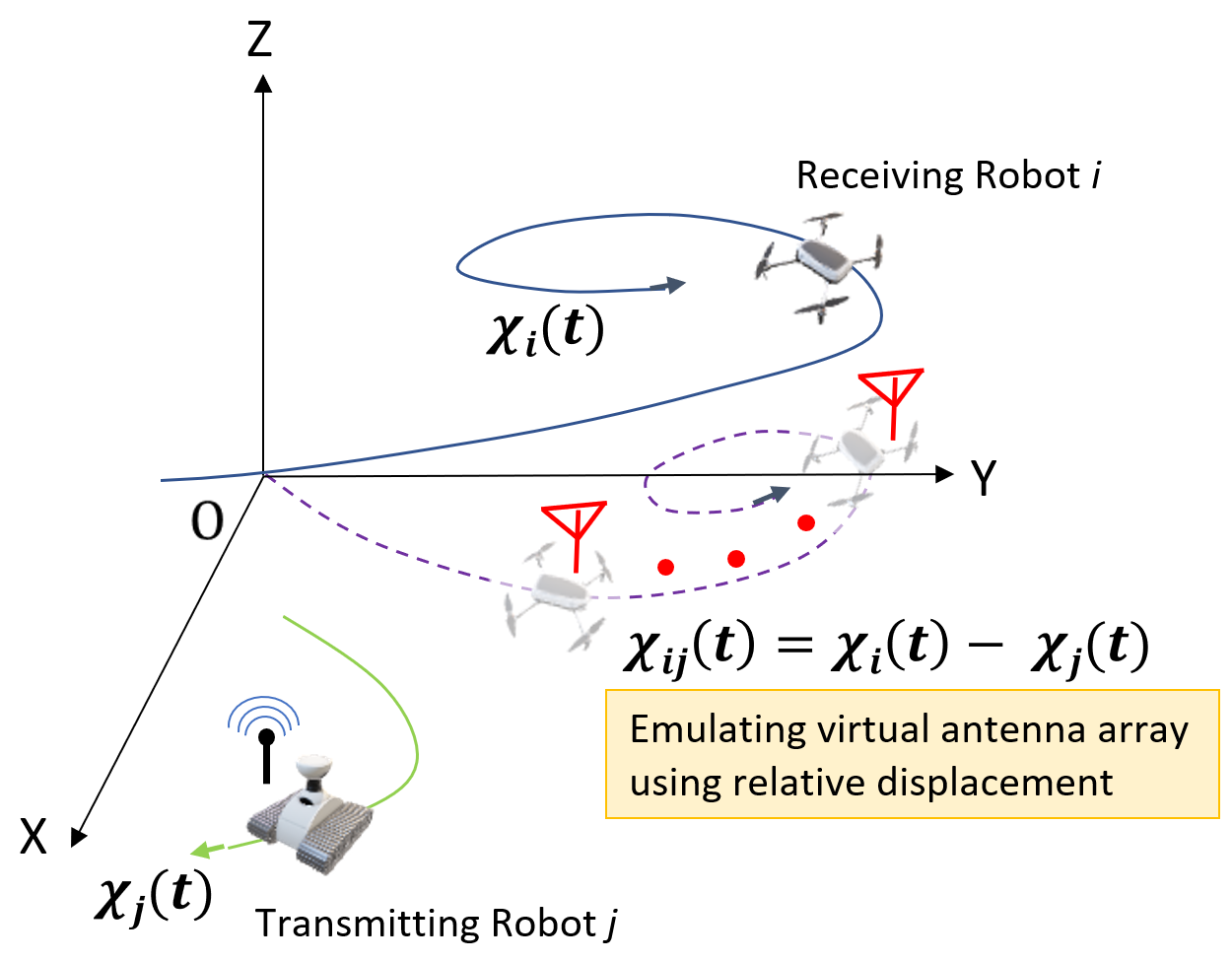}
	\caption{\footnotesize{\emph{Schematic showing relative displacement $\boldmath{\chi_{ij}}\textbf{(t)}$ of the UAV with respect to a moving transmitting robot $j$. $\boldmath{\chi_i}\textbf{(t)}$ and $\boldmath{\chi_j}\textbf{(t)}$ denote the actual displacement of the robot $i$ (emulating the virtual antenna array) and robot $j$.}}}
	\label{fig:Moving_ends_schematic}
\end{figure}

In this section we generalize our framework from Sec.~\ref{sec:Development} to accommodate the case where both receiving and transmitting robots are moving simultaneously during the data collection phase. This case is particularly challenging since not accounting for relative displacements between robots can greatly impact the resulting signal phase measurements -- leading to errors in the generated AOA profile $F_{ij}(\phi,\theta)$. Here, we show that a receiving robot $i$ can calculate $F_{ij}(\phi,\theta)$ to a moving transmitting robot $j$ by using the relative displacement obtained from its own and robot $j$'s local displacement estimates -- namely, by using $\boldmath{\chi_{ij}}\textbf{(t)}$ as the steering vector (Fig. \ref{fig:Moving_ends_schematic}). One clear difficulty in computing $\boldmath{\chi_{ij}}\textbf{(t)}$ is that robot $i$ and robot $j$ \emph{do not share a common coordinate reference frame} in the most general case. However, we begin the development of this section assuming the availability of a shared reference frame and then generalize to the case of different local coordinate frames between robots.

For the time being, let us assume that the robots have a common frame of reference (an assumption that we later relax). Thus, any position $p_{ij}(t)$ for pose $\chi_{ij}(t)$ for robot $i$'s relative displacement $\boldmath{\chi_{ij}}\textbf{(t)}$ for $t \in [t_k,\hdots,t_l]$ is given as:
\begin{equation}\label{eq:rho_1}
      p_{ij}(t) =  (p_i(t) - p_j(t)) - (p_i(t_k) - p_j(t_k))
\end{equation}
This is because all poses during local displacement $\boldmath{\chi}_i\textbf{(t)}$ are relative to the pose $\chi_i(t_k)$ for any robot $i$. Essentially, as our method uses displacement starting at time $t_k$ for any given robot, the relative poses obtained as per Eqn.~\eqref{eq:rho_1} can thus be used to obtain the steering vector. $F_{ij}(\phi,\theta)$ can then be calculated from the 3D SAR formulation developed in Sec.~\ref{sec:Development}.

In reality though, we do not have access to a global frame of reference and the local coordinate frames of the robots will have a rotation and translation drifting offset compared to global frame, thus making the computation of a relative displacement difficult to obtain. We introduce the following lemma which shows how a robot can obtain $F_{ij}(\phi,\theta)$ in the presence of these offsets. We first define the concept of a \emph{null vector} which captures the translation independent character of wireless signal phase measurements.

\begin{figure}
    \centering
    \hspace{15pt}\textbf{Null Vector}\\
	\includegraphics[width=8.0cm,height=6.5cm]{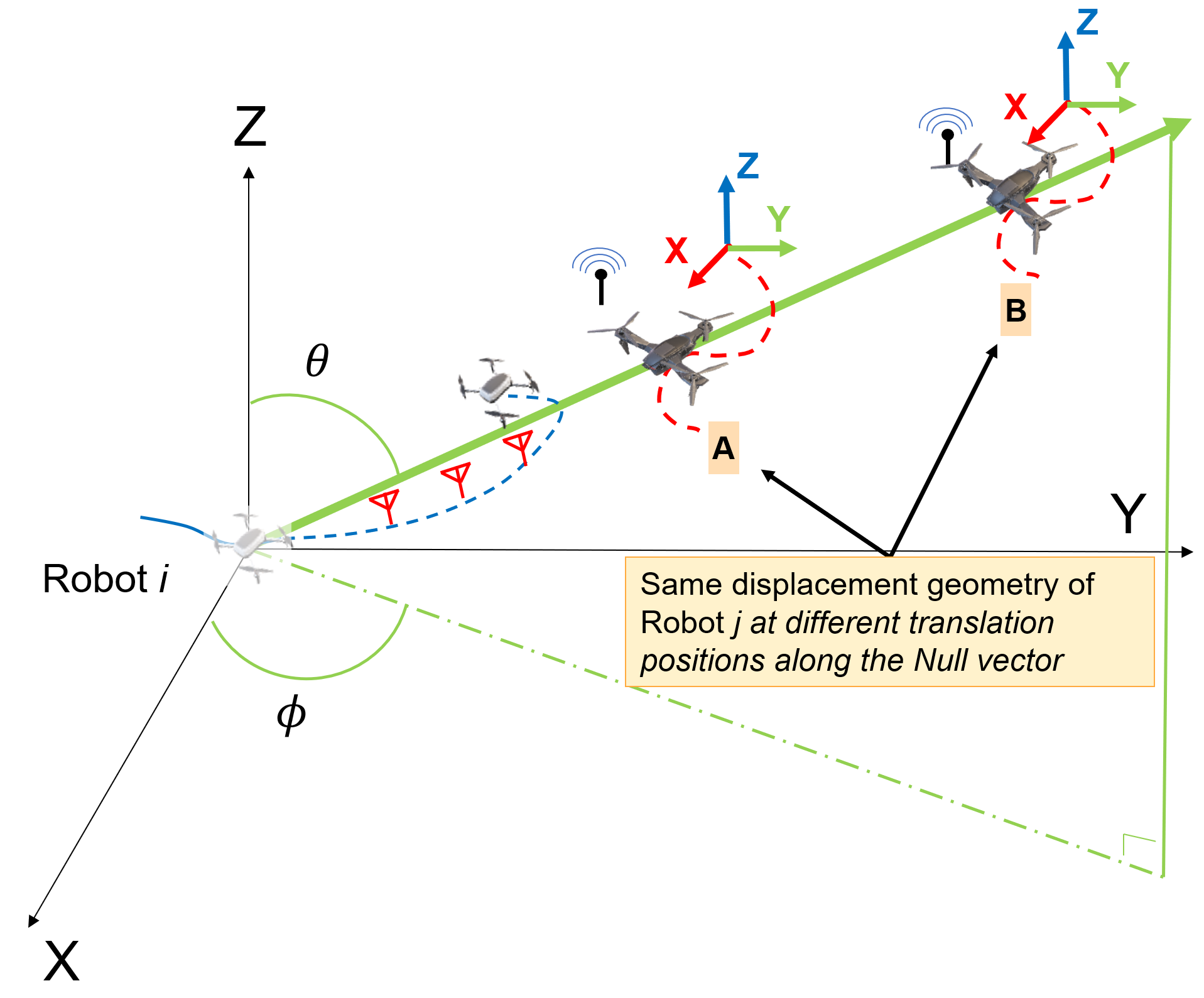}
	\caption{\footnotesize{\emph{Schematic representation of the Null vector (green) of robot $i$ (emulating the virtual antenna array) for Angle-of-Arrival ($\phi$,$\theta$). Rotation offset between any robot $j\in\mathcal{N}_i$ and robot $i$'s local coordinate frames can be removed by aligning the robots to true NORTH and DOWN direction, perhaps, using on-board sensors like magnetometer and gyroscope/gravity sensor. At any position on Null vector (e.g position A and B), robot $j$ will lead to same phase change in wireless signal, if the displacement geometry is same (dotted red line).}}}
	\label{fig:null_vector}
\end{figure}

\begin{figure}
    \centering
    \hspace{15pt}\textbf{AOA using relative displacement in 3D}\\
    \vspace{5pt}
    \includegraphics[width=8.0cm,height=7.5cm]{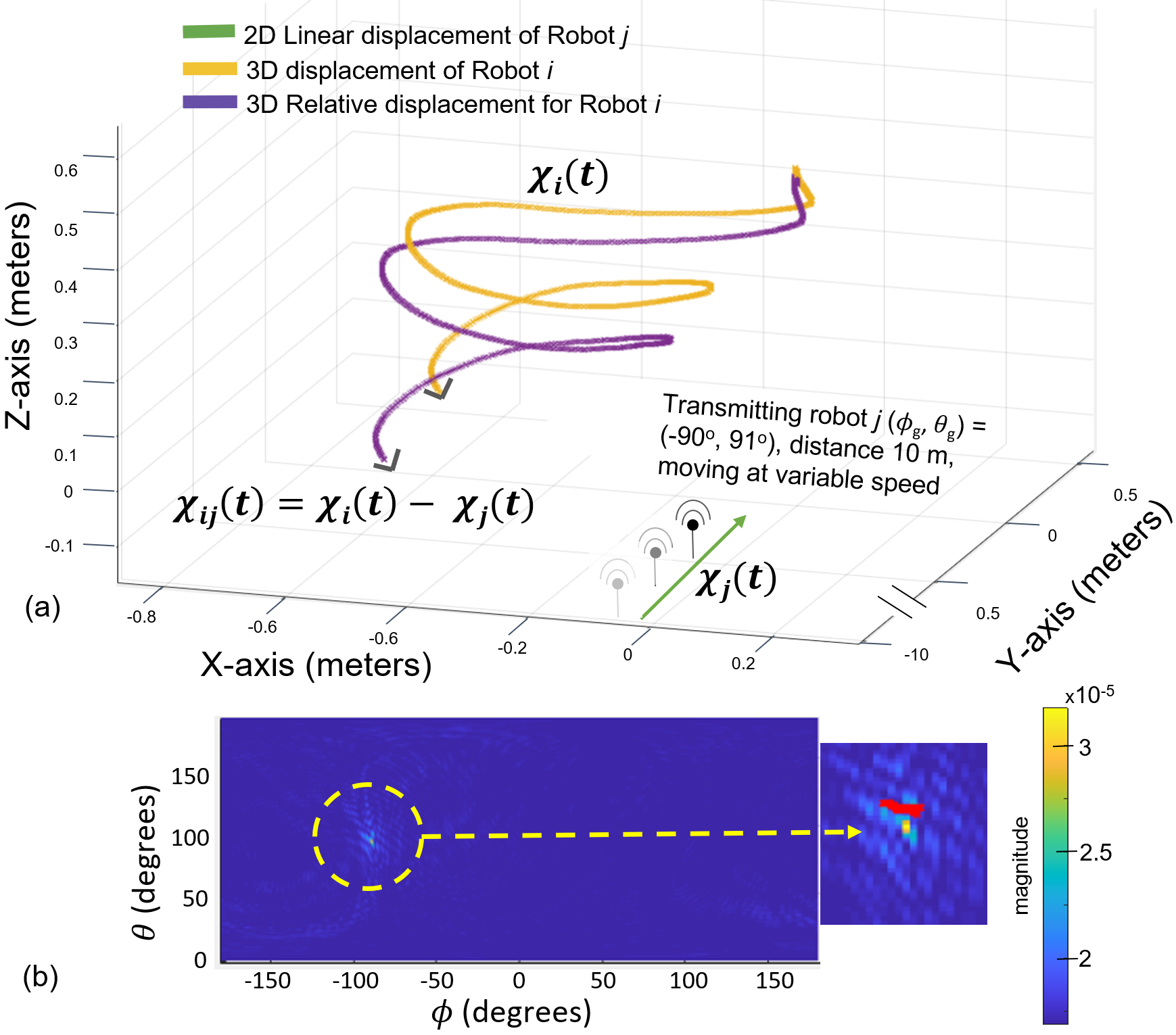}
	\caption{\footnotesize{\emph{(a) Hardware experiment showing the 3D displacement $\boldmath{\chi_i}\textbf{(t)}$ for the UAV and the corresponding relative displacement $\boldmath{\chi_{ij}}\textbf{(t)}$ due to displacement $\boldmath{\chi_j}\textbf{(t)}$ of transmitting robot $j$. Our approach for moving ends gives similar profile results for 3D motion as seen in Fig.~\ref{fig:3D_SAR_AOA}. (b) AOA profile showing true Angle-Of-Arrival ($\phi_g$, $\theta_g$) = (\ang{90}, \ang{91}), denoted by red 'x' in the profiles.}}}
	\label{fig:Relative_Traj_comparison_3D}
\end{figure}

\begin{definition}\underline{Null Vector:} We define the \emph{Null Vector} of a receiving robot $i$ as a vector along a specific angle-of-arrival from it's start position $p_i(t_k)$. A transmitting robot $j$ at any position on this \emph{Null Vector} and performing a specific motion will lead to \emph{same change in phase} of the wireless signal intercepted by robot $i$ (see Fig. \ref{fig:null_vector}).
\label{def:Null_vector}
\end{definition}

\noindent This is due to the fact that channel phase measurements are independent of the distance between a signal transmitter and receiver (for a detailed discussion, refer \cite{Ioannides2005UniformCA}). Thus, using the concept of a \emph{Null vector} we show that robots sharing a common North vector and gravity vector is sufficient to compute $\boldmath{\chi_{ij}}\textbf{(t)}$ as needed by our steering vector in Eqn.~\eqref{eqn:steering_vector}.

\begin{lemma}
\label{lemma:movingEnds}
Given displacement of two robots $i$ and $j\in\mathcal{N}_i$, each in their own local frame of reference with rotation offset $\tilde{R_i}$, $\tilde{R_j}$ with respect to a shared North vector, and a shared gravity vector and translation offset $\tilde{t}_{ij}$ with respect to each other, robot $i$ can calculate relative $AOA_{max}$ to robot $j$ by correcting for rotation offset and ignoring translation offset.
\end{lemma}

\noindent \textit{Proof: } As stated in Sec.~\ref{sec:Problem Formulation}, we assume that on-board sensors like magnetometer and gyroscope/gravity sensor can be used to rotate $\boldmath{\chi_i}\textbf{(t)}$ of a robot $i$ towards true \emph{North} and \emph{Down} by applying a rotation matrix, i.e $\boldmath{\tilde{\chi_i}\textbf{(t)}}$ = $\tilde{R_i}\hspace{-4pt}$ $\boldmath{\chi_i}\textbf{(t)}$. Similarly, any robot $j\in\mathcal{N}_i$ can align its coordinate frame such that $\boldmath{\tilde{\chi_j}\textbf{(t)}}$ =  $\tilde{R_j}\hspace{-4pt}$ $\boldmath{\chi_j}\textbf{(t)}$. Following def.~(\ref{def:Null_vector}), a robot $j$ along true AOA ($\phi$,$\theta$) relative to robot $i$ will give similar \emph{AOA} using our approach as long as it's anywhere on the \emph{Null Vector} and performing the same motion (Fig. \ref{fig:null_vector}). The relative displacement $\boldmath{\chi_{ij}\textbf{(t)}}$ thus becomes $\boldmath{\chi_{ij}\textbf{(t)}}$ = $\boldmath{\tilde{\chi_i}\textbf{(t)}}$ - $\boldmath{\tilde{\chi_j}\textbf{(t)}}$, since the presence of a translation offset $\tilde{t}_{ij}$ does not impact $AOA_{max}$ calculation and one can use Eqn.~(\ref{eq:rho_1}) to compute the steering vector.

\medskip
We validate our approach in hardware experiment  (Fig. \ref{fig:Relative_Traj_comparison_3D}), demonstrating our ability to compute $F_ij(\phi,\theta)$ for this moving ends problem in the Results Section of this paper (Sec.~\ref{sec:results}). As stated in our assumptions, the relative AOA is always computed with respect to position $p_i(t_k)$ of receiving robot $i$. Therefore we assume that robots are far enough away from each other such that local motions (which could be as small as $12~\text{cm}$ when $5~\text{GHz}$ WiFi is used) will not significantly change their true relative AOA.

In the next section we analyze the impact of different displacement geometries on $F_{ij}(\phi,\theta)$, i.e. characterizing \emph{informativeness} and the effect of errors in displacement estimation on $AOA_{max}$.

\section{Analyzing the Impact of Robot Displacement on AOA Estimation}\label{sec:Analysis_impact_factors}
We establish theoretical performance bounds of our system as they relate to the AOA profile, $F_{ij}(\phi,\theta)$, accuracy. We do so by characterizing the effect of robot displacement on the resulting accuracy of the obtained $AOA_{max}$ that corresponds to the strongest signal direction, which is the direct signal path in our case. The two major factors that we analyze are:
\begin{enumerate}
    \item \textbf{\emph{Displacement geometry (2D, 3D)}}:  To analyze the system performance due to variations in a robot's displacement geometry under ideal conditions, we formulate the Cramer-Rao bound for a \emph{3D helix} as well as the two special cases of \emph{Planar Curve} and \emph{Linear} displacement in Sec.~\ref{sec:cr-bound}. Our analysis shows that 3D displacement generally lead to lower variance (higher informativeness) in AOA estimation compared to 2D displacement.
    \item \textbf{\emph{Displacement estimation errors}}: We analyze the performance of our system in the presence of both constant and varying displacement estimation error. We derive the mathematical relation between displacement estimation error and AOA estimation error in Sec.~\ref{sec:characterization-trajectory-err-analysis}. 
\end{enumerate}
\noindent Although we analyze these factors separately, in reality they will act simultaneously to impact the accuracy of the resulting $F_{ij}(\phi,\theta)$ and $AOA_{max}$ estimation. However, our extensive empirical analyses from hardware experiments demonstrate good agreement with all analytical results presented in this section, despite the simultaneous impact of these factors.

\subsection{Cramer-Rao Bound for different displacement geometries}\label{sec:cr-bound}
Here, we consider the impact of the shape of the virtual antenna array, emulated during robot displacement, on \emph{AOA} estimation. Our main tool of analysis here will be the Cramer-Rao Bound (CRB) which provides a lower bound on the variance of the $AOA_{max}$ estimation for a given displacement geometry as captured by the steering vector $\textbf{a}(\phi,\theta)\textbf{(t)}$ (cf. Eqn.~\eqref{eqn:steering_vector}). This variance is inversely proportional to the informativeness of the robots' motion such that a lower variance indicates higher informativeness for a given displacement geometry (cf. Defn.~\ref{def:Informativeness}). 

We note that the CRB is a function of source location at a given $(\phi,\theta)$. In other words, the variance of the resulting $AOA_{max}$ estimate is a function of the different relative directions between the transmitting and receiving robots. Intuitively, depending on the shape of the virtual antenna array, the phase differences in the received signal from a specific direction will be easier or harder to discern. In this section we analyze this discerning capability of different displacement geometries via CRB. Specifically, we develop a closed-form expression for the CRB as a function of robot displacement, represented by  $\textbf{a}(\phi,\theta)\textbf{(t)}$, and a candidate relative direction of the transmitting robot i.e $(\phi,\theta)$.

We start by deriving the general form of CRB for arbitrary robot motion. Given that $\mathbf{h_{ij}(t)}$ is complex, deterministic, and known at the receiver robot, it is clear from Eqn. \ref{eq:array_model} that  output model \textbf{Y(t)} satisfies the \emph{Conditional observation model}; i.e. it is multivariate Gaussian with mean $\textbf{a}(\theta,\phi)\textbf{(t)}~\mathbf{h_{ij}(t)}$ and variance $\sigma^2I$~\citep{Vu2013-3D}. The CRB expression for an arbitrary 3D geometry array as given in Eqn. \ref{eq:CRB_equation} (dropping the notation $t$ and subscript $i,j$ for brevity):
\begin{align*}
&\hspace{-3.2in}\text{CRB} = \frac{\sigma^2}{2\mathbf{h}^H\mathbf{h}} \times \\
{\underbrace{
\begin{bmatrix}
\underbrace{Re(\frac{\partial \textbf{a}^H(\theta,\phi)}{\partial\theta} \frac{\partial \textbf{a}(\theta,\phi)}{\partial\theta})}_{\textbf{A}}  & 
\underbrace{Re(\frac{\partial \textbf{a}^H(\theta,\phi)}{\partial\theta} \frac{\partial \textbf{a}(\theta,\phi)}{\partial\phi})}_{\textbf{C}} \\
\underbrace{Re(\frac{\partial \textbf{a}^H(\theta,\phi)}{\partial\phi} \frac{\partial \textbf{a}(\theta,\phi)}{\partial\theta})}_{\textbf{C}} & \underbrace{Re(\frac{\partial \textbf{a}^H(\theta,\phi)}{\partial\phi} \frac{\partial \textbf{a}(\theta,\phi)}{\partial\phi})}_{\textbf{B}}
\end{bmatrix}
}_{FIM}}^{-1}
\end{align*}

It is also inversely proportional to the FIM as indicated above. Following the development in~\citep[Sec. 6]{Vu2010PerformanceAO}, derivatives of the $u^{th}$ element of the steering vector with $M$ total elements are given as:
\begin{align*}
    &\frac{\partial a_u(\theta,\phi)}{\partial \theta}= \\ 
    &\frac{2\pi \rho_u \Upsilon_u\sqrt{-1}}{\lambda}(\cos\theta \sin\xi_u\cos(\phi-\varphi_u)-\cos\xi_u\sin\theta)
\end{align*}
\begin{align*}
    \frac{\partial a_u(\theta,\phi)}{\partial \phi}=-\frac{2\pi \rho_u \Upsilon_u\sqrt{-1}}{\lambda}(\sin\theta \sin\xi_u\sin(\phi-\varphi_u)
\end{align*}
\noindent where 
\begin{align*}
    \Upsilon_u=e^{\frac{2\pi \rho_u\sqrt{-1}}{\lambda}(\sin\theta\sin\xi_u\cos(\phi-\varphi_u)+\cos\xi_u\cos\theta)}
\end{align*}

\smallskip
\noindent After some simplification the partial derivative terms in the FIM become:
\begin{align*}
&\hspace{-3.0in}\textbf{A} = \frac{\partial \textbf{a}^H(\theta,\phi)\partial \textbf{a}(\theta,\phi)}{\partial\theta \partial\theta}=\\
\sum_{u=1}^M \frac{4\pi^2 \rho_u ^2}{\lambda^2}(\cos\theta \sin\xi_u \cos(\phi - \varphi_u)-\cos\xi_u\sin\theta)^2
\end{align*}
and:
\begin{align*}
&\hspace{-2.5in}\textbf{B} = \frac{\partial \textbf{a}^H(\theta,\phi)\partial \textbf{a}(\theta,\phi)}{\partial\phi \partial\phi}= \\
\hspace{-.2in}\sum_{u=1}^M \frac{4\pi^2 \rho_u ^2}{\lambda^2}(\sin\theta \sin\xi_u \sin(\phi - \varphi_u))^2
\end{align*}
and finally:
\begin{align*}
&\hspace{-3.1in} \textbf{C} = \frac{\partial \textbf{a}^H(\theta,\phi)\partial \textbf{a}(\theta,\phi)}{\partial\phi \partial\theta}=\\
\hspace{.2in}\sum_{u=1}^M \frac{4\pi^2 \rho_u ^2}{\lambda^2}(\cos\theta \sin\xi_u \cos(\phi - \varphi_u)-\cos\xi_u\sin\theta)\\
\times (\sin\theta \sin\xi_u \sin(\phi - \varphi_u))
\end{align*}

\noindent Taking the determinant of the FIM, we arrive at the general form for the CRB in each the $\theta$ and $\phi$ directions:
\begin{align}
CRB_\theta = \frac{\sigma^2}{2\norm{h}^2Det(FIM)\textbf{A}} \nonumber \\
CRB_\phi =\frac{\sigma^2}{2\norm{h}^2Det(FIM)\textbf{B}}
\label{eq:CRB_helix_closed_form}
\end{align}
where $\sigma^2$ is the variance of our noise and \norm{h} is the magnitude of our signal, that is, $\frac{\sigma^2}{2\norm{h}^2}$ = $\frac{1}{\text{SNR}}$, where SNR is the signal to noise ratio.

\begin{figure}
    \centering
    \hspace{15pt}\textbf{Cramer-Rao Bound}\\
     \hspace{15pt}\textbf{for simulated robot motion}\\
    \vspace{3pt}
	\includegraphics[width=7.5cm,height=10.0cm]{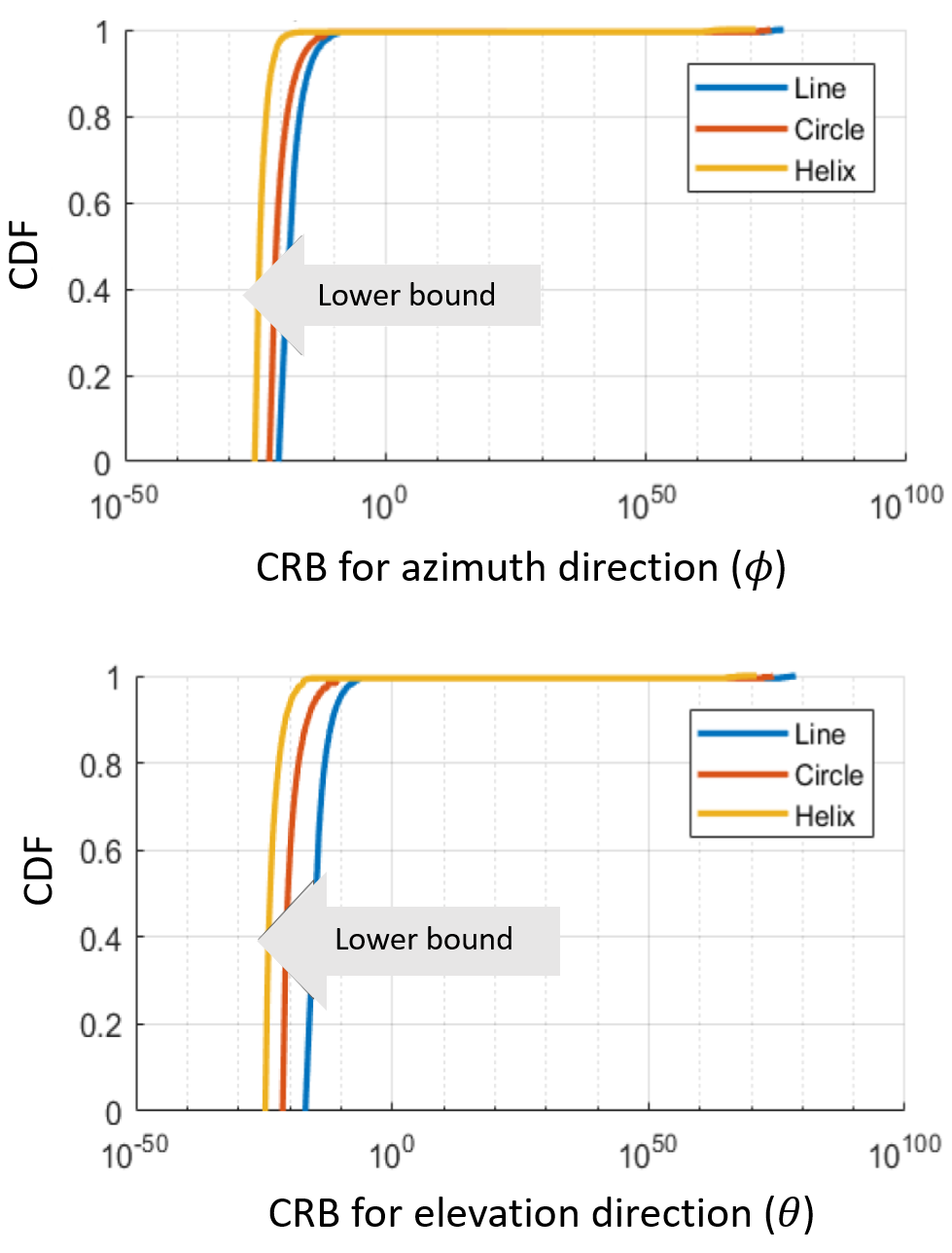}
	\caption{\footnotesize{\emph{CDF plot showing the effect of displacement geometry on Cramer-Rao Bound in the $\phi$ direction (top) and $\theta$ direction (bottom) for a receiving robot's simulated motion of fixed displacement, from 64800 simulated locations of transmitting robot that are uniformly distributed in 3D space. We see that 3D motion (helix) has a lower bound compared to 2D motion (circle and line). Cumulative result from hardware experiments (Fig.~\ref{fig:CRB_vs_AOA_error}) demonstrates that indeed the informativeness (inverse of the CRB) calculated for each motion successfully captures the improved performance in AOA estimation for displacement in 3D versus that in 2D.}}}
	\label{fig:cr-theta-phi}
	\vspace{-0.20in}
\end{figure}

\subsubsection{\textbf{Analysis of special geometries.}}
\noindent In this section, we include specific forms of the CRBs for three types of symmetrical displacement geometries that can be obtained on robot platforms - 2D linear, 2D curved and a 3D helix, to obtain in-depth insights on how their corresponding CRB values impact relative AOA. In section \ref{sec:results_informativeness} we show that these insights generalize well for arbitrarily shaped 2D and 3D displacement. Additional details of the derivations of CRB for these geometries can be found in the Appendix section.
\paragraph{\underline{(a) 3D spherical helix displacement.}} We begin by considering the case of displacement along a 3D spherical helix to build intuition on the impact of 3D motion on the CRB. For such motion, the spherical coordinates (dropping $t$ and subscript $i$ in the notations for brevity) are given by:
\begin{align*}
   \xi_u= \tau, \hspace{0.1in} \varphi_u= c\tau, \hspace{0.1in}
   \rho_u= r
    \end{align*}
for a spherical helix of radius $r$, spiral climb rate of $c$, and parameterization $\tau\in[0,2\pi)$. We assume that each antenna element $u$ is uniformly spaced along the helix so that $\tau=\{0,\Delta,2\Delta,\hdots,(M-1)\Delta\}$ where $\Delta=\frac{2\pi}{M}$. Further, we assume a climb rate  $c=1$ for a simplification of our closed form, though any constant can be similarly substituted before integration. So, the CRB terms for displacement along the 3D spherical helix become (derivation in Appendix \ref{sec:CRB_math} (I)):
\begin{align*}
    &\textbf{A}=-\frac{\pi ^2 r^2 \left(\cos ^2(\theta ) \cos (2 \phi )+\cos (2 \theta )-3\right)}{\lambda ^2}, \\
    &\textbf{B} = \frac{\pi ^3 r^2 \sin ^2(\theta ) (\cos (2 \phi )+2)}{\lambda ^2}, \\
    &\textbf{C}=\frac{2 \pi ^3 r^2 \sin (\theta ) \cos (\theta ) \sin (\phi ) \cos (\phi )}{\lambda ^2}.
\end{align*}
\noindent The Fisher Information for $\theta$ and $\phi$ are given by $\frac{2\norm{h}^2}{\sigma^2}$\textbf{A} and $\frac{2\norm{h}^2}{\sigma^2}$\textbf{B} respectively~\citep{Vu2013-3D}.
We calculate the determinant of FIM as follows:
\begin{align*}
&\textbf{Det}(\text{FIM}) = \\ &\frac{\pi ^5 r^4}{8 \lambda ^4} \sin ^2(\theta ) (8 \cos (2 (\theta -\phi ))+8 \cos (2 (\theta +\phi ))\\
& +\pi  \cos (2 (\theta +2 \phi ))+\cos (2 (\theta +2 \phi ))\\
&+(1+\pi ) \cos (2 (\theta -2 \phi ))\\
& -2 (\pi -9) \cos (2 \theta )-16 \cos (2 \phi )+2 \pi  \cos (4 \phi )\\
&+2 \cos (4 \phi )-2 \pi +50)
\end{align*}

\noindent Substituting \textbf{A}, \textbf{B}, and \textbf{Det}(FIM) into Eqn. \ref{eq:CRB_helix_closed_form} thus gives the corresponding closed form equation of the CRB for a spherical helix.
\begin{figure}
    \centering
    \hspace{20pt}\textbf{Informativeness of displacement geometries}\\
    \vspace{5pt}
    \includegraphics[width=8.5cm,height=6.0cm]{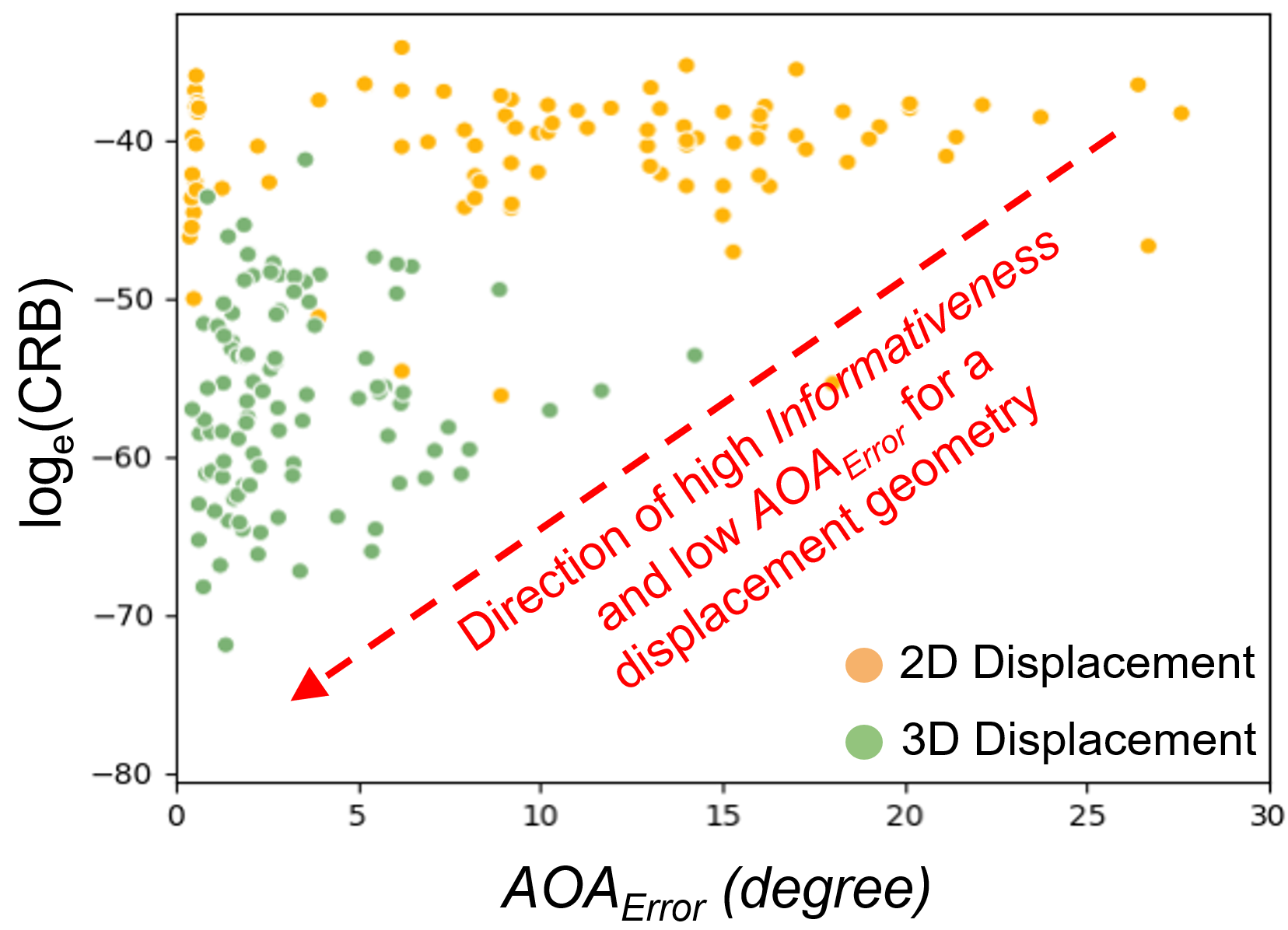}
    \caption{\footnotesize{\emph{ Aggregate results comparing 2D and 3D ground truth displacement geometries obtained from hardware experiments in terms of $AOA_{Error}$ (Eqn.~\ref{eqn:AOA_euclidean_error}) and Cramer-Rao Bound. A low value of CRB indicates that the corresponding displacement geometry has high informativeness. From the distribution of CRB and AOA error for 3D and 2D displacement, the former has a lower $AOA_{Error}$ on account of being more informative than the latter.}}}
    \label{fig:CRB_vs_AOA_error}
\end{figure}

\paragraph{\underline{(b) Planar circular displacement.}} For $z=0$ the helix is a circle of radius $r$ and the CRB equations should be the same as a circular planar curve.  For this case we have (dropping $t$ and subscript $i$ for brevity): 
\begin{align*}
    \xi_u= \tau,\hspace{0.1in} \rho_u=r,\hspace{0.1in} \varphi_u=\cos^{-1}(0)=\pi/2
\end{align*}

The CRB equations reduce to (derivation in Appendix \ref{sec:CRB_math} (II)):
\begin{align*}
    &\hspace{-1.25in}\textbf{A}= \frac{2\pi^2r^2}{\lambda^2}\cos^2\theta
\end{align*}
\begin{align*}
    &\hspace{-1.25in}\textbf{B}=\frac{2\pi^2r^2}{\lambda^2}\sin^2\theta
\end{align*}
\begin{align*}
    &\hspace{-1.25in}\textbf{C}=0.
\end{align*}

\noindent Substituting the values \textbf{A} and \textbf{B} in Eqn. \ref{eq:CRB_helix_closed_form} gives the corresponding closed form equation of the CRB for the planar circular motion.

\medskip
\paragraph{\underline{(c) Planar linear displacement.}} A line can be obtained by setting $r\rightarrow\infty$. For simplicity, we consider a line segment of length $2\pi$r at angle (a, b). This is parameterized in spherical coordinates as follows (dropping $t$ and subscript $i$ for brevity):
\begin{align*}
   \xi_u = a,\hspace{0.1in} \varphi_u= b,\hspace{0.1in} \rho_u= r\tau
\end{align*}

Thus, our CRB terms are:
\begin{align*}
&\hspace{-0.05in}\textbf{A}=\sum_{u=1}^M \frac{4\pi^2 (r(u-1)\Delta) ^2}{\lambda^2}\\
&(\cos\theta \sin(r(u-1)\Delta)\cos(\phi - b)-\cos(a)\sin\theta)^2
\end{align*}
\begin{align*}
&\hspace{0.04in}\textbf{B}= \sum_{u=1}^M \frac{4\pi^2 (r(u-1)\Delta) ^2}{\lambda^2}(\sin\theta \sin(a) \sin(\phi - b))^2
\end{align*}
\begin{align*}
&\hspace{-0.27in}\textbf{C}=\sum_{u=1}^M \frac{4\pi^2 (r(u-1)\Delta) ^2}{\lambda^2} \\ &*(\cos\theta \sin(a) \cos(\phi - b)-\cos(a)\sin\theta)\\
&\hspace{-0.1in}* (\sin\theta \sin(a) \sin(\phi - b))
\end{align*}

\noindent  We do not show a closed form for the CRB for a line, as the `u' term is not inside a trigonometric function and the CRB quickly inflates to infinity, except in certain directions where the trigonometric terms go to zero. That is, depending on the direction, linear displacements are largely uninformative.
\begin{figure}
        \centering
        \hspace{20pt}\textbf{Error in AOA for displacement geometries}\\
	    \vspace{5pt}
        \includegraphics[width=8.5cm,height=5.5cm]{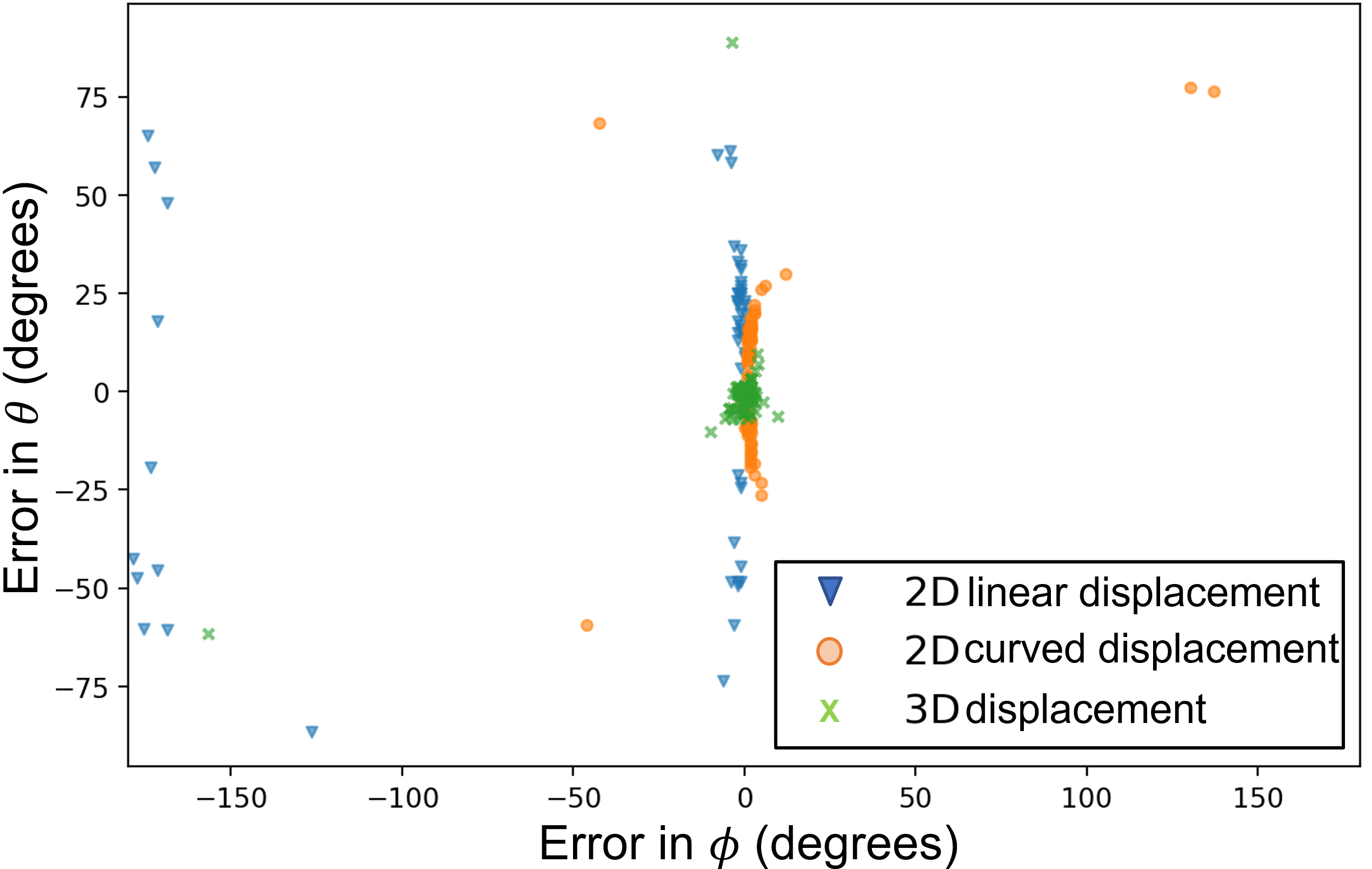}
        \caption{\footnotesize{\emph{Aggregate benchmark results for hardware experiments using ground truth displacement of the receiving robot that show AOA error in azimuth angle $\phi$ and elevation angle $\theta$ (def. \ref{def:AoA-error}). The AOA profiles are generated for 101 hardware experiment trials each for the three displacement geometries (2D linear, 2D curved and arbitrary 3D). 3D displacement (green) has higher AOA accuracy in both azimuth $(\phi)$. elevation $(\theta)$ directions. 2D Curved displacement shows high error in $\theta$. Moreover 2D linear displacement have additional errors in $\phi$ as well.}}}
        \label{fig:Different_mocap_Traj_comparison_AOA_error}
        \vspace{-0.1in}
\end{figure}
\subsubsection{\textbf{Observations:}}
We note a few observations from our analysis by comparing the 2D circle to the 3D helix. The off-diagonal terms of our CRB matrix (Eqn. \ref{eq:CRB_equation}), are always zero for the case of 2D circle. This results in a decoupling of the CRB in the $\theta$ and $\phi$ directions. When the transmitting robot is located in the x-y plane i.e $\theta=\frac{\pi}{2}$, $CRB_{\theta}$ for the planar circular case tends to infinity (as the \textbf{A} term becomes 0) while it remains finite for the 3D geometry case, thus the variance of estimated AOA for the 2D displacement increases (i.e. its informativeness decreases) as compared with the 3D displacement case for some relative angles between receiving and transmitting robots.
Further, we note the \textbf{B} term for a helix is always higher than for the planar circle by a factor of at least $\pi$, resulting in a lower bound for $\phi$ for the helix.
\begin{figure*}
	\centering
    \includegraphics[width=16.3cm,height=8.5cm]{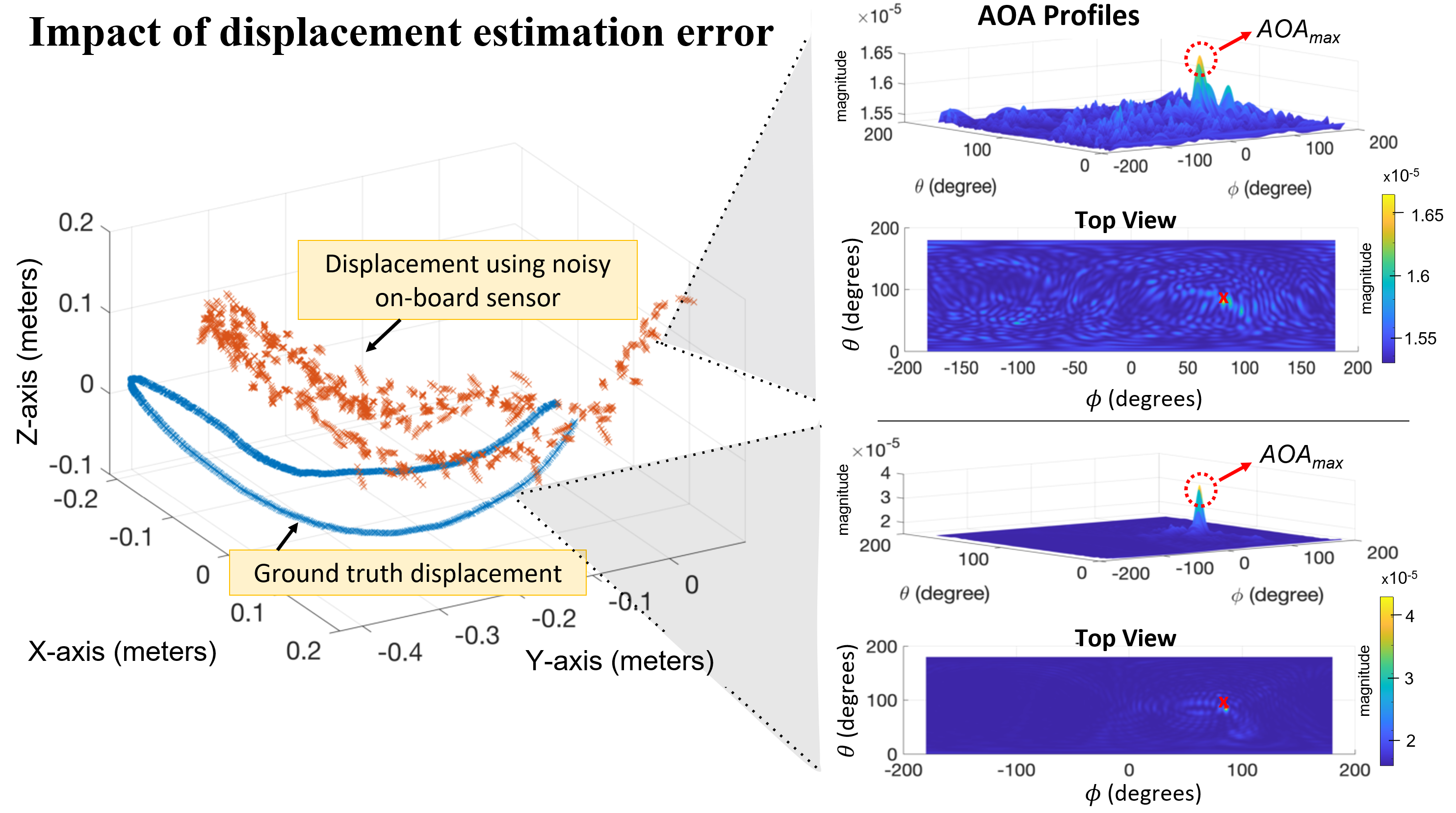}
	\caption{\footnotesize{\emph{Shows the AOA profile $F_{ij}(\phi,\theta)$, obtained during a hardware experiment, using ground truth displacement and estimated displacement (from Intel Realsense Tracking Camera T265). The peak $AOA_{max}$ in the in $F_{ij}(\phi,\theta)$ corresponds to the strongest signal direction of transmitting robot. Error in the estimated displacement increases the error in $AOA_{max}$ estimation and results in a more noisy profile (top right). The groundtruth AOA angles are marked by red 'x' in the Top View of the $F_{ij}(\phi,\theta)$. We see that there is a clear attenuation of the $AOA_{max}$ and a slight shift when using displacement from T265 camera compared to the peak obtained from ground truth displacement. Aggregate results comparing the AOA using displacement from ground truth and T265 camera are shown in Fig. \ref{fig:Mocap_vs_camera_AOA_cdf} and \ref{fig:Mocap_vs_camera_AOA}.}}}  
	\label{fig:Traj_comparison}
\end{figure*}

We also note some observations on the CRB from simulated motion of the receiving robot. This data was collected by simulating motion of a fixed displacement in each of the aforementioned shapes: a circle, a line, and a spherical helix. Then, the CRB (cf. Eqn. \ref{eq:CRB_helix_closed_form}) was calculated for a candidate transmitting robot direction along each angular position between [1, 360] degrees in $\phi$ and [1, 180] degrees in $\theta$, for a total of 64800 possible directions. The CDF of the CRBs across all directions Fig.~(\ref{fig:cr-theta-phi}) show that the lowest bound is achieved by the helix, followed by the circle and then the line, thus ranking these common displacement geometries in terms of informativeness.

Aggregate results in Fig.~(\ref{fig:CRB_vs_AOA_error}) for actual hardware experiments also demonstrate that the informativeness calculated for each displacement geometry successfully captures the behavior of $AOA_{Error}$ for 3D versus 2D displacement. The corresponding the AOA error along both $\phi$ and $\theta$ directions for 3D displacement geometries is lower compared to that for 2D linear and 2D curved displacement geometries (Fig.~(\ref{fig:Different_mocap_Traj_comparison_AOA_error})). Additional results for the hardware experiment pertaining to the impact of displacement geometry on AOA error are presented in Sec.~\ref{sec:results_informativeness}.

\subsection{Characterizing the impact of displacement estimation error on Angle-of-Arrival estimation}
\label{sec:characterization-trajectory-err-analysis}

As an implementation of SAR requires knowledge of local robot displacements (see Section~\ref{sec:Development}), one must rely on local estimation techniques for robots operating \emph{in the wild}. Most often, robot displacements can be estimated using onboard sensors such as inertial measurement units (IMU), cameras, or LiDAR. However the error in a local displacement $\boldmath{\hat{\chi_i}}\textbf{(t)}$ due to different estimation approaches translates into error in steering vector (incorrect antenna positions) when used for SAR. Thus, it is necessary to analyze our system's performance under cases where the measured displacement is different from ground truth displacement. 

Since, our method requires only \emph{local} displacement information, it is not impacted by large accumulations of errors over the entire robot trajectory. Specifically, to obtain an AOA profile, a robot uses its displacements over a small time window from $t=[{t_{k},\hdots,t_{l}}]$ to emulate a virtual antenna array (See Fig.~\ref{fig:3DcoordinateSystem}). 
This means that at any specific instance of 3D SAR, robot positions at timesteps not between $t_{k},\hdots,t_{l}$ are irrelevant for AOA estimation. Thus, our development here focuses on impact of $\boldmath{\hat{\chi_i}}\textbf{(t)}$ and the residual errors (def. \ref{def:residual_error}) accumulated over $\boldmath{\hat{\chi_i}}\textbf{(t)}$, on \emph{AOA}. Comparing the results for ground truth and displacement using on-board sensor during hardware experiments (Fig.~\ref{fig:Traj_comparison}) we see that higher error in the latter impacts the estimated AOA profile. Our goal in this section is to mathematically model and analyze this impact as a function of displacement estimation error. 
\begin{figure}
        \centering
        \hspace{15pt}\textbf{Comparison of robot displacements}\\
        \vspace{5pt}
        \includegraphics[width=8.0cm,height=5.25cm]{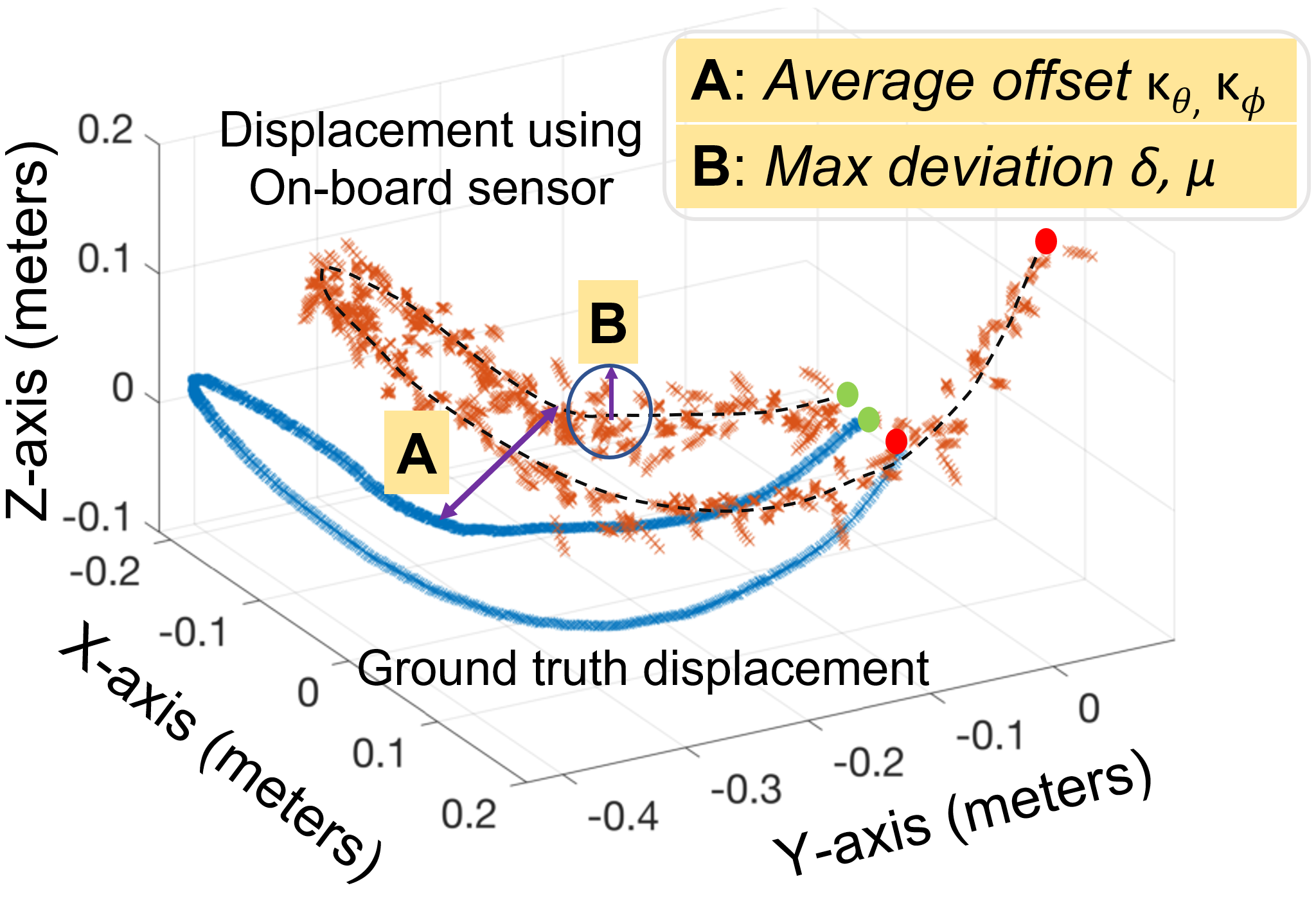}
    	\caption{\footnotesize{\emph{Hardware experiment showing the drift in estimated displacement, from Intel Realsense Tracking Camera T265, compared to ground truth displacement. The end positions points (red) are much further away from each other than the start points (green) indicating drift accumulation.}}}
    	\label{fig:drift_accumulation}
\end{figure}

We use the angular drift metric (def.~\ref{def:angular-drift}) to quantify the residual error in displacement estimation for our analysis since our system is based in Spherical coordinate system. We introduce two lemmas to show the impact of constant and varying angular drift on AOA estimation. To simplify our analysis, we use displacement along an ideal circular arc in 3D and show that results for arbitrary 3D displacement from hardware experiments closely follow our theoretical findings.

\subsubsection{Constant offset:}
We first analyze the case of a constant offset; for example, as would occur in the case of a bias or drift in the estimated displacement. We use a circular 3D motion of radius $R$ to simplify the analysis.
\begin{lemma}
If the estimated displacement in spherical coordinates is incorrect by a constant azimuth ($\phi$) offset $\kappa_{\phi}$ and a constant elevation ($\theta$) offset $\kappa_{\theta}$, the observed AOA peak will occur at a shifted angle of $\phi-\kappa_{\phi}$, $\theta-\kappa_{\theta}$.
\end{lemma}
\noindent \textit{Proof: } Assume that the transmitting robot $j$ at a point $(D,\phi,\theta)$ is very far from the receiving robot $i$ moving along the arc of a circle of radius $R$. The distance from robot $j$ to the first position of the arc is given by $D-R\cos\phi\cos\theta$. By rotating our coordinate system, we get $D-R\cos(\phi-\varphi(t_u))\cos(\theta-\xi(t_u))$ as the distance from robot $j$ to an arbitrary point on our arc, given robot $j$'s $(\phi,\theta)$ and virtual antenna element location $(\varphi(t_u),\xi(t_u))$ along robot $i$'s displacement. Thus, applying the general model of the wireless channel (Eqn. \ref{eqn:channel_basic_eqn}) for a circular motion based on distance $D$ and radius $R$ of the receiving robot's displacement, our forward-reverse channel product at a given ($\varphi(t_u),\xi(t_u)$) is the following (dropping notation $t$ for brevity):
\begin{align}\label{eqn:forward-reverse-sim}
    &h_{ij}(t_u) = \frac{1}{D^2}e^{\frac{-4\pi\sqrt{-1}~D}{\lambda}}*e^{\frac{-4\pi\sqrt{-1}~\rho~cos(\phi-\varphi_u)cos(\theta-\xi_u)}{\lambda}}
\end{align}
Due to some constant drift $\kappa_{\phi}, \kappa_{\theta}$, however, we instead measure a peak at some angle $\hat{\phi}$ rather than at $\phi$, and some angle angle $\hat{\theta}$ rather than at $\theta$, defined from the AOA profile (cf. Eqn.~\eqref{eqn:bartlett_estimator_fin}) as follows:
\begin{align}
\begin{split}
    &F_{ij}(\phi,\theta)= \frac{1}{D^2}\sum_{u=1}^{M}e^{\frac{-4\pi\sqrt{-1}~D}{\lambda}}* \\
    &\ \ \ \ \ \ \ \hspace{-10pt}e^{\frac{-4\pi\sqrt{-1}~\rho[cos(\phi-\varphi_u)cos(\theta-\xi_u)-cos(\hat{\phi}-\varphi_u+\kappa_{\phi})cos(\hat{\theta}-\xi_u+\kappa_{\theta})]}{\lambda}}
    \label{eqn:beta}
\end{split}
\end{align}
This takes the form of the well-known Bessel function, which is maximized at $\hat{\phi}=\phi-\kappa_{\phi}$ and $\hat{\theta}=\theta-\kappa_{\theta}$~\cite{circularAntennaArrays}.

\subsubsection{\textbf{Varying offset}}
Beyond constant offsets such as drift or bias, other errors that may complicate our analysis include the vibrations of drones and time-varying drifts in displacement across narrow data-capture windows. Our following analysis considers the effect of this varying offset (Fig. \ref{fig:Traj_comparison}).
\begin{lemma}
\label{lemma:trajError}
For an estimated robot displacement incorrect by some average offset $\kappa_{\phi}$ with a max deviation of $\delta$ around $\kappa_{\phi}$ in the azimuth direction, and some average offset $\kappa_{\theta}$ with a max deviation of $\mu$ around $\kappa_{\theta}$, there will exist a peak located at angle $(\phi-\kappa_{\phi}, \theta-\kappa_{\theta})$, in line-of-sight conditions. This peak will attenuate by a factor no more than $cos(\frac{\delta+\mu}{2})$, relative to the peak in absence of such offsets.
\end{lemma}

\noindent \textit{Proof: }  We consider the \emph{phase} component $\angle F(\phi, \theta)$ of Equation \ref{eqn:beta} (in the exponent) from the previous lemma, substitute a= $\phi-\varphi_u$ and b= $\theta-\xi_u$ and modify it for a varying $\kappa_{\phi}$ and $\kappa_{\theta},$ indicated by subtracting deviations from the mean for each position, $\delta_u, \mu_u$:
\begin{equation}
\begin{split}
    &\hspace{-0.1in}\angle\textit{F}(\phi, \theta)= \\ &\hspace{-0.1in}\frac{-4\rho\pi\sqrt{-1}}{\lambda}[cos(a)cos(b)-cos(a-\delta_u)cos(b-\mu_u)]
\end{split}
\end{equation}

\noindent Assuming $\delta_u$ and $\mu_u$ are relatively small, we can simplify our equation using trigonometric identities to the following:
\begin{equation}
\begin{split}
     &\hspace{-0.1in}\angle \textit{F}(\phi,\theta)=\\
     &\hspace{-0.1in}\frac{-4\rho\pi\sqrt{-1}}{\lambda}[\delta_u~sin(a)cos(b)+\mu_u~sin(b)cos(a)]
\end{split}
\end{equation}

\noindent Consider the worst-case scenario, where $\delta=\delta_{max}$ and $\mu=\mu_{max}$, and $\delta=-\delta_{max}$ and $\mu=-\mu_{max}$, alternately. For brevity, we represent the quantity $\frac{1}{D^2}e^{\frac{-4D\pi\sqrt{-1}}{\lambda}}$ with $\alpha$, and group these alternate phases together (thus changing the bounds of our summation). We can rewrite our profile as follows:
\begin{equation*}
    F_{ij}(\phi,\theta)\geq~\alpha\sum_{u=1}^{M/2}e^{\frac{\sqrt{-1}~(\delta_{max}+\mu_{max})}{2}}+e^{\frac{-\sqrt{-1}~(\delta_{max}-\mu_{max})}{2}}
\end{equation*}
\begin{equation}
    \text{i.e.}~F_{ij}(\phi,\theta)\geq~\alpha M cos{\frac{\delta_{max}+\mu_{max}}{2}}
\end{equation}
\noindent Thus, our peak at $\phi-\kappa_{\phi}$, $\theta-\kappa_{\theta}$ from Lemma 1 is attenuated by a factor that is at most $cos\frac{\delta_{max}+\mu_{max}}{2}$ vs. the peak in the absence of drift.

Our hardware experiment results bolster these theoretical findings. A representative experiment sample comparing the robot displacements obtained from motion capture versus an on-board sensor as well as the corresponding AOA profile, are show in Fig.~\ref{fig:Traj_comparison}. The side view for the profiles indicates an attenuation of the maximum AOA peak as expected from the analysis of this section. From a top view the predicted peak shift can also be seen for the noisy estimated displacement versus the case where ground truth displacement is used. 

We note that to directly apply these analytical results for robot motion in Spherical coordinate system to that in Cartesian coordinates, the corresponding $ATE_{trans}$ error from Eqn.~\ref{eq:ATEtran}, which is commonly used metric~\citep{Sturm2012ABF}, needs to be converted to angular error. However, results with $ATE_{trans}$ error demonstrate an increasing linear trend for the AOA error. Figure~\ref{fig:aoa_vs_ATE_error} demonstrates this for different values of simulated $ATE_{trans}$ error and corresponding AOA error in $\phi$ and $\theta$ for an actual robot displacement obtained from hardware experiments. Section~\ref{sec:results_traj_erro_AOA} also includes aggregate results showing the impact of error in displacement estimation on AOA error on average, where robot displacements are estimated using an Intel RealSense Tracking Camera T265.
\begin{figure}
    \centering
    \hspace{15pt}\textbf{AOA using relative displacement in 2D}\\
    \vspace{3pt}
    \includegraphics[width=8.5cm,height=7.5cm]{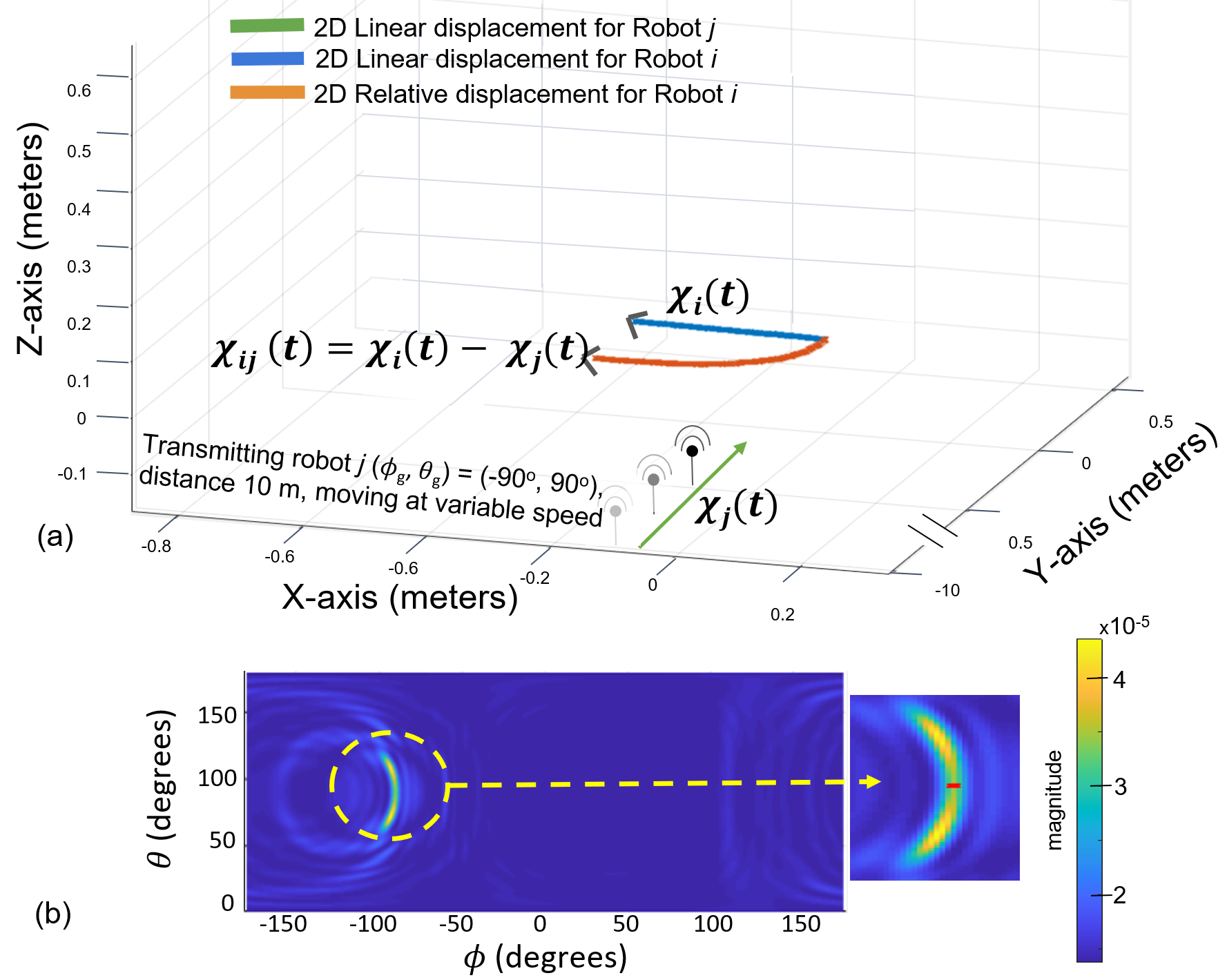}
	\caption{\footnotesize{\emph{(a) Hardware experiment showing linear 2D displacement $\boldmath{\chi_i}\textbf{(t)}$ for signal receiving robot $i$ and the corresponding curved relative displacement $\boldmath{\chi_{ij}}\textbf{(t)}$ due to the displacement $\boldmath{\chi_{j}}\textbf{(t)}$ of transmitting robot $j$. Using the relative displacement, our approach for moving ends (Sec. \ref{sec:moving_ends}) gives similar profile results for 2D curved displacement (Fig.~\ref{fig:3D_SAR_AOA}) with ambiguity only along elevation angle. (b) AOA profile showing ($\hat{\phi}$, $\hat{\theta}$) = (\ang{-90}, \ang{73}) while true AOA ($\phi_g$, $\theta_g$)  = (\ang{-90}, \ang{90}), denoted by red 'x' in the profiles. Relative displacement $\boldmath{\chi_{ij}}\textbf{(t)}$ (curved 2D) thus has a more informative geometry than actual displacement $\boldmath{\chi_i}\textbf{(t)}$ (linear 2D).}}}
	\label{fig:Relative_Traj_comparison_2D}
	\vspace{-5pt}
\end{figure}

\section{Results}\label{sec:results}
This section presents the results of extensive simulation and hardware experiments that demonstrate an agreement of practical implementation with our analytical results. As clarified in Sec.~\ref{sec:intro}, our \emph{WSR} framework can be deployed on actual hardware with any RF signal as long as the signal phase is accessible. For experimental evaluation, we use WiFi since it is ubiquitous to current robot platforms and its signal phase for off-the-shelf Intel 5300 WiFi card can be readily accessed by deploying the \emph{CSI Toolbox}~\citep{Halperin_csitool} on the robots' on-board computer. Using this WiFi card, we show 1) the feasibility of attaining accurate AOA profiles $F_{ij}(\phi,\theta)$ using relative displacement of a receiving robot with respect to the transmitting robot to solve the \emph{moving ends} problem, 2) the informativeness of different displacement geometries in 2D and 3D, and 3) the impact of displacement estimation errors on attainable $AOA_{max}$ accuracy. We find that:
\vspace{-0.1in}
\begin{itemize}
    \item Our developed framework from Sec.~\ref{sec:moving_ends} solves the \emph{moving ends} problem where $F_{ij}(\phi,\theta)$ is attainable even when both the transmitting and receiving robots are simultaneously moving. Furthermore, we observe that relative displacement geometries can have more informativeness, resulting in higher $AOA_{max}$ accuracy.
    \item Experiments using ground truth displacement show that arbitrary 3D displacement geometries generate highly accurate $F_{ij}(\phi,\theta)$ (less than $10\degree$ error in $AOA_{max}$ for $95\%$ of trials). In contrast, for planar 2D curved and linear displacement of a robot, only $50\%$ and $7\%$ of trials show error below $10\degree$ respectively.
    \item For robot displacement estimates using on-board VIO sensor, with an average of $7\%$ estimation error ($0.2$ m error in position over $2.8$ m distance), attainable $AOA_{max}$ median estimation accuracy was shown to be $7.58\degree$ in azimuth, $3.21\degree$ in elevation. 
\end{itemize}

\noindent We subsequently describe our experimental setup in section \ref{sec:result_exp_setup}. The results for moving ends experiments are detailed in section \ref{sec:results_movingEnds}. Section \ref{sec:results_informativeness} and \ref{sec:results_traj_erro_AOA} show the results for displacement geometry informativeness and the impact of displacement estimation respectively.

\subsection{Experimental setup} \label{sec:result_exp_setup}
\subsubsection*{\textbf{Testbed and hardware setup:}}We use an off-the-shelf Turtlebot3 Waffle as the ground robot and a NXP KIT-HGDRONEK66 drone as the aerial robot (Fig.~\ref{fig:hardware_robots}). Each robot is mounted with a lightweight 2 or 5 dBi omnidirectional WiFi antenna, UP Squared board with Intel 5300 WiFi module and a VIO sensor - Intel Realsense Tracking Camera T265 - which gives local pose estimates at 200 Hz. We evaluate our system in a $300$ m$^2$ indoor environment covered with the Optitrack motion capture system that enables obtaining ground truth measurements for robot displacement and AOA. We first evaluate our system using ground truth displacement to obtain AOA accuracy baseline. We then evaluate how $F_{ij}(\phi,\theta)$ and $AOA_{max}$ are impacted when noisy displacement estimation, obtained from on-board T265 VIO camera, are used instead of ground truth displacements.

For experiments with ground truth displacement, the center of the rigid body model, generated by the motion capture system, is aligned with the physical placement of the WiFi antenna on the robots' chassis. For experiments with the T265 VIO camera, the antenna and the camera are aligned on the robots' chassis, and their offset, if any, is accounted for. The robots use minimal communication (broadcasting ping packets) over the center frequency channel of the 802.11 $5$ GHz bands. Our system's total data transmission overhead amounts to a maximum of $15~\text{KB}$ in 3 seconds when a receiving robot performs 3D SAR. Note that the frequency of such transmission would highly depend on the application. For dynamic rendezvous demonstration using real hardware (Sec. \ref{sec:hardware_application}), as the receiving robot continuously requires $AOA_{max}$ for navigation, this leads to an average $5~\text{KB/s}$ transmission overhead. This includes \emph{CSI} of the transmitted signals between robots and local position estimates from neighboring robots (in their coordinate frame), collected on the on-board computer in real time. The current nature of our code-base requires processing this data on an off-board computer, but the computation requirements for our system are satisfied by the onboard computer as well. The AOA profile $F_{ij}(\phi,\theta)$ is generated using the Bartlett estimator (Eqn.~\ref{eqn:bartlett_estimator_fin}).
\begin{figure*}
	\centering
    \hspace{20pt}\textbf{\large{CDF of error in AOA for robot displacement geometries}}\\
    \vspace{5pt}
    \includegraphics[width=17.5cm,height=5.0cm]{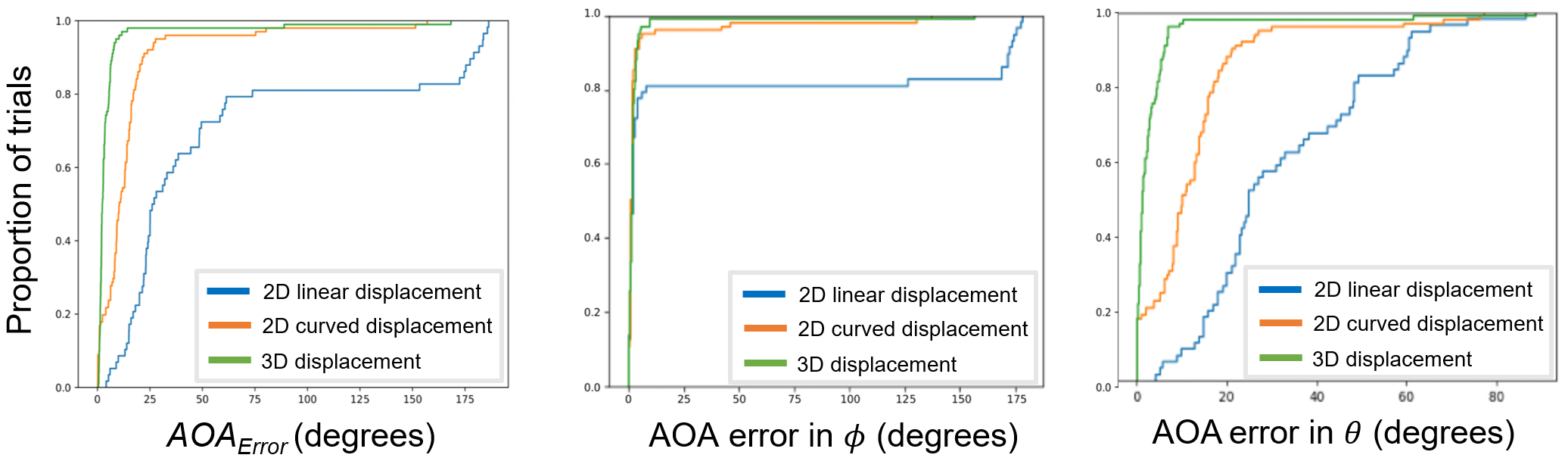}
	\caption{\footnotesize{\emph{Benchmark results obtained using ground truth displacement for a receiving robot in hardware experiments. The transmitting robot is at a distance of 10m. (a) CDF plot for $AOA_{Error}$ error for 101 trials of each motion type (linear, curved and arbitrary 3D). 3D displacement have the lowest $AOA_{Error}$ (def. \ref{def:AoA-error}) which is less than $\ang{10}$ for 95\% of trials. For planar 2D curved and linear displacement only 50\% and 7\% of trials show error below \ang{10} respectively. (b) and (c) show the individual CDF plots for absolute error in AOA for $\phi$ and $\theta$ respectively. 2D linear displacement have the highest error overall. We note that even ground truth displacement incur AOA error due to factors mentioned in Sec. \ref{sec:ps_trajectory_impact}}.}}
    \label{fig:CDF_phi_theta}
\end{figure*}

\subsubsection*{\textbf{Software setup:}} We use Ubuntu 16.04 with kernel 4.15 as the OS for the on-board computer. It runs the following: control code for the robots that uses ROS and MAVROS, pose estimation for the T265 VIO camera using realsense libraries\endnote{Code samples used for T265 VIO camera to get position estimates can be found in their github repository : https://github.com/IntelRealSense/librealsense} and a modified WiFi driver with 802.11 CSI toolbox~\citep{Halperin_csitool} that collects CSI data which is used to calculate the signal phase. The AOA profile is calculated off-board using Matlab. As per our assumption, global clock synchronization among the robots is achieved by using NTP on the on-board computers. 

\subsection{Moving ends}\label{sec:results_movingEnds}
In this section, we show results for hardware experiments that demonstrate the solution to the \emph{moving ends} problem (Sec.~\ref{sec:moving_ends}). Essentially, we show the accuracy of the AOA profile $F_{ij}(\phi,\theta)$ obtained using relative displacement $\boldmath{\chi_{ij}\textbf{(t)}}$ of the receiving robot $i$ with respect to a moving transmitting robot $j\in\mathcal{N}_i$. Relative displacement can be obtained using different on-board sensors, however for proof-of-concept purposes of our developed framework, we use estimates of local robot displacement obtained using the external motion capture system. The following scenarios are evaluated:
\begin{enumerate}
     \item Robot $i$ performs a helix motion at 0.5 m/s (Fig.~\ref{fig:Relative_Traj_comparison_3D})
     \item Robot $i$ performs a linear motion at 0.1 m/s (Fig.~\ref{fig:Relative_Traj_comparison_2D})
\end{enumerate}
The transmitting ground robot $j$ moves along a linear trajectory with varying speed (max speed of $0.1$ m/s), and is at a distance of $10$ m from robot $i$. A sample result for these scenarios shows that $F_{ij}(\phi,\theta)$ obtained using relative displacement of robot $i$ for a moving robot $j$ is similar (Fig.~\ref{fig:Relative_Traj_comparison_3D}) or better (Fig.~\ref{fig:Relative_Traj_comparison_2D}) than that obtained for a stationary robot $j$ (Fig. \ref{fig:Traj_comparison}).

We make interesting observations with respect to the use of relative displacement. For example, when the robot $i$ and $j$ are moving linearly along different directions and with different speeds, the relative displacement $\boldmath{\chi_{ij}\textbf{(t)}}$ has a curved geometry thus eliminating ambiguities along $\phi$ (Fig. \ref{fig:Relative_Traj_comparison_2D}). We postulate that depending on the direction and variation in speed of the moving robots, displacement geometries with more informativeness and thus less ambiguity can be naturally obtained. Further investigation of this observation, namely, exploiting relative displacement for obtaining higher informativeness, is an interesting avenue for further research. Results pertaining to the informativeness of different displacement geometries are discussed in the following section.

\subsection{Informativeness of 2D and 3D displacement geometry}\label{sec:results_informativeness}
\begin{figure*}
	\centering
        \hspace{20pt}\textbf{\large{Ambiguity in AOA profile for displacement geometries}}\\
	    \vspace{5pt}
        \includegraphics[width=16.3cm,height=8.5cm]{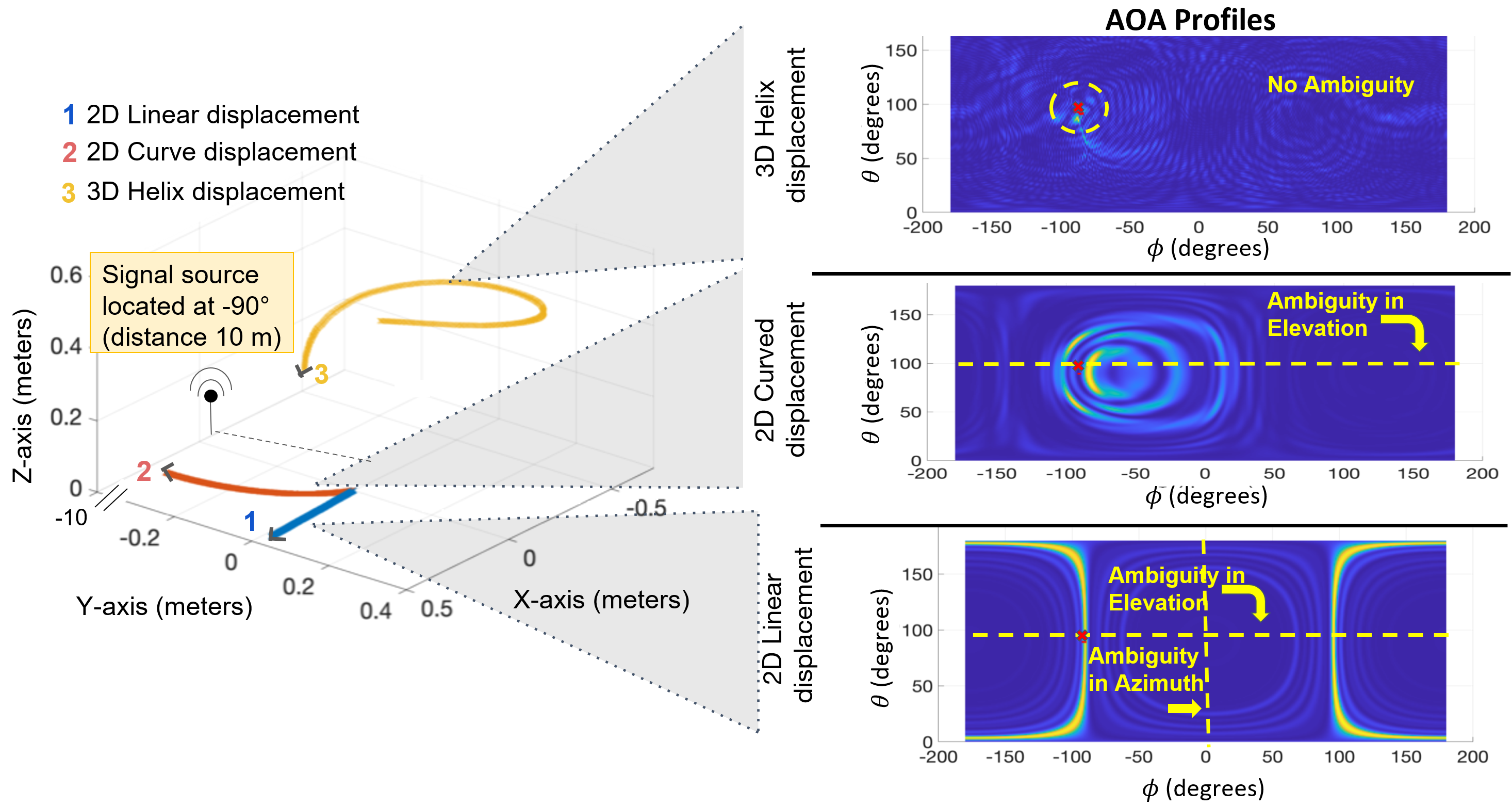}
    	\caption{\footnotesize{\emph{ Shows different robot motions evaluated during hardware experiments and ambiguities in the AOA profile $F_{ij}(\phi,\theta)$ (right). True AOA ( $\phi_g$, $\theta_g$) = ( $\ang{-90}$, $\ang{91}$) to the signal transmitting robot is denoted by red 'x' in the profiles. 3D displacement geometries have higher informativeness and have no ambiguities in the AOA profile.}}}
    	\label{fig:Traj_ambiguity}
\end{figure*}

In this set of experiments, our objective is to study the impact of displacement geometry on $AOA_{max}$ estimation. To analyze this impact without presence of other noise inducing factors, the results of this section use the robots' ground truth displacement obtained from motion capture (the impact of displacement estimation errors on $AOA_{max}$ is presented subsequently in Sec.~\ref{sec:results_traj_erro_AOA}). In these experiments, the relative $AOA_{max}$ is computed between a stationary transmitting robot and the receiving robot having the following motions - a) 2D linear, b) 2D curved, and c) arbitrary 3D displacement (See Fig. \ref{fig:Traj_ambiguity}). The robots are separated by a distance of 10m and are in the line-of-sight of each other. A ground robot is used to obtain 2D displacement samples while an aerial robot is used for 3D displacement. The application of our system to non-line-of-sight scenarios is presented subsequently in Section~\ref{sec:hardware_application}.

Fig.~\ref{fig:CDF_phi_theta} shows the aggregate performance over 101 trials for each robot motion. Overall, the $AOA_{Error}$ (Fig.~\ref{fig:CDF_phi_theta} (a)) is significantly lower for 3D displacement as compared to its 2D counterparts. Specifically, 3D displacement geometries demonstrate less than 10 degrees $AOA_{Error}$ for 95\% trials, where as only 50\% of 2D curved displacement and 7\% of 2D linear displacement show similar accuracy. This trend is again shown empirically in our application scenario results in Section~\ref{sec:hardware_application}.

Hardware results also reveal important intuition behind the informativeness of different displacement geometries (Fig.~\ref{fig:Traj_ambiguity}). For motion with lower informativeness such as the linear 2D displacement, ambiguities are observed in $F_{ij}(\phi,\theta)$ for relative direction $(\hat{\phi}, \hat{\theta})$ of the transmitting robot. This is due to the inability of the linear antenna array geometry traced by receiving robot to uniquely distinguish a transmitting robot's location for certain direction tuples $(\phi,\theta)$ using phase differences. It aligns with the behavior of physical antenna arrays; planar arrays cannot distinguish between directions which are symmetric with respect to the array plane~\citep{manikas2004differential}. This problem is thus greatly alleviated by using displacement geometries with higher informativeness.

In conclusion, our results indicate that 3D displacement leads to the highest informativeness amongst all displacement geometries studied, resulting in a more accurate estimation of $AOA_{max}$ in both azimuth and elevation as predicted by the analysis from Section~\ref{sec:Analysis_impact_factors}.

\subsection{Effect of displacement estimation error on AOA}\label{sec:results_traj_erro_AOA}
Computing AOA profiles $F_{ij}(\phi,\theta)$ requires knowing a robot's local displacement. Since most robots \emph{in the wild} rely on on-board sensors to estimate their position, in this section, we experimentally study the impact of error in local displacement estimation on $F_{ij}(\phi,\theta)$; specifically on the accuracy of $AOA_{max}$ (maximum magnitude path in $F_{ij}(\phi,\theta)$).
\begin{figure}
	\centering
        \hspace{20pt}\textbf{Displacement estimation error for}\\
        \vspace{1pt}
        \hspace{20pt}\textbf{Intel Tracking Camera T265}\\
        \vspace{3pt}
        \includegraphics[width=8.25cm,height=4.0cm]{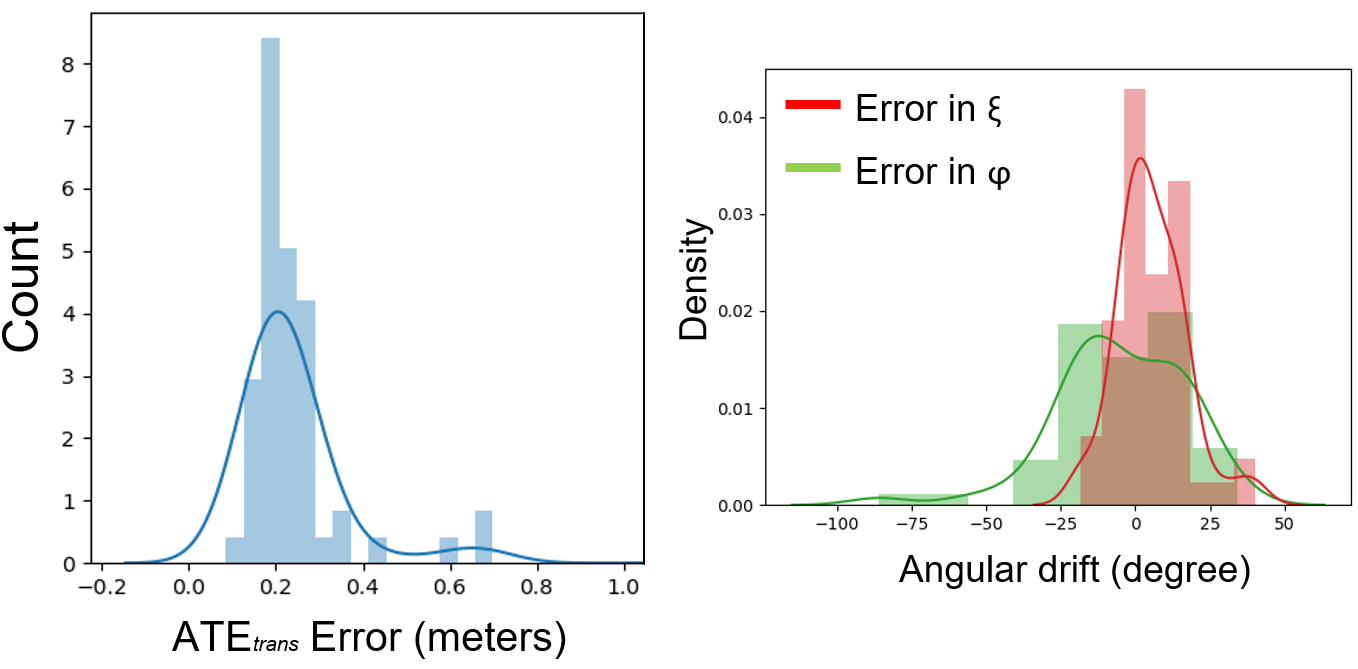}
        \vspace{-0.05in}
        \caption{\footnotesize{\emph{Estimation error ($ATE_{trans}$, Eqn. 7) in robot displacement when using Intel Realsense Tracking Camera T265 from 57 hardware trials. The cumulative length of robot displacement for the trails is 2.8 m on average. (Left): $ATE_{trans} error$ (Cartesian coordinates) across all trials, the mean error being 0.2m. (Right): Corresponding mean angular drift (See def.~\ref{def:angular-drift}) along Spherical coordinates. A higher drift is observed along $\varphi$ compared to $\xi$.}}}
    	\label{fig:ATE_trans_error_cameraTraj}
    	\vspace{-0.2in}
\end{figure}

Our empirical analysis includes both a simulation and a hardware component. The simulation provides a controlled study in which noise can be added systematically to reveal error trends and in the absence of other impacting factors like varying displacement geometry or noisy signal phase and multipaths. The hardware component provides aggregate results showing the impact of displacement estimation error on $AOA_{max}$ accuracy in an actual all on-board sensing robot system. For the hardware experiments, we use displacement estimates provided ``out-of-the-box'' from the T265 VIO camera. No additional optimization techniques were used to improve displacement estimation though we do note that it is possible to do so using state-of-the-art methods (See Sec. \ref{sec:related_works}). Finally, the $AOA_{max}$ estimates obtained are compared against benchmark results from using ground truth displacement.
\begin{figure}
	\centering
        \hspace{20pt}\textbf{Simulated displacement geometries}\\
        \vspace{1pt}
        \includegraphics[width=8.25cm,height=3.5cm]{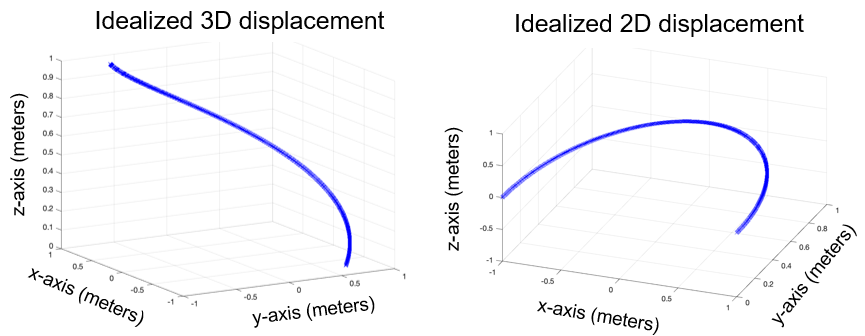}
        \caption{\footnotesize{\emph{Simulated idealized robot displacement geometries in 2D and 3D. The cumulative length for each of them is 3.14 m.}}}
    	\label{fig:ideal_geometries}
    	\vspace{-0.05in}
\end{figure}
\begin{table}
\vspace{-0.05in}
\begin{center}
\begin{tabular}{| c | c | c | c | c |}
\hline
\textbf{$ATE_{trans}$} & \multicolumn{2}{ c |}{\textbf{3D displacement}} & \multicolumn{2}{ c |}{\textbf{2D displacement}}  \\ 
\cline{2-5}
(meters) & \textbf{$\phi$ error} & \textbf{$\theta$ error} & \textbf{$\phi$ error} & \textbf{$\theta$ error}\\
\hline
0.1 m & 2.5$\degree$ & 7.8$\degree$ & 5.7$\degree$ & N/A \\ \hline
0.15 m & 3.75$\degree$ & 10.9$\degree$ & 8.24$\degree$ & N/A \\ \hline
0.2 m & 5$\degree$ & 14$\degree$ & 10.8$\degree$ & N/A  \\ \hline
\end{tabular}
\end{center}
\vspace{-0.05in}
\caption{\footnotesize{\emph{Shows $ATE_{trans}$ error for the noisy displacement geometries obtained after adding simulated angular noise to ideal robot motion (Fig~\ref{fig:ideal_geometries}), as per the noise model explained in Sec.~\ref{sec:traj_error_impact_results_sim}). The cumulative length for each of them is 3.14 m. The corresponding Angle-Of-Arrival estimation error (degree) in azimuth($\phi$) and elevation ($\theta$) directions, is shown for varying levels of displacement estimation error.}}}
\label{table:sim_error_impact}
\vspace{-0.15in}
\end{table}
\begin{figure*}
    	\centering
        \hspace{20pt}\textbf{\large{Impact of displacement estimation for arbitrary 3D displacement}}\\
        \vspace{3pt}
        \includegraphics[width=17.0cm,height=5.0cm]{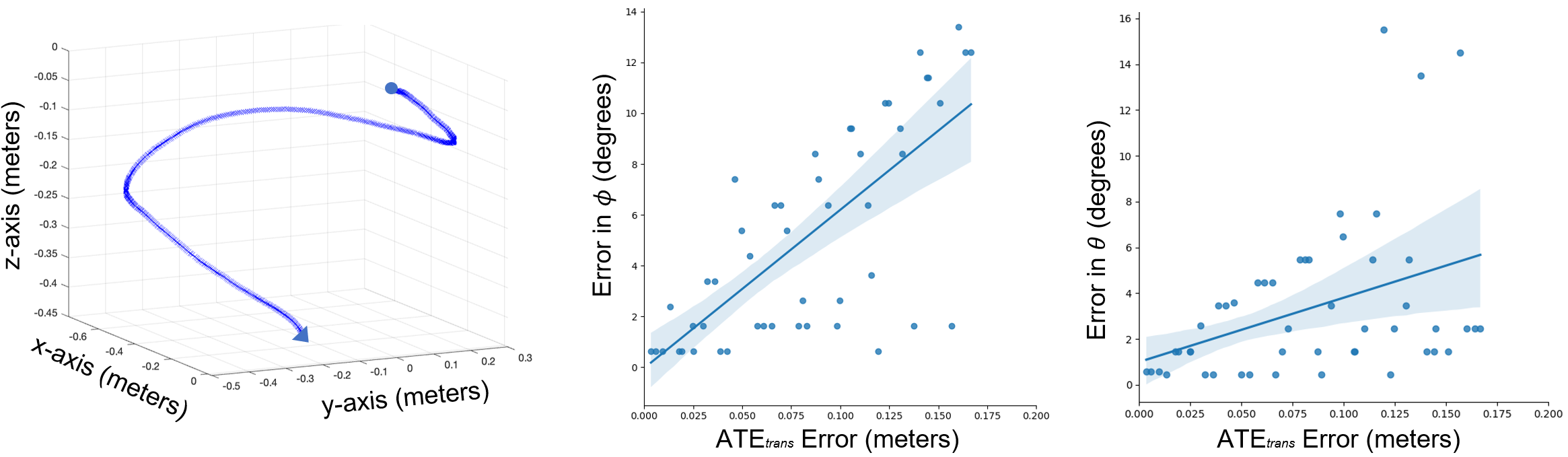}
        \vspace{-0.05in}
        \caption{\footnotesize{\emph{Results showing Angle-Of-Arrival estimation error vs $ATE_{trans}$: Eqn. \ref{eq:ATEtran}) for an actual ground truth 3D displacement of the receiving robot obtained during hardware experiments (leftmost plot). Error is added using simulated noise to the ground truth displacement as per the method detailed in Sec.~\ref{sec:traj_error_impact_results_sim}. The maximum range of $ATE_{trans}$ on X-axis reflects the observed median error in Intel Realsense Tracking Camera T265 displacement estimation (Fig. \ref{fig:ATE_trans_error_cameraTraj}). This result shows an agreement with our analytical results from Sec.~\ref{sec:characterization-trajectory-err-analysis} about the linear relationship between displacement error and AOA error.}} 
    	\label{fig:aoa_vs_ATE_error}}
    	\vspace{-0.1in}
\end{figure*}

\subsubsection{\textbf{Simulation study:}}\label{sec:traj_error_impact_results_sim} We firstly simulate multiple noisy displacement estimates $\boldmath{\hat{\chi_i}}\textbf{(t)}$. We do so by adding accumulating Gaussian noise to idealized robot displacement geometry shown in Fig.~\ref{fig:ideal_geometries}. $F_{ij}(\phi,\theta)$ is then generated for these $\boldmath{\hat{\chi_i}}\textbf{(t)}$ to obtain an $AOA_{max}$ estimate $(\hat{\phi},\hat{\theta}$) which is then compared with the true simulated AOA direction $(\phi_g,\theta_g)$ to compute AOA error (cf. def.~\ref{def:AoA-error}).

In order to generate an $F_{ij}(\phi,\theta)$ (cf. Eqn.~\eqref{eqn:bartlett_estimator_fin}) using simulated noisy displacements as described above, we first simulate the wireless channel $\mathbf{h_{ij}(t)}$, from a stationary transmitting robot $j$, received over $\boldmath{\hat{\chi_{i}}\textbf{(t)}}$. We do so by fixing the true virtual location $p_j(t_k)$ of robot $j$ at $100$m from position $p_i(t_{k})$ in $\boldmath{\chi_i}\textbf{(t)}$. For any position $p_i(t)$, $t \in [t_{k},\hdots,t_{l}]$ where $k$=0, $l$=500 and $d_{ij}(t)$ is the distance between $p_i(t)$ and $p_j(t_k)$, the channel measurements $h_{ij}(t)$ can thus be simulated from Eqn. \ref{eqn:channel_basic_eqn} to obtain $\bold{h_{ij}(t)}$. We then simulate noise in $\boldmath{\chi_i}\textbf{(t)}$ by applying increasing cumulative error in position ($x_i(t)$, $y_i(t)$, $z_i(t)$ for each $p_i(t)$ in $\boldmath{\chi_i}\textbf{(t)}$.

$\boldmath{\hat{\chi_{i}}\textbf{(t)}}$ are simulated in such a way that their $ATE_{trans}$ error (cf. Eqn.~\eqref{eq:ATEtran}) are representative of those characteristic from using on-board sensors. This is based on our observations of estimation error using T265 VIO camera (see Fig. \ref{fig:ATE_trans_error_cameraTraj}). In particular, for each position $p_i(t)$ in $\boldmath{\chi_i}\textbf{(t)}$, we sample position error along $x_i(t)$, $y_i(t)$, \text{and} $z_i(t)$ from a zero-mean Gaussian distribution and add it to each pose cumulatively. This generates $\boldmath{\hat{\chi_{i}}\textbf{(t)}}$ from $\boldmath{\chi_i}\textbf{(t)}$ with an accumulating $ATE_{trans}$ error. To generate additional noisy displacement estimates, we increase the sampled position error for each successive simulated robot displacement steadily so that the $ATE_{trans}$ error ranges from $0.0$ m to $0.2$ m. The corresponding angular drift (See Def~\ref{def:angular-drift}) in $\varphi$ and $\xi$ , for robot pose in spherical coordinates, ranges from $0\degree$ to $10\degree$. A total of 50 such samples are generated for any displacement geometry that is evaluated.   

Our empirical results support our analytical development from \emph{Lemma} 2 and 3 (Sec. \ref{sec:characterization-trajectory-err-analysis}). Table~\ref{table:sim_error_impact} shows results of simulation for idealized robot displacement geometries in 2D and 3D (Fig.~\ref{fig:ideal_geometries}). As shown by our analysis in Sec.~\ref{sec:characterization-trajectory-err-analysis}, higher $ATE_{trans}$ leads to an increase in angular drift that is reflected in the AOA estimation for $\phi$ and $\theta$. It also shows an AOA estimation error $5\degree$ in $\phi$ and $14\degree$ in $\theta$ corresponding to $ATE_{trans}$ 0.2 m in the absence of other factors that impact $F_{ij}(\phi,\theta)$, for perfectly symmetric displacement geometries. We note that given equal error accumulation in $x_i(t)$, $y_i(t)$, and $z_i(t)$, higher error angular drift in corresponding Spherical coordinate $\xi$ is expected compared to that in $\varphi$, thus resulting in higher AOA error in $\theta$ compared to $\phi$.   

To validate the relationship between displacement estimation error and AOA error further, we add simulated noise to a sample 3D ground truth displacement $\boldmath{\chi_i}\textbf{(t)}$ of the receiving robot $i$ obtained from hardware experiments. The AOA error shows an increasing linear relationship to the corresponding $ATE_{trans}$ error for the robot displacement (Fig.~\ref{fig:aoa_vs_ATE_error}). These results can be used as indicators of the expected degradation of AOA estimation from our system given the displacement estimation error in robot's motion. We note that the scale of AOA error in $\phi$ and $\theta$ directions slightly varies due to the use of asymmetric displacement geometries that impact how error in $x_i(t)$, $y_i(t)$, \text{and} $z_i(t)$ (i.e., $ATE_{trans}$ error in Cartesian coordinates) translates to angular drift in $\varphi~\text{and}~\xi$ (in spherical coordinates).
\begin{figure*}
	\centering
    \hspace{20pt}\textbf{\large{CDF of AOA error error for hardware experiments}}\\
    \vspace{5pt}
    \includegraphics[width=17.5cm,height=4.75cm]{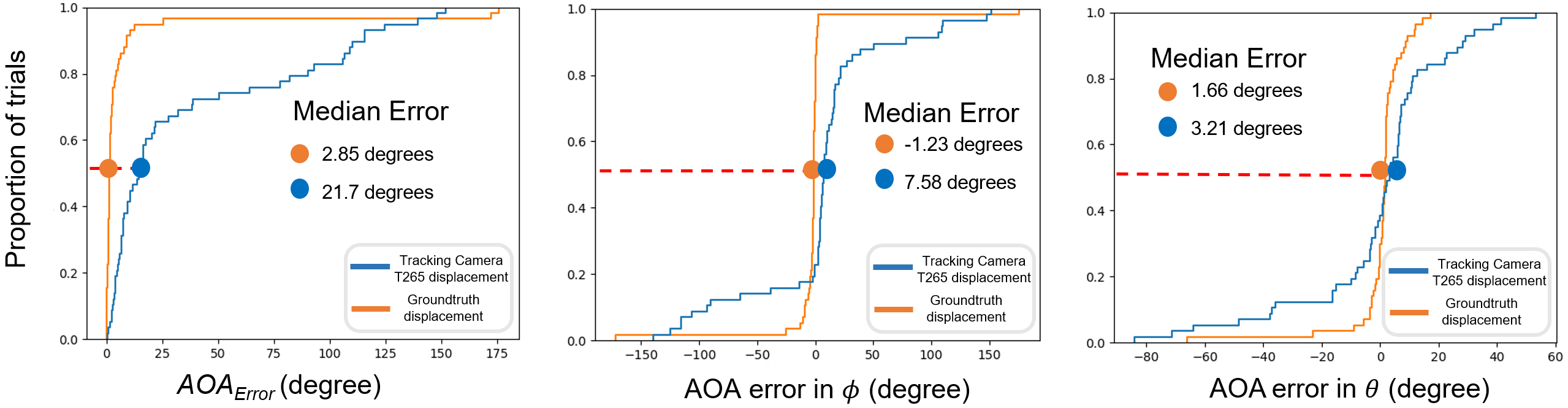}
    \caption{\footnotesize{\emph{Aggregate results showing CDF of AOA estimation error for hardware experiments consisting of 57 trials using UAV. Each experiment records the 3D displacement for the receiving robot (UAV) using the Intel Realsense Tracking Camera T265 camera and motion capture (ground truth), along with their respective $F_{ij}(\phi,\theta)$. The median error for each sub-plot shows higher accuracy for ground truth displacement compared to estimated displacement from T265 camera. $AOA_{Error}$ is obtained using Eqn.~\ref{eqn:AOA_euclidean_error}. The corresponding distribution of error along azimuth and elevation directions is shown in Fig.~\ref{fig:Mocap_vs_camera_AOA}.}}}
    \label{fig:Mocap_vs_camera_AOA_cdf}
    \vspace{-0.1in}
\end{figure*}
\begin{figure*}
	\centering
    \hspace{20pt}\textbf{\large{AOA error distribution comparison for hardware experiments}}\\
    \vspace{5pt}
	\begin{minipage}{.45\linewidth}
         \includegraphics[width=8.0cm,height=4.75cm]{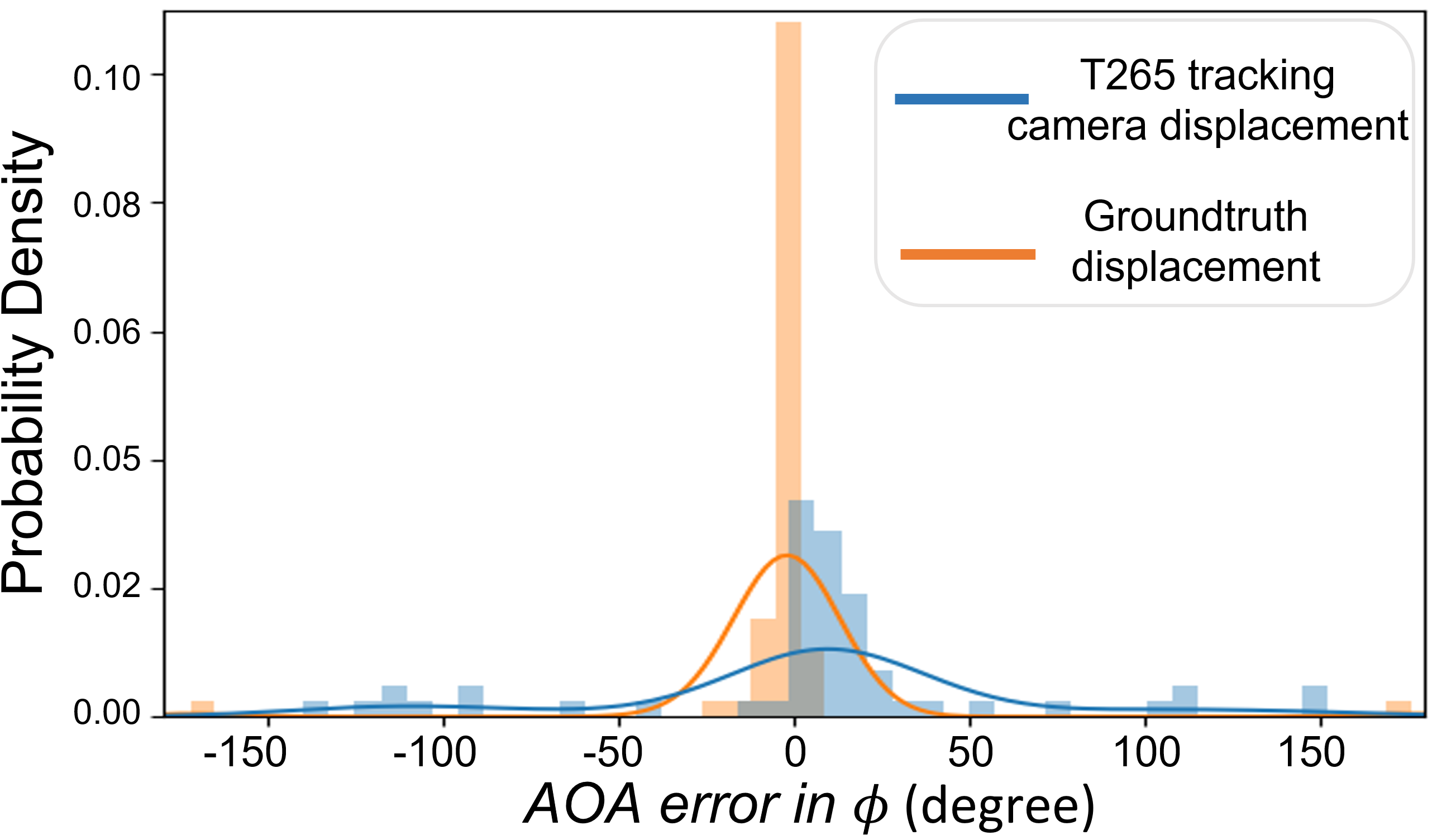}
	\end{minipage}
	\hspace{6pt}
	\begin{minipage}{.45\linewidth}
        \includegraphics[width=8.0cm,height=4.75cm]{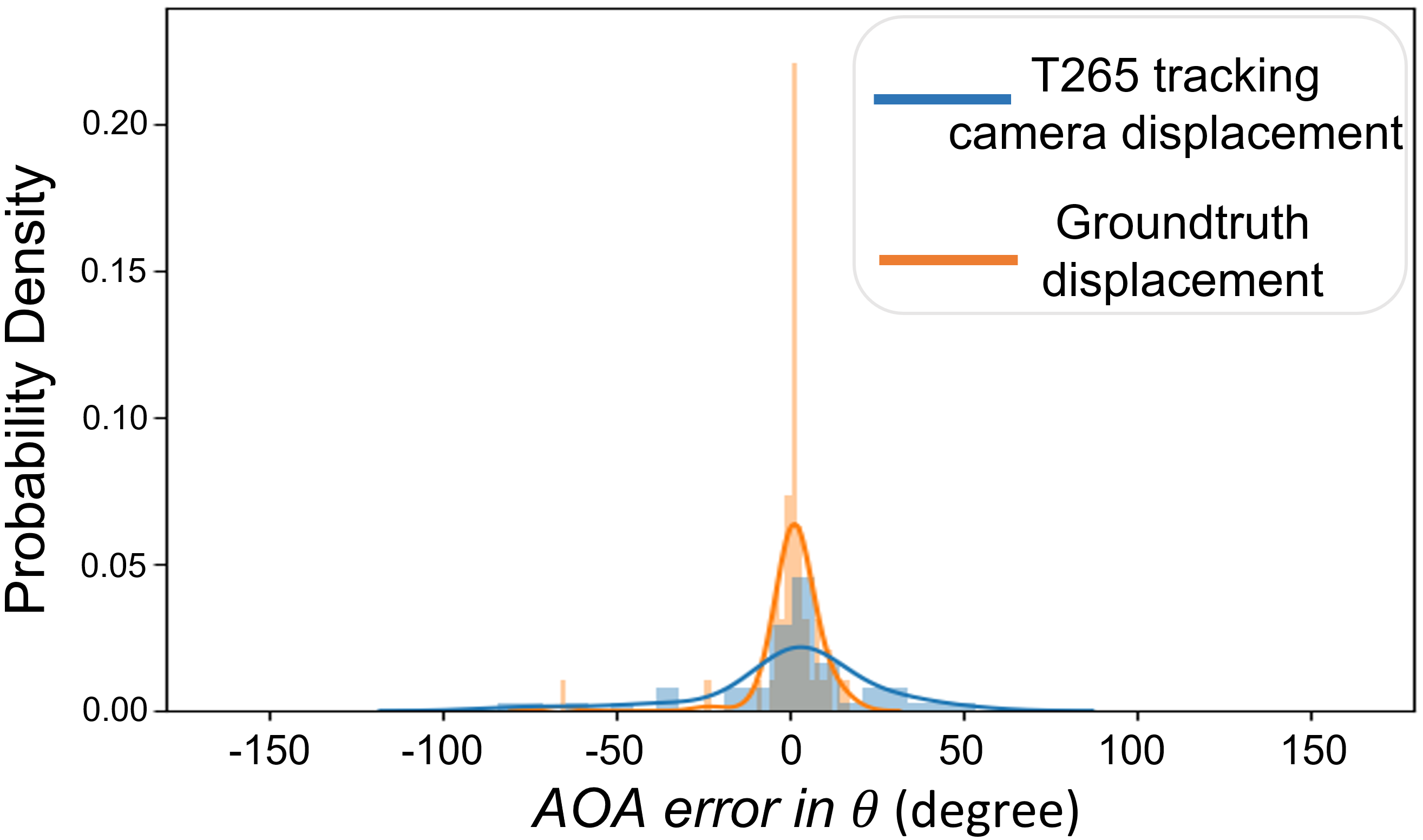}
	\end{minipage}
    \caption{\footnotesize{\emph{Aggregate results for hardware experiments consisting of 57 trials showing error distribution for AOA estimation for hardware experiments. Ground truth AOA $(\phi_g, \theta_g)$ = ( $\ang{0}$, $\ang{91}$). Sub-plot on the left shows error in $\phi$ with mean, std dev of -2.13 $\degree$, 33.23 $\degree$ for ground truth displacement and 4.86 $\degree$, 58.45 $\degree$ for T265 camera displacement. Sub-plot on the right shows error in $\theta$ with mean, std dev -0.65 $\degree$, 10.63 $\degree$ for ground truth displacement and -0.28 $\degree$, 25.22 $\degree$ for T265 camera displacement. Thus, with high error in estimated displacement, the corresponding AOA accuracy is low, compared to that obtained from ground truth displacement. These AOA errors can be also be approximated using a Gaussian distribution as evident from the aggregate plots. A higher error is observed in $\phi$ compared to that in $\theta$ on account of higher angular drift (def.~\ref{def:angular-drift}) in $\varphi$ than in $\xi$ (See Fig.~\ref{fig:ATE_trans_error_cameraTraj}).}}}
    \label{fig:Mocap_vs_camera_AOA}
    \vspace{-0.05in}
\end{figure*}

\subsubsection{\textbf{Hardware experiments:}} We compare $AOA_{max}$ estimation using 3D robot displacement estimates obtained from T265 VIO camera against benchmark $AOA_{max}$ estimates obtained using ground truth. Displacements for the receiving robot are collected simultaneously using the T265 VIO camera (mounted on the aerial receiving robot) and the external motion capture system. Experiments are conducted in realtime, while keeping the transmitting robot stationary (results involving a mobile transmitting robot are presented in Section~\ref{sec:results_movingEnds} and Section~\ref{sec:hardware_application}). The average cumulative displacement of the receiving robot is 2.8 m per AOA profile generated, so as to obtain a sufficiently informative displacement geometry similar to a 3D helix. We run 59 such trials for this experiment. Two trials were discarded due to loss of pose tracking for the T265 VIO camera during the experiments. The CDF plots for AOA estimation error (Fig.~\ref{fig:Mocap_vs_camera_AOA_cdf}) show a low median $AOA_{Error}$ as well lower errors in $\phi$ and $\theta$ using the ground truth displacement compared to that using the estimated displacement.
Figure~\ref{fig:Mocap_vs_camera_AOA} shows the corresponding kernel density estimation distribution plot comparing $AOA_{max}$ estimation error in azimuth angle $\phi$ and elevation angle $\theta$. The mean, std dev of the error in $\phi$ is $-2.13\degree$, $33.23\degree$ respectively for ground truth displacement and $4.86\degree$, $58.45\degree$ respectively for the estimated displacement from the T265 VIO camera. The mean, std dev of the error in $\theta$ is $-0.65\degree$, $10.63\degree$ respectively for the ground truth displacement and $-0.28\degree$, $25.22\degree$ respectively for the estimated displacement.    

The mean $ATE_{trans}$ error in estimated displacement is $0.2$ m for $2.8$ m long cumulative displacement of the receiving robot on average (Fig. \ref{fig:ATE_trans_error_cameraTraj}). Thus, the results shown in Fig.~\ref{fig:Mocap_vs_camera_AOA} represent $AOA_{max}$ estimation error given $\sim7.0\%$ average displacement estimation error. Higher error along $\phi$ than $\theta$ is due to asymmetric angular drift in displacement (See Fig.~\ref{fig:ATE_trans_error_cameraTraj}). We also note that given these are the results for actual hardware experiment using only the robot's local sensors, other factors such as real wireless signal phase, UAV vibrations and small variations in the shape of the displacement geometry also contribute the observed error in AOA estimation. On the other hand, results corresponding to ground truth displacement represent an empirical lower bound for $AOA_{max}$ error that could be attainable using improved position estimation accuracy while accounting for other factors that impact $F_{ij}(\phi,\theta)$.
\vspace{-0.1in}
\section{Application Study}
\subsection*{Dynamic rendezvous between ground/air robots} \label{sec:hardware_application}
In this section, we demonstrate the performance of our system for a complete experiment of multi-robot rendezvous task between a UAV (signal receiving) and a ground robot (signal transmitting). The goal is to evaluate the utility of our system for a multi-robot coordination application that involves a heterogeneous team of robots. Our evaluation is based on the following three criteria:
\begin{itemize}
     \item \emph{System performance in the presence of occlusions}: The UAV should navigate towards the ground robot using the AOA profile $F_{ij}(\phi,\theta)$ despite of any visual occlusions.
     \item \emph{Performance of our system in the presence of moving ends}: The UAV should be able to navigate towards a moving ground robot using its relative displacement and our 3D SAR formulation, while also maintaining continuous mobility.
    \item \emph{Impact of displacement geometry}: Following the development in Sec.~\ref{sec:cr-bound}, verify that the informativeness of displacement geometry can be used to predict AOA estimation performance.
\end{itemize}
\begin{figure*}
	\centering
	\textbf{\large{Dynamic rendezvous experiment between UAV and ground robot}}\\
	\vspace{10pt}
	\begin{minipage}{.495\linewidth}
		\centering
        \includegraphics[width=8.5cm,height=5.5cm]{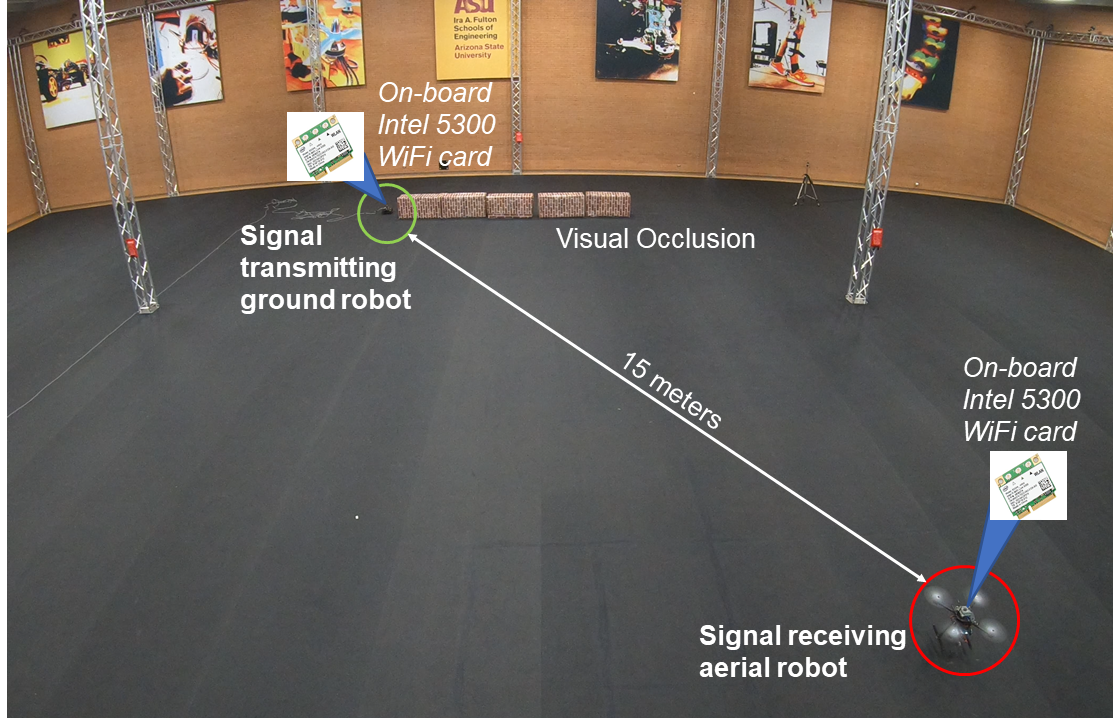}\\
        \footnotesize\emph{{(a) Initial setup}}
    	\vspace{7pt}
	\end{minipage}
	\begin{minipage}{.495\linewidth}
		\centering
        \includegraphics[width=8.5cm,height=5.5cm]{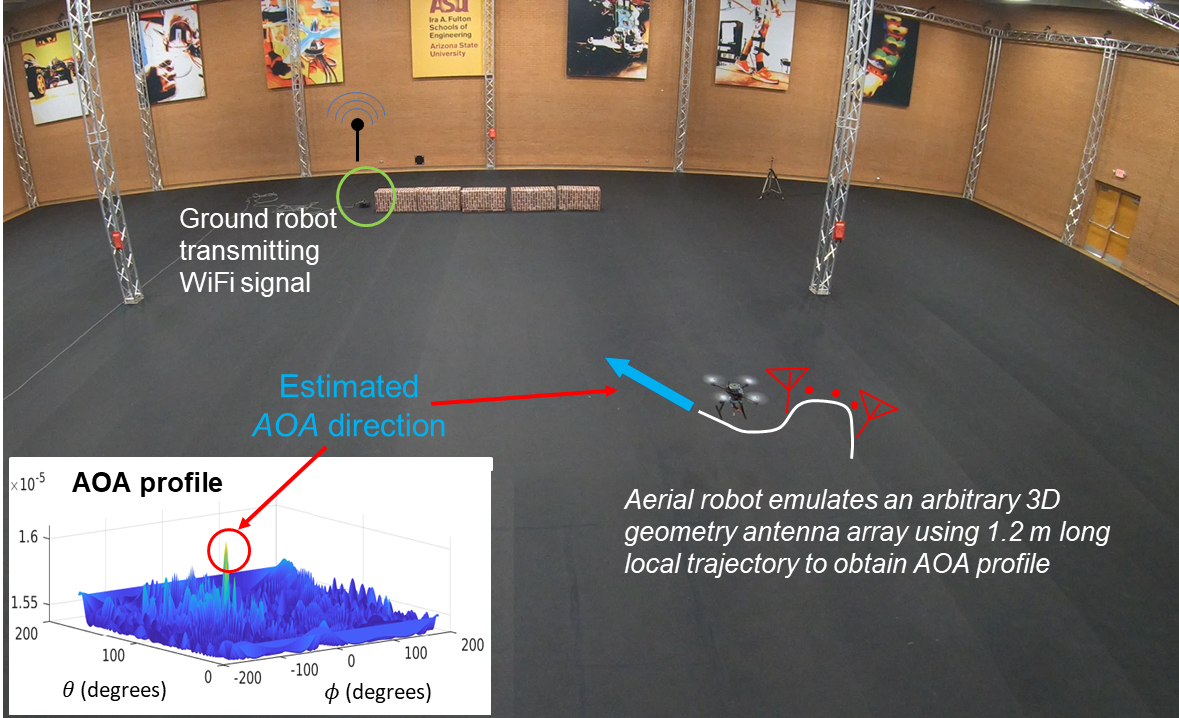}
        \footnotesize\emph{{(b) UAV starts navigating towards static ground robot}}
        \vspace{7pt}
	\end{minipage}
	\begin{minipage}{.495\linewidth}
		\centering
        \includegraphics[width=8.5cm,height=5.5cm]{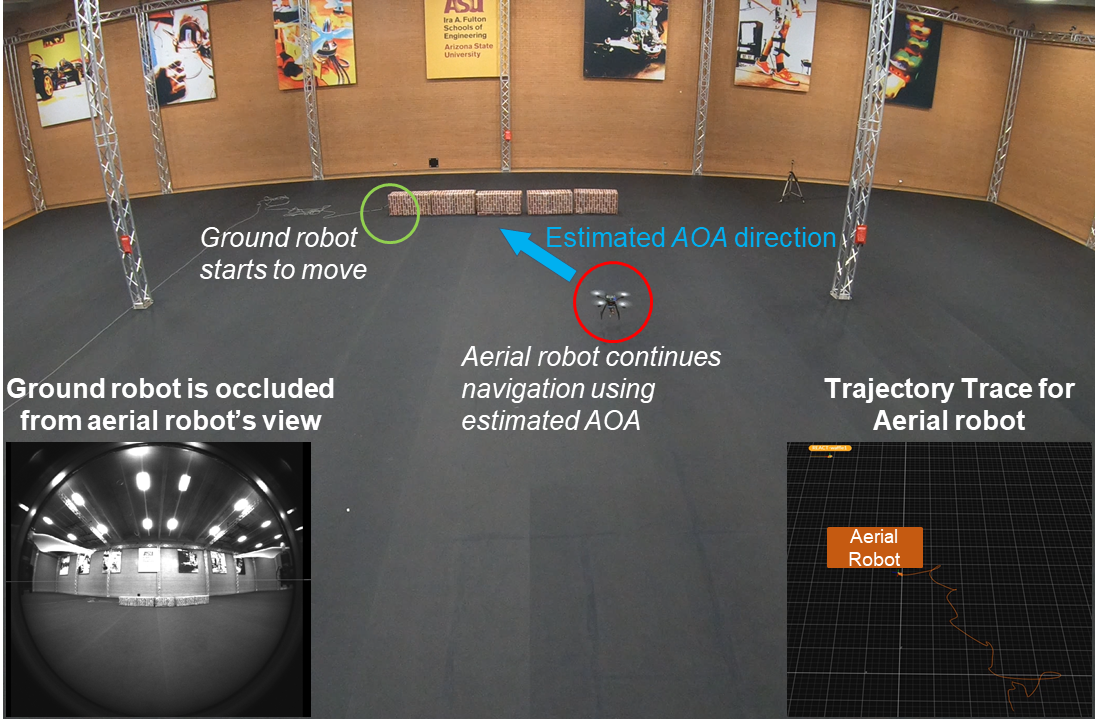}\\
        \footnotesize\emph{{(c) Ground robot starts to move and gets occluded}}
	\end{minipage}
	\begin{minipage}{.495\linewidth}
		\centering
        \includegraphics[width=8.5cm,height=5.5cm]{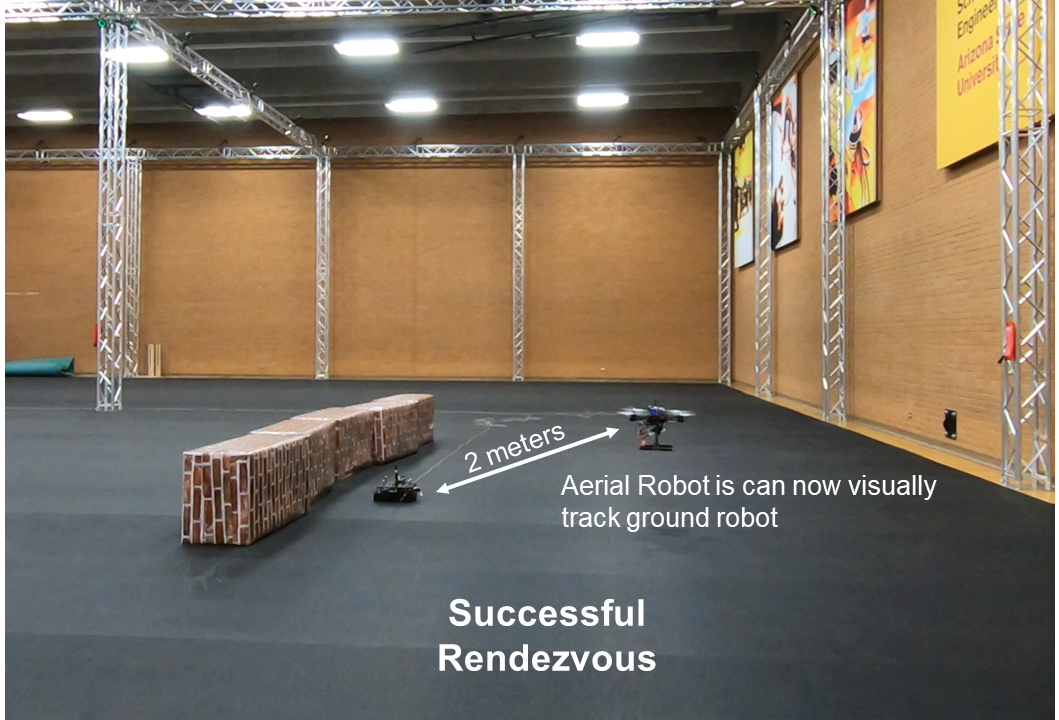}\\
        \footnotesize\emph{{(d) UAV achieves rendezvous with ground robot}}
	\end{minipage}
	\caption{\footnotesize\emph{{Shows different stages of dynamic rendezvous hardware experiment between signal transmitting ground robot and receiving UAV. Starting from top-left : a) The initial setup to test rendezvous of UAV with the ground robot in presence of occlusions. b) The UAV begins to navigate towards the stationary ground robot, using relative AOA information obtained from continuously generating AOA profiles using our arbitrary 3D SAR system which leverages the UAV's local displacement obtained from its natural motion. c) The ground robot starts to move and gets occluded (UAV loses visual of the ground robot as seen in the fisheye view image). d) The UAV continues its navigation towards the ground robot despite of occlusion until it is within 2 meters proximity, at which point rendezvous is declared successful.}}}
    \label{fig:HW_application}
\end{figure*}

\subsubsection*{\textbf{Experimental setup:}}  We consider an environment of size $300$ m$^{2}$ with one UAV and one ground robot separated by distance of $15$ m and occlusions (Fig.~\ref{fig:HW_application} (a)). We note that these occlusions block visual line-of-sight but are penetrable by WiFi signals. Motion capture is used to provide displacement information for the UAV (robot $i$). The ground robot $j$'s position is not known to robot $i$. True AOA is obtained between robot $i$ and $j$ using their respective positions obtained from the motion capture system.

The UAV, after take-off, starts to successfully navigate towards the stationary ground robot using relative $AOA_{max}$ obtained from $F_{ij}(\phi,\theta)$ and a simple motion controller (Fig.~\ref{fig:HW_application} (b)). The ground robot then starts to move and gets visually occluded by obstacles (Fig.~\ref{fig:HW_application} (c)), but the UAV continues to successfully navigate towards it. We consider the rendezvous successful when the UAV is within a $2$ m radius of the ground robot (Fig.~\ref{fig:HW_application} (d)).

\subsubsection*{\textbf{Evaluation:}} For evaluation of our system performance, we use $AOA_{Error}$ (Eqn. \ref{eqn:AOA_euclidean_error}) to measure the accuracy of our $AOA_{max}$ estimation against ground truth AOA. Our evaluation focuses on the following: 1) evaluating $AOA_{Error}$ over time, 2) evaluating $AOA_{Error}$ for the moving ends case where both the transmitting and receiving robots are moving, and 3) experimental characterization of the robot displacement geometry's informativeness versus $AOA_{Error}$. During the experiment, the UAV's average cumulative displacement is 1.2 m to measure a complete $F_{ij}(\phi,\theta)$. We refer to this as a \emph{SAR iteration}. For each SAR iteration, the UAV uses its relative displacement and received signals to calculate $F_{ij}(\phi,\theta)$ and updates its waypoint to navigate towards the ground robot using estimated $AOA_{max}$. 
\begin{figure*}
    \centering
	 \textbf{\large{Error in AOA during Dynamic rendezvous experiment}}\\
	\vspace{5pt}
    \includegraphics[width=17.2cm,height=9.5cm]{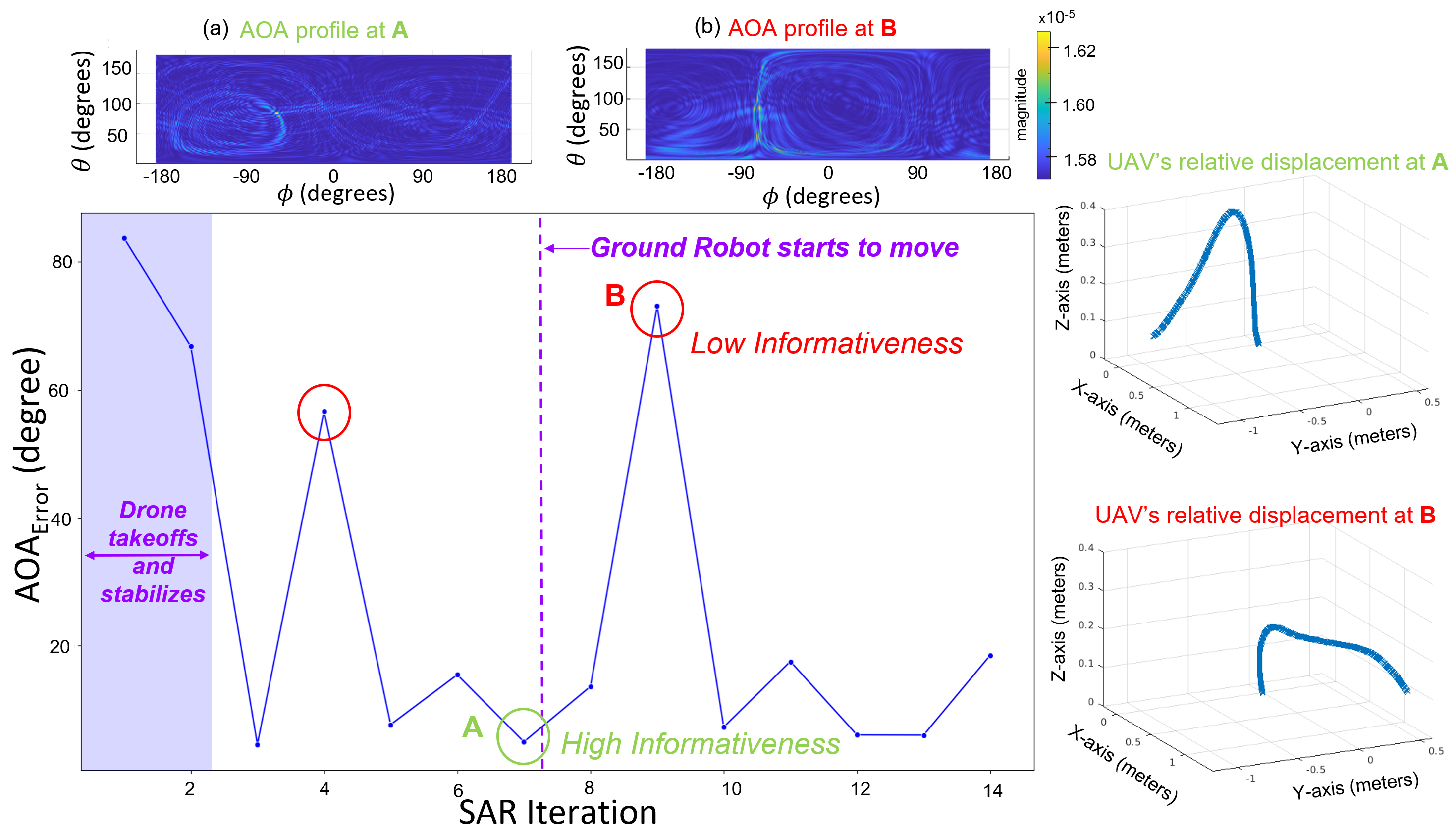}
	\caption{\footnotesize{\emph{Plot showing $AOA_{Error}$ (Eqn.\ref{eqn:AOA_euclidean_error}) for signal receiving UAV during the dynamic rendezvous hardware experiment. Our approach considers the motion of the transmitting robot (start indicated by vertical dotted line) while the UAV continues to navigate towards it. During each SAR iteration, the UAV travels a distance of 1.2 m on average. AOA profiles and the UAV displacement are shown for iteration 7 (point A) and iteration 9 (point B). High error at point B is due to low informativeness of the corresponding displacement geometry (show on the right) and an ambiguous $F_{ij}(\phi,\theta)$ (shown at the top). Similar behavior is observed for iteration 4. Conversely, the UAV displacement geometry at point A is highly informative and has almost no ambiguity leading to low $AOA_{Error}$. High AOA error for the first two iterations is due to sudden jerk and high UAV chassis vibrations during take-off.}}}
	\label{fig:HW_application_stats}
\end{figure*}

Fig.~\ref{fig:HW_application_stats} shows the resulting accuracy of the $AOA_{max}$ estimation for each iteration. Throughout the experiment, the mean $AOA_{Error}$ between robots is 10.17 degrees except for two outlier points at iteration 4 and 9. Note that $AOA_{Error}$ is not affected by the motion of the ground robot, which begins at iteration 7. A closer look at iteration 9 and 4 reveals that the $F_{ij}(\phi,\theta)$ contains ambiguous AOA directions owing to displacement geometries of low informativeness (high CRB) at these iterations (see Fig.~\ref{fig:HW_application_stats}, Profile (b)). Intuitively, this occurs when the motion of the UAV is in 2D with little displacement along the z-axis and is predicted by a high CRB value at these points in the experiment (see Fig.~\ref{fig:HW_application_stats_crb_error}). In contrast, all other iterations where the UAV's displacement geometry has high informativeness, result in more accurate $AOA_{max}$ and less ambiguous $F_{ij}(\phi,\theta)$ (e.g., Fig.~\ref{fig:HW_application_stats} AOA profile (a)).

\subsubsection*{\textbf{Observations:}} 
\begin{figure}
        \centering
        \hspace{20pt}\textbf{CRB for UAV displacement geometry}\\
	    \vspace{5pt}
        \includegraphics[width=8.5cm,height=5.0cm]{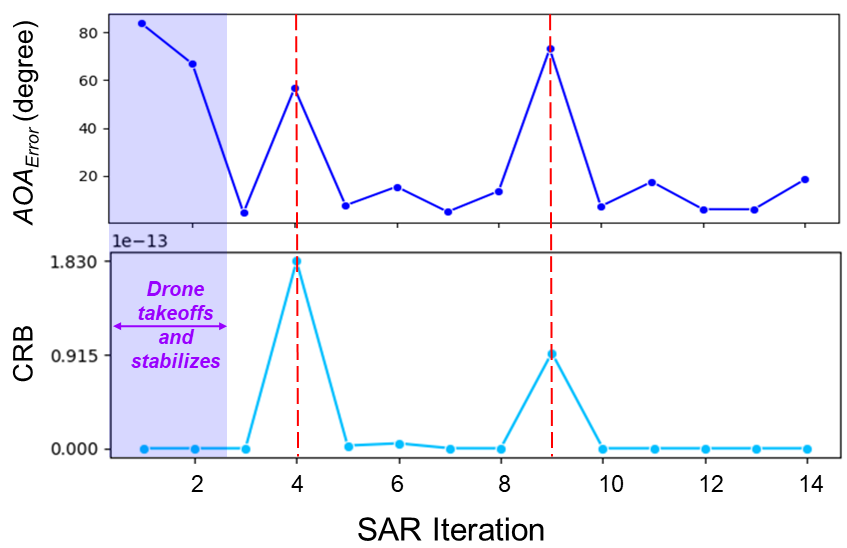}
	    \caption{\footnotesize{\emph{Shows that CRB can be used to presage AOA accuracy that would result from low informative displacement geometries. The high CRB values at timesteps 4 and 9 indicate lower informativeness and higher ambiguity in $F_{ij}(\phi,\theta)$ (see Fig.~\ref{fig:HW_application_stats}). Correspondingly, they have high $AOA_{Error}$. We note that despite the CRB being low for the first two iterations, the $AOA_{Error}$ is high, generally due sudden jerk and high vibrations of the UAV  chassis.}}}
    	\label{fig:HW_application_stats_crb_error}
\end{figure}

We make the following important observations during this experiment:
\begin{itemize}
    \item \emph{Tracking through visual occlusions:} Our system enables the UAV to sense the ground robot's spatial direction corresponding to $AOA_{max}$ obtained by continuously generating AOA profiles $F_{ij}(\phi,\theta)$ in
    realtime.
    \item \emph{Accurate $AOA_{max}$ estimation with moving ends:} Despite the presence of a moving transmitting ground robot, our 3D SAR formulation allows the signal receiving UAV to use it's relative displacement with respect to the ground robot and achieve rendezvous.
    \item \emph{Informativeness as a predictor of AOA accuracy:} Iterations 4 and 9, which have high error in $AOA_{max}$ (Eqn.~\ref{eqn:AOA_euclidean_error}), have corresponding high CRB values indicating that the corresponding displacement geometries have low informativeness (Fig.~\ref{fig:HW_application_stats_crb_error}). By contrast, low CRB values correctly predict low $AOA_{Error}$.
\end{itemize}
\noindent Thus our experimental results support our theoretical developments with regards to addressing the moving ends problem, and characterizing the accuracy of measured $AOA_{max}$ as a function of the informativeness of a robot's displacement geometry which is computable using Eqn.~\eqref{eq:CRB_helix_closed_form}.   

We note a few things about this experiment. As per the assumption in Sec. \ref{sec:Problem Formulation} and Sec. \ref{sec:moving_ends}, the relative $AOA_{max}$ to the robot $j$ is calculated with respect to the initial position $p_i(t_k)$ of the robot $i$. Thus if the true AOA with respect to this initial position undergoes a substantial change over a SAR iteration, as could happen when robots are at short distances to one another or when robots are moving very fast relative to each other, then the $AOA_{max}$ estimation will be outdated. Hence, for our experiments we limit the UAV and ground robot speeds to 0.5 $\texttt{m/s}$ and 0.1 $\texttt{m/s}$ respectively. Additionally, once the robots are close enough to each other, we end the experiment. At this point we assume that the UAV is equipped with an on-board camera that can detect the ground robot and maintain tracking. For example, at $2$ m distance between robots, April tags can be used to obtain relative position estimates with very high accuracy~\citep{april_tag2}. Additionally, the CRB is obtained via post-processing the collected data and using the true AOA direction of the ground robot in Eqn.~\eqref{eq:CRB_helix_closed_form}. Future implementations could consider online CRB computation using a \emph{worst-case} source direction which could then be used to \emph{optimize} robot motion for improved $AOA_{max}$ estimation.  We do not consider such optimizations here and rather leave this as an interesting direction of future work. 

This proof-of-concept application study demonstrates the applicability of our developed framework to multi-robot tasks (e.g coverage or exploration) that require robots to maintain continuous mobility in an environment with obstacles and visual occlusions. Moreover, use of 3D robot motion during the task improves the information quality obtainable by our system. We note that if only ground robots are used instead of a heterogeneous robot team, then it would be sufficient to rely only on AOA estimation along the azimuth direction as long as the robots are operating in the same 2D plane. On the other had, a ground robot can use AOA estimation to an aerial robots along the elevation direction in the range [$0\degree$, $90\degree$] while trading-off some accuracy.

\section{Discussion} \label{sec:discussion}
In this discussion section we address some practical considerations that need to be taken into account when integrating this system with robot platforms in its current form as well as avenues of improvement and future work. 

\subsection*{Practical considerations}
We consider how our system can be used by a team comprising of only ground robots as well as additional factors that can impact AOA estimation during implementation of our proposed framework.


\paragraph{Discretization:} 
We note that substantial discretization of the virtual antenna array elements resulting from a combined effect of displacement estimation errors and significant WiFi packet loss during transmission, is possible in some implementations. We leave this analysis as an avenue for investigation in future work and in this paper assume that the loss in packet transmission is not severe. For implementation scenarios, a future version of the work could include control of packet transmission as well as optimization of robot motion to diminish the occurrence and potential impact of discretization effects.

\paragraph{Impact of packet detection delay and channel sampling time offsets (STO):} 
For actual implementation, we make a similar assumption as that in~\citep{Gil2015AdaptiveCI} about the static nature of the channel, i.e coherence time of the phase is negligible compared to the time delay between the packet. Although in practice there is small delay (in microseconds) in packet transmission for the forward and reverse channel using the WiFi card, we found that the impact on final phase (after cancelling CFO) is not substantial. Since the impact of STO is more of an implementation challenge, we do not address it in this paper which focuses more on developing the analytical framework. It is one of the design challenges for the software toolbox release of our framework as part of future work and can be addressed by interpolating the phase from channel subcarriers (30 that are accessible in Intel 5300 WiFi card) to calculate the phase corresponding to true center frequency of the channel. In this paper, we use the phase corresponding to the subcarrier 15 of the WiFi channel. We also plan to relax the time synchronization requirement for cancelling CFO in future work.

\subsection*{Future work}
Here, we summarize a few directions of interesting future work that could build upon our developed framework and accompanying analysis.

\paragraph{Optimizing informativeness:} 
This paper develops the concept of \emph{informativeness} of different displacement geometries for the purpose of obtaining AOA profiles $F_{ij}(\phi,\theta)$ between robots. Interesting avenues of future investigation could include studying the trade-offs between different levels of informativeness and the modification of the robot's natural motion, i.e. co-optimizing AOA estimation with robot motion to improve \emph{informativeness} as the robots achieve their primary coordination task. Additionally using relative displacement could naturally lead to more \emph{informativeness} as seen in Sec. \ref{sec:results_movingEnds}.

\paragraph{Localization and SLAM:}
As our proposed method allows for co-optimizing the AOA estimation with the robot’s motion in order to improve AOA accuracy, it shows a promising potential for active perception capability. Our method also returns the AOA profile $F_{ij}(\phi,\theta)$, which captures the physical interactions of the transmitted signals with the environment. This is in contrast to commonly used commercial range-only technologies (e.g., UWB) that provide only bearing or range measurements. Interesting avenues for future research include using the full AOA profile for improving robustness of multi-robot localization/SLAM methods to adversarial action. For example this profile has been shown to be a unique fingerprint for each mobile robot~\citep{Gil2015Spoof-Resilient} or can potentially be used as a trust metric when robots are exchanging information with each other~\citep{MallmannTrenn2020CrowdVR}. Another potential application would be to use a single AOA profile measured at some location for loop closures as opposed to other wireless signal based approaches that use multiple RSS measurements from stationary signal access points~\citep{Yen2020OrientationCF,Liu2020CollaborativeSB}.
\begin{figure}
    \centering
    \textbf{Arbitrary motion of manipulator arm}\\
    \vspace{5pt}
    \includegraphics[width=7.5cm,height=6.7cm]{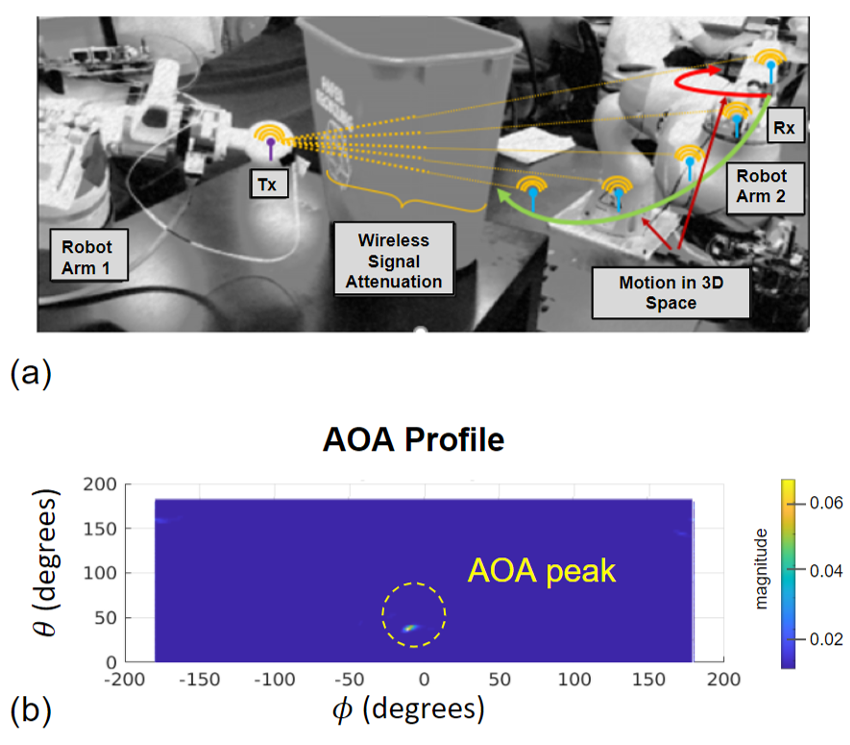}
	\caption{\footnotesize{\emph{A proof-of-concept hardware experiment where an AOA profile is obtained from arbitrary 3D motion of a manipulator arm using the methodologies presented in this paper. Here, a WiFi patch antenna was adhered to the manipulator at the end of the robot arm and ping packets from a second manipulator arm were collected for AOA profile generation.}}}
	\label{fig:future_work_manipulator_arm}
\end{figure}

\paragraph{Multi-robot perception:} 
The developed framework can improve \emph{multi-robot} perception where sensed AOA profiles can be shared with neighboring robots. This could enable several interesting capabilities such as positioning for the entire team, multi-robot mapping without the need for a common reference frame, and/or performing sensor fusion of this information with other more traditional on board sensors such as cameras or LiDARs to learn about the environment.  

\paragraph{Robot manipulation:} 
One perhaps surprising outcome of our work on generalizing SAR capabilities to arbitrary robot displacement in 3D, is the applicability of this method to robot manipulator arms, that also traverse 3D space. Additionally for these systems, the arbitrary 3D SAR development from this paper can be applied fruitfully, taking advantage of the fact that the manipulator displacements in 3D space can be estimated with very high precision using inverse kinematics. The analysis on informativeness of displacement geometries can also be used to co-optimize manipulator arm motion for enhancing attainable $F_{ij}(\phi,\theta)$, as the manipulator achieves its primary task, such as approaching or manipulating an object. Fig.~\ref{fig:future_work_manipulator_arm} shows a proof-of-concept where we measured $F_{ij}(\phi,\theta)$ information between two manipulator arms working in unison at the Stanford Robotics Lab in Palo Alto, California.


\section{Conclusion}
\label{sec:conclusion}
This paper develops a system that enables robots to use their local information to obtain signal multipaths i.e AOA profiles for other robots on their team with which they can exchange ping packets. This has implications for positioning, adhoc robot networks and security, among others. Our key idea is to exploit estimated robot displacement to trace virtual 3D antenna arrays for attaining AOA profiles, even in the presence of robot motion at both ends (receiving and transmitting robots). We provide complete theoretical analysis, supported by in-depth experiments, for understanding the informativeness of different robot motions, as well as the impact of displacement estimation error on AOA accuracy. To our knowledge, this is the most general implementation of such a capability to date. Specifically, our framework is compatible with multi-robot systems that are mobile in 2D and 3D space, are heterogeneous, and must coordinate in potentially cluttered environments with large inter-robot distances as shown by our experimental results and application study.

\begin{acks}
We gratefully acknowledge the funding support through the NSF, Sloan Research Fellowship and MIT Lincoln Laboratories. Experiments were conducted in the Arizona State University Drone studio, the REACT Lab and the Stanford Robotics Lab.
\end{acks}

\begin{dci}
 The Authors declare that there is no conflict of interest.
\end{dci}

\begin{funding}
\textbf{The authors disclosed receipt of the following financial support for the research, authorship, and/or publication of this article:} This work was supported by Lincoln Labs Line grant, Sloan Research Fellowship 2021 [FG-2020-13998] and National Science Foundation awards [grant numbers: 1845225, 1718435 and 1837607]
\end{funding}
\theendnotes

\bibliographystyle{SageH}
\bibliography{references}
\vspace{-14pt}
\vspace{0.3in}
\section*{APPENDIX}
\appendix
\section{Cramer Rao Bound Development} \label{sec:CRB_math}

\textbf{I. 3D Spherical Helix Displacement:} We parameterize a spherical helix in spherical coordinates as given by:
\begin{align*}
   \xi_u= \tau, \hspace{0.1in} \varphi_u= c\tau, \hspace{0.1in}
   \rho_u= r
    \end{align*}
for a spherical helix of radius $r$, spiral climb rate of $c$, and parameterization $\tau\in[0,2\pi)$. We assume that each antenna element $u$ is uniformly spaced along the helix so that $\tau=u=\{0,\Delta,2\Delta,\hdots,(M-1)\Delta\}$ where $\Delta=\frac{2\pi}{M}$. Further, we assume a climb rate  $c=1$ for a simplification of our closed form, though any constant can be similarly substituted before integration. We set M approaching infinity (i.e. assume a very large number of packets received) to simplify analysis. So for a robot's 3D spherical helical displacement, the CRB terms are:
\begin{align*}
    &\hspace{-0.1in}\textbf{A}= \frac{\partial \textbf{a}^H(\theta,\phi)\partial \textbf{a}(\theta,\phi)}{\partial\theta \partial\theta}=\\
    &\sum_{u=1}^M \frac{4\pi^2 r^2}{\lambda^2}(\cos\theta \sin((u-1)\Delta)\cos(\phi - c(u-1)\Delta)\\
    &-\cos((u-1)\Delta)sin(\theta))^2\\
    & \text{Setting} \hspace{5pt} M \xrightarrow{} \infty, c=1\text{ we get}\\
    &\hspace{0.5in} \textbf{A}=-\frac{\pi ^2 r^2 \left(\cos ^2(\theta ) \cos (2 \phi )+\cos (2 \theta )-3\right)}{\lambda ^2}
\end{align*}
\begin{align*}
    &\textbf{B}= \frac{\partial \textbf{a}^H(\theta,\phi)\partial \textbf{a}(\theta,\phi)}{\partial\phi \partial\phi}=\\
    &\hspace{0.1in} \sum_{u=1}^M \frac{4\pi^2r^2}{\lambda^2}(\sin\theta \sin((u-1)\Delta)\sin(\phi - c(u-1)\Delta)))^2\\
    &\text{Setting} \hspace{5pt} M \xrightarrow{} \infty, c=1 \text{ we get} \\
    &\hspace{0.5in}\textbf{B} = \frac{\pi ^3 r^2 \sin ^2(\theta ) (\cos (2 \phi )+2)}{\lambda ^2}
\end{align*}
\begin{align*}
    &\textbf{C}= \frac{\partial \textbf{a}^H(\theta,\phi)\partial \textbf{a}(\theta,\phi)}{\partial\phi \partial\theta}=\\
    &\ \ \ \hspace{0.1in} \sum_{u=1}^M \frac{4\pi^2 r^2}{\lambda^2}(\cos\theta \sin((u-1)\Delta) \cos(\phi - c(u-1)\Delta))\\
    &-\cos((u-1)\Delta)\sin\theta)(\sin\theta \sin((u-1)\Delta)\\ &\ \ \ \ \ \ \ \ \ \ \ \ \  *\sin(\phi -c(u-1)\Delta))\\
    &\text{Setting} \hspace{5pt} 
    M \xrightarrow{} \infty, c=1 \text{ we get} \\  
    &\hspace{0.5in}\textbf{C}=\frac{2 \pi ^3 r^2 \sin (\theta ) \cos (\theta ) \sin (\phi ) \cos (\phi )}{\lambda ^2}.
\end{align*}

\noindent \textbf{II. Planar Circular Displacement:} We parameterize a planar circle as follows:
\begin{align*}
    \xi_u= \tau,\hspace{0.1in} \rho_u=r,\hspace{0.1in} \varphi_u=\cos^{-1}(0)=\pi/2
\end{align*}

The CRB equations reduce to
\begin{align*}
    &\textbf{A}= \sum_{u=1}^M \frac{4\pi^2r^2}{\lambda^2}\cos^2\theta\cos^2(\phi-(u-1)\Delta)\\
    & \hspace{0.1in}= \frac{4\pi^2r^2}{\lambda^2}\cos^2\theta \sum_{u=1}^M (\frac{1}{2}+\frac{\cos(2\phi-2(u-1)\Delta)}{2}) \\
    &\text{Setting} \hspace{0.1in} M \xrightarrow{} \infty, \text{we get}\\ 
    &\hspace{0.5in} \textbf{A}= \frac{2\pi^2r^2}{\lambda^2}\cos^2\theta
\end{align*}
and
\begin{align*}
    &\textbf{B}= \sum_{u=1}^M \frac{4\pi^2r^2}{\lambda^2}\sin^2\theta\sin^2(\phi-(u-1)\Delta)\\
    &\hspace{0.1in}=\frac{4\pi^2r^2}{\lambda^2}\sin^2\theta\sum_{u=1}^M (\frac{1}{2}-\frac{\cos(2\phi-2(u-1)\Delta)}{2})\\
    &\text{Setting} \hspace{0.1in}M \xrightarrow{} \infty, \text{we get}\\
    &\hspace{0.5in}\textbf{B}=\frac{2\pi^2r^2}{\lambda^2}\sin^2\theta
\end{align*}
and finally,
\begin{align*}
    &\textbf{C}= \sum_{u=1}^M -\frac{4\pi^2r^2}{\lambda^2}\cos\theta\cos(\phi-(u-1)\Delta)\\
    &\hspace{0.9in}\times (\sin\theta\sin(\phi-(u-1)\Delta))\\
    & \hspace{0.1in} = -\frac{4\pi^2r^2}{\lambda^2}\cos\theta\sin\theta\sum_{u=1}^M \sin(2\phi-2(u-1)\Delta) \\
    &\text{Setting} \hspace{0.1in}M \xrightarrow{} \infty, \text{we get}\\
    &\hspace{0.5in} \textbf{C}=0.
\end{align*}

\section{Notations and Terminology} \label{sec:notations}
\vspace{-10pt}
\nomenclature[C,01]{$t$}{time}
\nomenclature[C,02]{($\rho_i(t)$, $\varphi_i(t)$, $\xi_i(t)$)}{spherical coordinates of receiving robot's position denoting displacement, azimuth and elevation angles wrt. the first position along its displacement $\boldmath{\chi_i}\textbf{(t)}$}
\nomenclature[C,03]{$(\phi, \theta$)}{signal transmitting robot’s relative azimuth and elevation direction wrt. the initial position of the receiving robot}
\nomenclature[C,04]{($\hat{\phi},\hat{\theta}$)}{Estimated relative AOA to the transmitting robot (AOA peak in $F_{ij}(\phi,\theta)$)}
\nomenclature[C,05]{$(\phi_g, \theta_g$)}{True AOA (obtained from motion capture system) between transmitting and receiving robot}
\nomenclature[C,06]{$p_i(t)$}{groundtruth position of the robot $i$ at time $t$}
\nomenclature[C,07]{$(\kappa_\rho, \kappa_\phi, \kappa_\theta)$}{position estimation error in $\rho, \phi, \theta$ accordingly in polar coordinates }
\nomenclature[C,08]{$\delta$}{deviation around $\kappa_\phi$ in azimuth}
\nomenclature[C,09]{$\mu$}{deviation around $\kappa_\theta$ in elevation}
\nomenclature[C,10]{$\delta_{max}$}{max deviation around $\kappa_\phi$ in azimuth}
\nomenclature[C,11]{$\mu_{max}$}{max deviation around $\kappa_\theta$ in elevation}
\nomenclature[C,12]{$\delta_u$}{deviation around $\kappa_\phi$ in azimuth at position $p_i(t_u)$}
\nomenclature[C,13]{$\mu_u$}{deviation around $\kappa_\theta$ in elevation at position $p_i(t_u)$}
\nomenclature[C,14]{$\hat{p_i}(t_k)$}{Estimated position estimate of robot $i$ at time $t$}

\nomenclature[R,01]{$\chi_i(t_k)$}{Ground truth pose of robot $i$ at time $t_k$}
\nomenclature[R,02]{$\hat{\chi_i}(t_k)$}{Estimated pose of robot $i$ at time $t_k$}
\nomenclature[R,03]{$\boldmath{\chi_i}\textbf{(t)}$}{Local ground truth displacement of robot $i$ for $t\in[t_k,\hdots,t_l]$ which is a vector of positions $\chi_i(t)$ }
\nomenclature[R,04]{$\boldmath{\hat{\chi_i}}\textbf{(t)}$}{Local estimated displacement of robot $i$ for $t\in[t_k,\hdots,t_l]$ which is a vector of positions $\hat{\chi_i}(t)$}
\nomenclature[R,05]{$\mathcal{N}_i$}{all transmitting robots for which robot i can hear ping packets from}
\nomenclature[R,06]{$R_i(t)$}{Orientation of the robot i at time $t$ }
\nomenclature[R,07]{$R_{\epsilon}(t)$}{Accumulated error in robot orientation $R_i(t)$ at time $t$ }
\nomenclature[R,08]{$p_{\epsilon}(t_k)$ }{Accumulated error in robot position $p_i(t)$ at time $t$ }
\nomenclature[R,09]{($x_i(t)$, $y_i(t)$, $z_i(t)$)}{Robot position $p_i(t)$ in cartesian coordinates}
\nomenclature[R,10]{$SO(3)$}{Special Orthogonal group in 3D}
\nomenclature[R,11]{$\mathbb{R}^3$}{Euclidean space in 3D}
\nomenclature[R,12]{$\epsilon(t)$}{Residual error}
\nomenclature[R,13]{$\boldmath{\chi_{ij}}\textbf{(t)}$ }{Relative displacement of robot $i$ wrt. robot $j$}
\nomenclature[R,14]{$\boldmath{\tilde{\chi_i}}\textbf{(t)}$}{Displacement aligned with respect to true NORTH and DOWN direction}
\nomenclature[R,15]{$\tilde{R_{i}}$}{Rotation offset of robot $i$'s coordinate frame with respect to true NORTH and true DOWN direction}
\nomenclature[R,16]{$\tilde{t}_{ij}$}{Translation offset between robot $i$ and robot $j$}

\nomenclature[S,01]{$\lambda$}{Wireless signal wavelength}
\nomenclature[S,02]{$h_{ij}(t)$}{Wireless signal measurements between signal transmitting robot $j$ and receiving robot $i$}
\nomenclature[S,03]{$d_{ij}(t)$}{Ground truth distance between robot positions $p_i(t)$ and $p_j(t)$}
\nomenclature[S,04]{$\mathbf{Y_{ij}(t)}$}{Multi-antenna array output observation model}
\nomenclature[S,05]{$Y(t_k)$}{Received signal at robot position $p_i(t_k)$}
\nomenclature[S,06]{$\textbf{a}(\theta,\phi)\textbf{(t)}$}{Steering Vector}
\nomenclature[S,07]{$\textbf{n({t)}}$}{Received signal noise vector}
\nomenclature[S,08]{$M$}{Number of received signal packets}
\nomenclature[S,09]{$\mathbf{h_{ij}(t)}$}{Wireless channel}
\nomenclature[S,10]{$F_{ij}(\phi,\theta)$}{AOA profile, refers to relative paths a wireless signal traverses between a given pair of signal transmitting robot $j$ and receiving robot $i$ It is thus a 2D matrix calculated for all possible incoming signal directions along azimuth and elevation (360 x 180).}
\nomenclature[S,11]{$\sigma^2$}{Variance in signal noise}
\nomenclature[S,12]{$I$}{Identity matrix}
\nomenclature[S,13]{$\norm{h}$}{Magnitude of wireless signal} 
\nomenclature[S,14]{$a(\theta,\phi)(t)$}{Element of steering vector $\textbf{a}(\theta,\phi)\textbf{(t)}$ at time $t$ }
\nomenclature[S,15]{$AOA_{max}$}{Strongest signal direction i.e maximum magnitude path, in the full AOA profile $F_{ij}(\phi,\theta)$}
\nomenclature[S,16]{$\Delta_f$}{Frequency offset}
\nomenclature[S,17]{$h^r_{ij}(t)$}{Reverse wireless signal channel between robot $i$ and robot $j$}
\nomenclature[S,18]{$\hat{h}_{ij}(t)$}{Observed forward wireless channel}
\nomenclature[S,19]{$\hat{h}^r_{ij}(t)$}{Observed reverse wireless channel}
\nomenclature[S,20]{$\textbf{A, B, C}$}{Partial derivative terms in Cramer Rao Bound formulation for \emph{informativeness} of a displacement geometry.}
\nomenclature[S,21]{FIM}{Fisher Information Matrix}
\nomenclature[S,22]{SNR}{Signal To Noise Ration}
\nomenclature[S,23]{$\Delta$}{$\frac{2\pi}{M}$}
\nomenclature[S,24]{$\tau$}{Uniform spacing between antenna elements used for CRB analysis of special geometries.}
\nomenclature[S,25]{c}{climb rate for simulated robot motion}
\nomenclature[S,26]{\textbf{Det}}{Determinant}
\nomenclature[S,27]{$\angle\textit{F}(\phi, \theta)$}{Phase component of the AOA profile $F_{ij}(\phi, \theta)$}

\printnomenclature

\end{document}